\documentclass{article}

\usepackage[final,nonatbib]{neurips_2022}

\usepackage{subfigure}
\usepackage{soul}
\usepackage[toc,page,header]{appendix}
\usepackage{minitoc}
\usepackage[colorlinks=true,linkcolor=blue]{hyperref}

\usepackage[utf8]{inputenc} 
\usepackage[T1]{fontenc}    
\usepackage{hyperref}       
\usepackage{url}            
\usepackage{booktabs}       
\usepackage{amsfonts}       
\usepackage{nicefrac}       
\usepackage{microtype}      
\usepackage{wrapfig}
\usepackage[dvipsnames]{xcolor}
\usepackage{amsmath}
\usepackage{amssymb}
\usepackage{graphicx}
\usepackage{stackengine,collcell,makecell}
\usepackage{multirow}
\usepackage{lipsum}  
\usepackage[ruled]{algorithm2e}
\usepackage{caption}
\usepackage{pifont}
\usepackage{physics}
\usepackage[numbers, sort&compress]{natbib}
\usepackage[para]{footmisc}

\title{Data-IQ: Characterizing subgroups with heterogeneous outcomes in tabular data}
\author{%
  Nabeel Seedat\\
  University of Cambridge\\
  \texttt{ns741@cam.ac.uk} \\
   \And
   Jonathan Crabb\'{e} \\
   University of Cambridge \\
   \texttt{jc2133@cam.ac.uk} \\
   \AND
   Ioana Bica \\
   University of Oxford \\
   The Alan Turing Institute \\
   \texttt{ioana.bica@eng.ox.ac.uk} \\
   \And
   Mihaela van der Schaar \\
   University of Cambridge \\
   The Alan Turing Institute \\
   UCLA\\
   \texttt{mv472@cam.ac.uk} \\
}

\begin{document}

\doparttoc
\faketableofcontents

\maketitle

\newcommand{\X}{\mathcal{X}}
\newcommand{\Y}{\mathcal{Y}}
\newcommand{\Z}{\mathcal{Z}}
\renewcommand{\H}{\mathcal{H}}
\newcommand{\C}{\mathcal{C}}
\newcommand{\G}{\mathcal{G}}
\newcommand{\Dtrain}{\mathcal{D}_{\textrm{train}}}
\newcommand{\Dtest}{\mathcal{D}_{\textrm{test}}}
\newcommand{\D}{\mathcal{D}}
\renewcommand{\P}{\mathcal{P}}
\newcommand{\Ph}{\hat{\P}}
\newcommand{\R}{\mathbb{R}}
\newcommand{\N}{\mathbb{N}}
\newcommand{\xh}{\hat{x}}
\newcommand{\yh}{\hat{y}}
\newcommand{\zh}{\hat{z}}
\newcommand{\rxh}{\hat{X}}
\newcommand{\ryh}{\hat{Y}}
\newcommand{\rzh}{\hat{Z}}
\newcommand{\dzh}{\delta \hat{z}}
\newcommand{\dxh}{\delta \hat{x}}
\newcommand{\dyh}{\delta \hat{y}}
\newcommand{\rdzh}{\delta \hat{Z}}
\newcommand{\rdxh}{\delta \hat{X}}
\newcommand{\rdyh}{\delta \hat{Y}}
\newcommand{\proba}[2][\C]{p_{#1}\left(#2 \right)}
\newcommand{\cproba}[3][\C]{\proba[#1]{#2 \mid #3}}

\newcommand{\set}[1]{\left\{ #1 \right\}}

\newcommand{\cmark}{\ding{51}}
\newcommand{\xmark}{\textcolor{lightgray}{\ding{55}}}

\newcommand{\easy}{\textcolor{ForestGreen}{Easy}}
\newcommand{\ambiguous}{\textcolor{Orange}{Ambiguous}}
\newcommand{\hard}{\textcolor{Red}{Hard}}
\newcommand{\expect}[2]{\mathbb{E}_{#1} \left[ #2 \right]}
\newcommand{\variance}[2]{\mathbb{V}_{#1} \left[ #2 \right]}
\newcommand{\val}{v_{\mathrm{al}}}
\newcommand{\vep}{v_{\mathrm{ep}}}
\newcommand{\xtrain}{x_{\mathrm{train}}}
\newcommand{\xtest}{x_{\mathrm{test}}}
\newcommand{\Clower}{C_{\mathrm{low}}}
\newcommand{\Cupper}{C_{\mathrm{up}}}

\begin{abstract}
High model performance, on average, can hide that models may systematically underperform on subgroups of the data. We consider the tabular setting, which surfaces the unique issue of \emph{outcome heterogeneity} - this is prevalent in areas such as healthcare, where patients with \emph{similar features} can have \emph{different outcomes}, thus making reliable predictions challenging. To tackle this, we propose \emph{Data-IQ}, a framework to systematically stratify examples into subgroups with respect to their outcomes. We do this by analyzing the behavior of individual examples during training, based on their predictive confidence and, importantly, the aleatoric (data) uncertainty. Capturing the aleatoric uncertainty permits a principled characterization and then subsequent stratification of data examples into three distinct subgroups ($\easy, \ambiguous, \hard$). We experimentally demonstrate the benefits of Data-IQ on four real-world medical datasets. We show that Data-IQ's characterization of examples is most robust to variation across similarly performant (yet different) models, compared to baselines. Since Data-IQ can be used with \emph{any} ML model (including neural networks, gradient boosting etc.), this property ensures consistency of data characterization, while allowing flexible model selection. Taking this a step further, we demonstrate that the subgroups enable us to construct new approaches to both feature acquisition and dataset selection. Furthermore, we highlight how the subgroups can inform reliable model usage, noting the significant impact of the $\ambiguous$~ subgroup on model generalization. 
\end{abstract}

\section{Introduction} \label{sec:introduction}
Most machine learning models are optimized using empirical risk minimization (ERM), to maximize average performance during training \cite{liu2021just}. However, in real-world settings, while models may perform well on average, they might underperform on specific subgroups of data \cite{recht2019imagenet, duchidistributionally, hashimoto2018fairness}.  Most of the current literature has focused on this problem in computer vision, where the underperforming subgroups are typically associated with data examples that have spurious correlations \cite{liu2021just,sohoni2020no} or mislabelling \cite{pleiss2020identifying}. 

In this paper, we focus on tabular data, the most ubiquitous format in medicine and finance, where data is based on relational databases  \citep{borisov2021deep, yoon2020vime}.  Specific to the tabular setting, we formalize an understudied source of underperformance, namely \emph{heterogeneity of outcomes}. This phenomenon is vital in healthcare, where patients with \emph{similar features} can have \emph{different outcomes} \cite{huo2021sparse,gartlehner2012clinical, yoon2016discovery}. For example, \cite{oh2021two} showed that prognostic models for risk prediction perform well on average, but underperform on specific cancer types due to heterogeneity of risk (outcome). Prior works have audited subgroups belonging to sensitive attributes (e.g. demographics, race or gender), as it is well-known that ML models generally underperform on these subgroups \cite{jiang2021towards,chen2021ethical}. However, this approach is limiting, as it needs the sensitive attributes to be specified, and it also does not capture the case where complex feature interactions may lead to underperformance. 
\begin{figure}[t]
    \centering
    \includegraphics[width=0.835\textwidth]{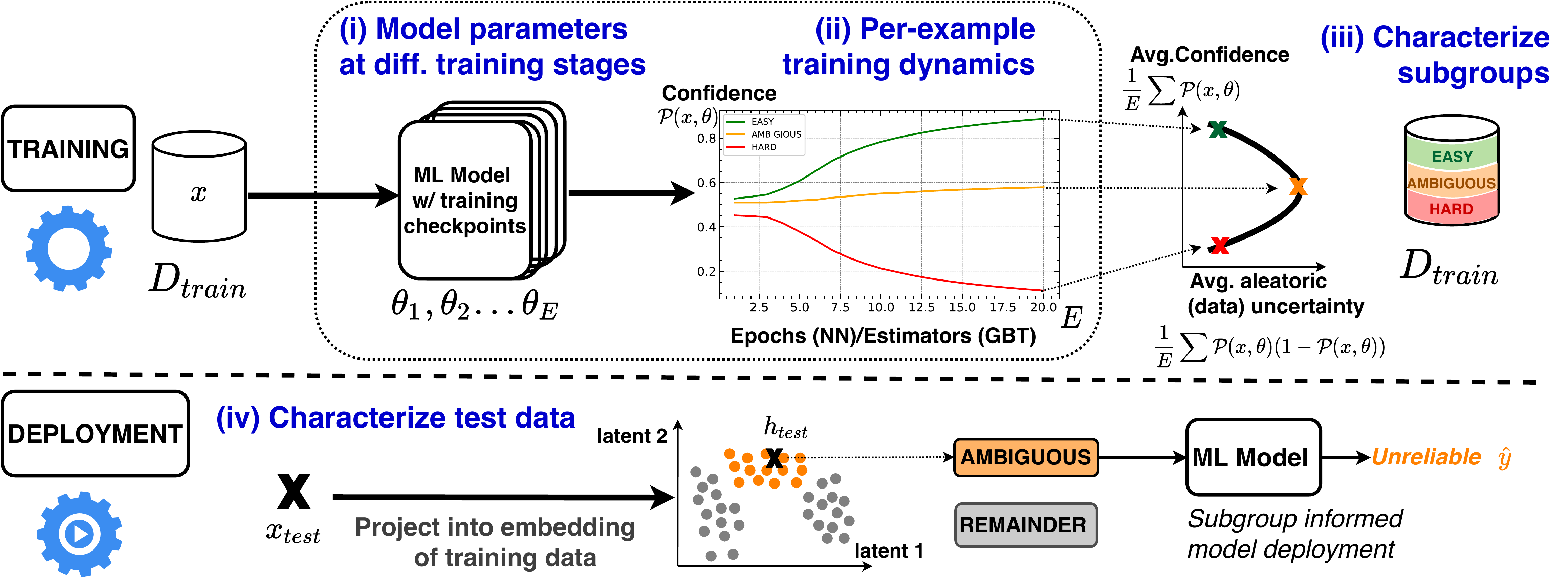}
    \caption{\footnotesize{Data-IQ systematically characterizes data into subgroups using \textcolor{blue}{(i)} any ML model, trained in stages (epochs/iterations) \& can be checkpointed. At training time, Data-IQ  \textcolor{blue}{(ii)} leverages the model's checkpoints to analyze the training behavior of individual examples and \textcolor{blue}{(iii)} characterizes each example based on its aleatoric (data) uncertainty \& prediction confidence. \textcolor{blue}{(iv)} At deployment, the subgroups inform model usage by embedding the training set in a representation space and characterizing the new data points in this representation space.}}
    \label{fig:overview}
     \vspace{-3mm}
    \rule{\linewidth}{.75pt}
    \vspace{-5mm}
\end{figure}

We take a different approach to automatically stratify data into subgroups, usable with \textit{any ML model trained in stages (epochs/iterations)}; e.g., neural networks, gradient boosting etc. Specifically, we study the behavior of individual examples during training, called \emph{training dynamics}. This allows us to formalize that examples can lie on the spectrum from easy to hard to predict. More concretely, let's consider the task of patient mortality prediction.  Based on their features, sicker patients more often have a mortality event. Thus, they are easy to learn for any model and will be predicted \emph{correctly} with \emph{high confidence} ($\easy$). However, a subgroup might have a heterogeneous outcome: survival despite their poor prognosis. This heterogeneity could result from randomness, making it practically impossible for a model to learn. These examples will be predicted \emph{incorrectly} yet with \emph{high confidence} (or equivalently have low confidence for the correct class) ($\hard$). In tabular data, there are also examples with inherent ambiguity where the predicted probability for the correct class remains low. They appear where the current features are insufficient to distinguish the example correctly, regardless of the model used \cite{ho2002complexity,lorena} ($\ambiguous$).  These subgroups naturally arise in real data; see Fig.\ref{fig:overview} (ii).

Identifying these subgroups is practically valuable, as improving accuracy and robustness often depends on the data's characteristics and quality \cite{jain2020overview,gupta2021dataB,renggli2021data,Sambasivan}.  As mentioned in \cite{ wagstaff2012machine,Sambasivan}, the ``data'' work is often undervalued as merely operational, yet failing to account for it can have immense practical harm \cite{Sambasivan,oh2021two}. Consequently, our goal is to build a systematic framework with the following desired properties (\textbf{P1-P4}), motivated by the considerations of practitioners at various stages of the ML pipeline. In satisfying \textbf{P1-P4}, we seek to address the ``dire need for an ML-aware data quality that is not only principled, but also practical for a larger collection ($\ldots$) of ML models'' \cite{renggli2021data}:\\\\
\textbf{(P1) Robust data characterization}: the characterization of data examples should be robust, such that it is consistent across similar performing models, that have different architectures/parametrizations.\\
\textbf{(P2) Principled data collection}: the characterization should be informative and actionable, providing practitioners insights that enable both quantitative feature collection and dataset selection.\\
\textbf{(P3) Reliable model deployment}: the characterization should enable reliable model usage, both by unmasking unreliable subgroups or using the subgroups to tailor the data for better performance.\\
\textbf{(P4) Plug \& play}: the characterization should be applicable to a variety of ML models widely used on tabular data, including neural networks, gradient boosting (and variants) etc.

To fulfill \emph{P1-P4}, we propose \textbf{Data-IQ}, a systematic framework that characterizes examples based on the inherent qualities (IQ) of the data; at both training and deployment time. As outlined in Fig.\ref{fig:overview}, Data-IQ leverages confidence and in the \emph{``data-centric AI''} spirit focused on the data: aleatoric uncertainty (i.e. uncertainty inherent to the data). This permits Data-IQ to provide ML-aware data quality that is principled and practical for a variety of ML models, making the following contributions:

\newpage
\textbf{Contributions:} 
\textbf{\textcolor{BrickRed}{\textcircled{1}}} Data-IQ models the aleatoric (data) uncertainty, which permits subgroup identification that is most robust to variation across different yet similar performing models/parameterizations, compared to other baselines, i.e. \emph{P1}.
\textbf{\textcolor{BrickRed}{\textcircled{2}}} Data-IQ aids with principled data collection \emph{P2} in two ways: Firstly, it permits to quantify the value of an acquired feature by measuring how the feature reduces the aleatoric uncertainty of the example. This information enables a more principled approach to feature acquisition. Secondly, it permits to compare datasets based on the proportion of ambiguous examples. We demonstrate that the proportions link to how well a model trained with the dataset generalizes. \textbf{\textcolor{BrickRed}{\textcircled{3}}}  Experimentally, the subgroups identified by Data-IQ can
inform reliable model deployment, i.e. P3. We highlight cases, where assessment on average might
mask unreliable performance, including data sculpting, model robustness, and uncertainty estimation
methods. 
\textbf{\textcolor{BrickRed}{\textcircled{4}}} Data-IQ by construction is ``plug-and-play'' i.e. \emph{P4} with \emph{any} ML model that can be checkpointed, granting practitioners flexibility to apply Data-IQ to their model of choice.

\section{Related work} \label{sec:related}
This paper primarily engages with the literature on data characterization and contributes to the nascent area of data-centric AI \cite{ng2021,polyzotis2021can}. An extended discussion of related work is found in Appendix \ref{extended_related}. 

\textbf{Data characterization.} 
The literature to characterize data samples has used a myriad of different metrics. However, their goals have typically been different, such as spurious correlation or mislabelling, compared to Data-IQ, whose goal is to characterize subgroups with respect to the outcome predictions. Furthermore, none of these methods completely addresses all the desired properties (P1-P4). The closest to our work on data quality is Data Maps \cite{swayamdipta2020dataset}. A key contrast to Data-IQ is that Data Maps use confidence and prediction variability to flag instances. In Sec. \ref{sec:formulation}, we show that this prediction variability corresponds to the model uncertainty (i.e. epistemic uncertainty). Alternatively, Data-IQ takes a different and more principled approach, capturing the inherent data uncertainty (known as \emph{aleatoric uncertainty}) \cite{kendall2017uncertainties}. 
Epistemic uncertainty is reducible by collecting more data. In comparison, aleatoric uncertainty is irreducible even with more samples. This is due to the fact that it captures properties inherent to the data \cite{der2009aleatory,kendall2017uncertainties,gavves}; only better features can reduce the aleatoric uncertainty \cite{gavves}. Later in Fig. \ref{fig:compare}, we show on real data that capturing the aleatoric uncertainty allows Data-IQ to be more robust to variation across different models, compared to Data Maps (\emph{P1}). This allows practitioners to characterize their data in such a way that the insights are more consistent.  We further show theoretically in Sec. \ref{formulation:stratification}, why the characterization by Data-IQ indeed provides a more principled definition for \textcolor{Orange}{\emph{Ambiguous}} examples, compared to Data Maps. 

Besides Data Maps, other related methods address specific computer vision problems:
identifying mislabelled images using area under the margin (AUM) \cite{pleiss2020identifying}, gradient norm to identify ``important examples'' to aid pruning during training \cite{paul2021deep}, or underperformance due to spurious image correlations \cite{liu2021just}. The tabular setting considered in this paper requires new methods, due to the specific problem of heterogeneous outcomes for examples with similar features (i.e. ``feature collision''). The ambiguity in the tabular, ``feature collision'' sense, is different or non-existent in modalities such as images.

\textbf{Data-Centric AI.}
The assessment of data quality is a critical but often overlooked problem in ML  \cite{Sambasivan}. While the focus in ML is typically on optimizing models, the task of ensuring high quality data (or even improving one's data) can be equally valuable to improving performance \cite{Sambasivan,jain2020overview}. Even when it is considered, the process of assessing datasets is adhoc or artisinal \cite{Sambasivan,ng2021,chug2021statistical,seedat2022data}. The recent growth of the data-centric AI space aims to build systematic tools for ``data collection, labeling, and quality monitoring processes for datasets to be used in machine learning'' \cite{seedat2022data,polyzotis2021can}. Data-IQ contributes to this nascent body of work, specifically around ML-aware data quality monitoring \cite{renggli2021data}.

\section{Formulation} \label{sec:formulation}
This section gives a detailed formulation of Data-IQ and motivates our proposed example stratification that uses aleatoric uncertainty and confidence. We then describe how Data-IQ stratifies examples into subgroups at both training and testing time. Finally, we show Data-IQ's formulation permits usage with \emph{any} ML model trained in stages, e.g. neural networks, GBDTs etc, unlike other approaches.

\subsection{Preliminaries}

We consider the typical supervised learning setting, where the aim is to assign an input $x \in \X \subseteq \R^{d_X}$ to a class $y \in \Y \subset \N$. We have a dataset $\mathcal{D}$  with $N \in \N^*$ examples, i.e. $\mathcal{D} = \set{(x^n, y^n) \mid n \in [N]}$ drawn IID from an unknown distribution. Our goal is then to learn a model $f_\theta: \X \rightarrow \Y$, parameterized by $\theta \in \Theta$. 
Typically, the parameters $\theta$ are learned to minimize empirical risk, by minimizing the average training loss , i.e. $\text{ERM}(\theta) = \frac{1}{n} \sum_{i=1}^n \ell(x_i, y_i; \theta)$,
with a loss function $\ell : \X \times \Y \times \Theta \to \R^+$. 

This brings us to the essence of the problem: ``not all examples are created equally''. e.g. patients with similar features might have heterogeneous outcomes, reflected in their labels $y$ being different. These correspond to subgroups within $\Dtrain$ on which a predictive model might systematically underperform. We formalize this concept of hidden heterogeneous subgroups by assigning to each example $x^n$ a hidden \emph{subgroup} label $g^n\,{\in}\,\G$, where $\G=\{\easy, \ambiguous, \hard \}$. Before giving a precise description of how those group labels are assigned, it is useful to detail the context. Several works have established that the training dynamics of a model, contains signal about the quality of the data itself~\cite{arpit2017closer,arora2019fine,li2020learning}. For instance, it takes more epochs/iterations for a model to assign the correct label to noisier/more difficult training examples. With Data-IQ, we build on those observations and assign a label $g^n$ to each example $x^n$ by studying its training dynamic, which is then used to estimate the aleatoric uncertainty and predictive confidence of each example. The following sections detail how this is done and how this contrasts with existing approaches.

\subsection{Uncertainty decomposition during training}
Recall that practitioners desire flexibility in the choice of the model. Hence,  we focus on \emph{any} ML model that is trained in stages and can be checkpointed during training, $f_\theta: \X \rightarrow \Y$ parameterized by $\theta$ and on a given example \emph{from the training set} $(x, y) \in \Dtrain$. Assume that the model $f_{\theta}$ corresponds to a conditional categorical distribution, assigning a probability to each class given the input $x$: $f_{\theta}(x) = P(Y \mid X=x, \vartheta =\theta)$. During iterative training, the model parameters $\theta$ vary, where over $E \in \N^*$ epochs/iterations, these parameters take $E$ different values at each checkpoint, i.e. $\theta_1 , \theta_2, \dots, \theta_E$. Since our analysis relies on the model's training dynamics, we want to take those different parameters into account. For the sake of notation, we introduce a random variable $\vartheta$ that has an empirical distribution over this set of parameters captured through the training process $\vartheta \sim P_{\mathrm{emp}}(\set{\theta_e \mid e \in [E]})$. The variability of the model's parameters at training time is then reflected by the variance $\variance{\vartheta}{\cdot}$.

The uncertainty we model is based on the random variable $Y \mid X = x$ that represents the possible labels given the input $x$. Since the ground-truth label $y$ is available for training examples, we would like to distinguish between 2 cases: \textcircled{1} the predicted label corresponds to the ground-truth label $Y = y$ and \textcircled{2} the predicted label is different from the ground-truth label $Y \neq y$. To this end, we introduce a binary random variable $\tilde{Y}$ that is set to one when the predicted label equals the ground-truth label ($\tilde{Y} = 1$ if $Y = y$) and that is zero otherwise ($\tilde{Y} = 0$ if $Y \neq y$). As discussed earlier, we are interested in the uncertainty on the predictive random variable $\tilde{Y} \mid X = x$. This uncertainty is modeled by the variance $v(x) = \variance{\tilde{Y} \mid X}{\tilde{Y} \mid X = x}$. We will now show that this quantity can be evaluated with the model predictions. 

We start by noting that the definition of $\tilde{Y}$ implies that $\tilde{Y} \mid X=x , \vartheta = \theta$ is a Bernoulli random variable with parameter\footnote{In this case, $[f_{\theta}(x)]_y$ denotes the component $y$ of the probability vector $f_{\theta}(x)$.} $\P(x, \theta) = P(Y=y \mid X=x , \vartheta = \theta) = [f_{\theta}(x)]_y$. From this observation, we can decompose $v(x)$ with the law of total variance and make each term explicit:

\vspace{-2mm}
\begin{equation} \label{eq:variance_decomposition}
\begin{aligned}
v(x) &=\underbrace{\variance{\vartheta}{\expect{\tilde{Y}|X,\vartheta}{\tilde{Y}|X=x,\vartheta}}}_{\text {Bernoulli Mean: $\mathcal{P}(x,\vartheta)$}}+\underbrace{\expect{\vartheta}{\variance{\tilde{Y}|X,\vartheta}{\tilde{Y}|X=x,\vartheta}}}_{\text{Bernoulli Var: $\mathcal{P}(x,\vartheta)(1-\mathcal{P}(x,\vartheta))$}}  \\
&=\underbrace{\variance{\vartheta}{\mathcal{P}(x,\vartheta)}}_{\text{Epistemic uncertainty: $\vep(x)$}}+\underbrace{\expect{\vartheta}{\mathcal{P}(x,\vartheta)(1-\mathcal{P}(x,\vartheta))}}_{\text{Aleatoric uncertainty: $\val(x)$}} .
\end{aligned}
\end{equation}

In Eq. \eqref{eq:variance_decomposition}, we have split the overall uncertainty into two components: \emph{epistemic} and \emph{aleatoric uncertainty}. This type of decomposition is similar to those in the context of Bayesian neural networks~\cite{kwon2018uncertainty,seedat2019towards}. To understand the distinction between uncertainties, it is useful to closely examine the variances in the \emph{first} equality of Eq. \eqref{eq:variance_decomposition}. For \emph{epistemic uncertainty} $\vep$, variance is evaluated on the \emph{model parameters $\vartheta$}. Hence, epistemic uncertainty originates from the fact that a model's predictions oscillate when we change its parameters. For the \emph{aleatoric uncertainty} $\val$, the variance is evaluated on the \emph{predicted label $\tilde{Y} \mid X, \vartheta$}. Hence, the variability originates from the inability to predict the correct label with high confidence. While existing works use epistemic uncertainty to stratify examples, we argue that aleatoric uncertainty is a better principled choice to capture the inherent data uncertainty.

\subsection{Stratification based on data uncertainty}\label{formulation:stratification}

We now explain how the above notion of uncertainty permits to assign a group label $g \in \G$ to each training example $x$. First, we use the empirical distribution $\vartheta \sim P_{\mathrm{emp}}(\set{\theta_e \mid e \in [E]})$ to explicitly write the two types of uncertainties in~ Eq. \eqref{eq:variance_decomposition},where $\Bar{\P}(x) = \nicefrac{1}{E} \sum_{e=1}^E \P(x, \theta_e)$:
\vspace{-2mm}
\begin{align}
    \vep(x) = \frac{1}{E} \sum_{e=1}^E \left[ \P(x, \theta_e) - \Bar{\P}(x) \right]^2 \hspace{.5cm} \val(x) = \frac{1}{E} \sum_{e=1}^{E} \mathcal{P}(x,\theta_e)(1-\mathcal{P}(x,\theta_e)),
\end{align}

\textbf{Stratification at training time.} Before giving a precise definition of the group labels, let us give an intuitive definition for each group. \textcircled{1} $\easy$: examples that have low data uncertainty that the model can correctly predict with high confidence, \textcircled{2} $\ambiguous$: examples that have high data uncertainty, hence the model is unable to predict with confidence and \textcircled{3} $\hard$: examples that have low data uncertainty that the model is unable to predict (i.e. predicted incorrectly yet with high confidence or equivalently have low confidence for the correct class). We note that we need the model's prediction for the ground-truth class to delineate $\easy$ and $\hard$ examples. In practice, we use the model's average confidence for the ground-truth class $\Bar{\P}(x)$ defined previously for this purpose. We make use of this concept to detail how labels are assigned to training examples $(x,y) \in \Dtrain$:
\begin{align} \label{eq:groups_training}
\vspace{-10mm}
    g(x,\Dtrain) = \begin{cases}
$\easy$ & \text{if} \ \Bar{\P}(x) \ge \Cupper \ \wedge \ \val(x) < P_{50}\left[ \val(\Dtrain) \right]  \\
$\hard$ & \text{if} \ \Bar{\P}(x) \le \Clower \ \wedge \ \val(x) < P_{50}\left[ \val(\Dtrain) \right]  \\
$\ambiguous$ & $\text{otherwise}$  \\
\end{cases}
\vspace{-5mm}
\end{align}
\vspace{-0.5mm}%
where $\Cupper$ and $\Clower$ are upper and lower confidence threshold resp. and $P_{n}$ the n-th percentile. We provide a practical method to set $\Cupper$ and $\Clower$, applicable to any dataset in Appendix \ref{sec:appendixA}.

\begin{wrapfigure}{r}{0.45\textwidth}
\captionsetup{font=footnotesize}
\vspace{-5mm}
  \centering
    \includegraphics[trim=8cm 0cm 3.5cm 0cm, width=0.45\textwidth]{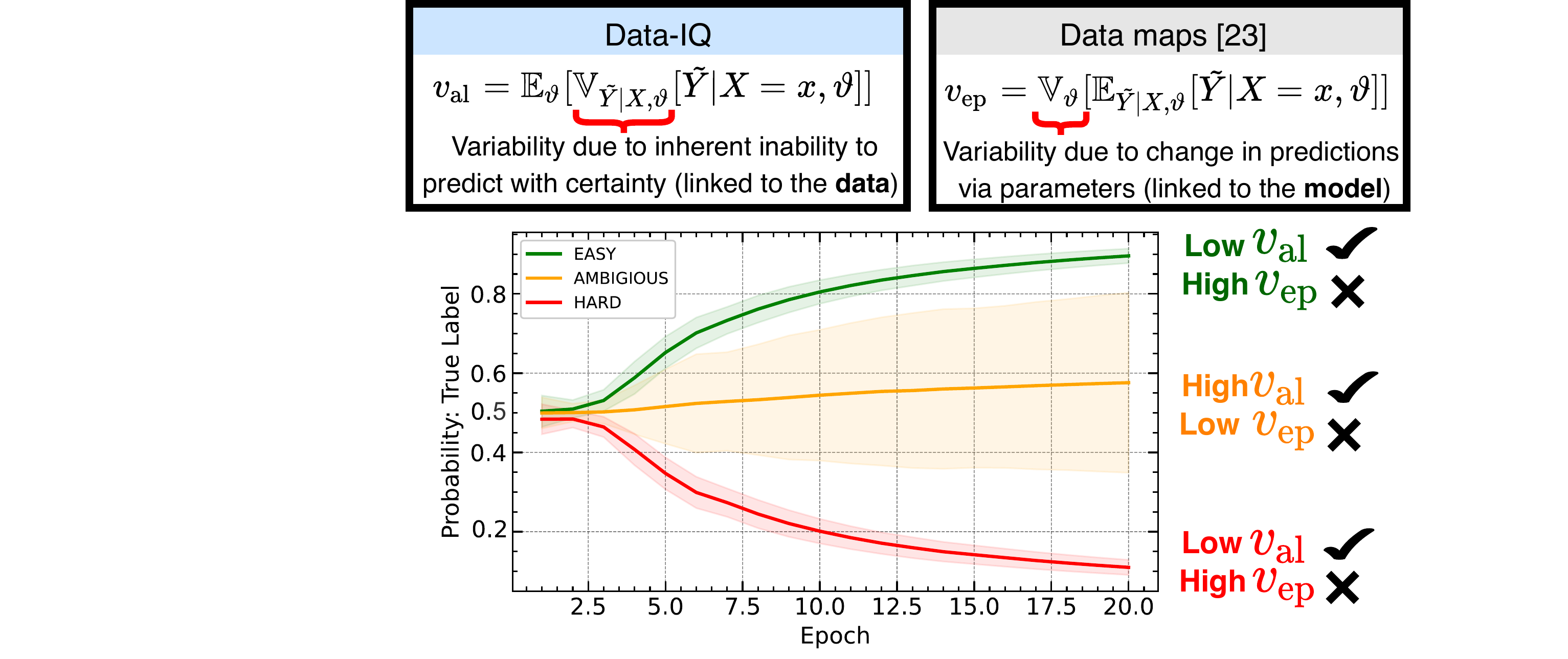}
   \vspace{-7mm}
   \caption{\footnotesize{Example outlining the differences between $\val$ (Data-IQ) and $\vep$ (Data Maps) \& showing the type of uncertainty matters.}}
    \vspace{-3mm}
    \rule{\linewidth}{.45pt}
   \vspace{-4mm}
    \label{fig:evolution_compare}
\end{wrapfigure}

In contrast to Data-IQ which uses Aleatoric uncertainty $\val(x)$; Data Maps~\cite{swayamdipta2020dataset} identifies ambiguous training examples $(x, y) \in \Dtrain$ as those with high epistemic uncertainty $\vep(x)$. We consider a typical scenario to see how this characterization might cause problems – illustrated in Fig. \ref{fig:evolution_compare}. Consider an example $x$ in which the model cannot classify confidently during the entire training $\P(x, \theta_e) = 0.5 \ \forall e \in [E]$. In this case, the epistemic uncertainty $\vep(x)$ vanishes, as the prediction is consistently unconfident (i.e. low variability of the model predictions). This implies that Data Maps would consider this example as non-ambiguous, despite the ambiguous model prediction for this example. This problem can be traced back to the definition of epistemic uncertainty, which measures the sensitivity of a model prediction with respect to the model's parameters.

A more principled definition for ambiguous examples should capture examples for which the model cannot predict the appropriate label with high confidence (i.e. data uncertainty). This is precisely what the aleatoric uncertainty $\val(x)$ captures (Data-IQ). Furthermore, it is easy to verify that the previous example $\P(x, \theta_e) = 0.5 \ \forall e \in [E]$ maximizes the aleatoric uncertainty (see Fig. \ref{fig:evolution_compare}). Since high aleatoric uncertainty captures ambiguous examples for various values of the model's parameters, we believe that it better reflects the inherent quality of the data. In that sense, we expect this quantity to be more stable and robust to variation for different ML model parameters/architecture changes (\emph{P1}). We experimentally validate the consistency in Sec. \ref{sec:experiments}.

\textbf{Stratification at inference time.} Most previous methods are only applicable at training time. To address this limitation and improve the practical utility of our method, we also stratify examples into subgroups at deployment time. However, if we try to apply the above stratification for incoming data at deployment time, we face a problem: $\Bar{\P}(x)$ requires the ground-truth class $y$.

For this reason, we follow an alternative approach based on representation learning that does not require access to ground-truth labels. The idea is the following: we construct a low-dimensional UMAP embedding~\cite{mcinnes2018umap} $h : \X \rightarrow \H$ of the training set's examples $\xtrain \in \Dtrain$. In doing this, we note two things (see Appendix \ref{sec:appendixC}): \textcircled{1} $\ambiguous$ examples have distinctive features and are clustered in embedding space. Thus, it is possible to distinguish the $\ambiguous$ examples using the embedding. \textcircled{2} It is not possible to reliably distinguish $\easy$ examples from $\hard$ examples based on the embedding, because $\hard$ examples are a minority with outcome randomness that have similar features, as the $\easy$ examples. Combining these observations, we note it is possible to identify $\ambiguous$ test examples. This label is assigned by computing the related embedding $h(\xtest)$ and comparing this embedding to the nearest neighbor embedding from the training set, i.e. $d[h(\xtest),h(\xtrain)] \forall \xtrain \in \Dtrain$. For models like neural networks with an implicit representation space, the same analysis can be done using the model's representation space.

\subsection{Using Data-IQ with a variety of models, beyond Neural Networks (P4)}\label{p4-formulation}

The baseline methods discussed are primarily applicable only to neural networks. However, practically in tabular settings (e.g. healthcare/finance etc), practitioners often use other highly performant iterative learning algorithms such as Gradient Boost Decision Trees (GBDTs) or variants \cite{shwartz2022tabular,borisov2021deep}.  Data-IQ's formulation by construction is naturally adaptable to \emph{any} ML model trained in stages, that can be checkpointed. This satisfies \emph{P4}, which allows practitioners the flexibility to use Data-IQ with their application-specific model of choice.  Appendix \ref{sec:appendixA} provides guidelines, space and time considerations, as well as, discussing the specifics of how Data-IQ is easily adapted, for example to GBDTs.

\section{Experiments} \label{sec:experiments}
This section presents a detailed empirical evaluation demonstrating that Data-IQ \footnote{https://github.com/seedatnabeel/Data-IQ}\footnote{https://github.com/vanderschaarlab/Data-IQ}  satisfies (\textbf{P1}) Robust data characterization, (\textbf{P2}) Principled data collection and (\textbf{P3}) Reliable Model Deployment, introduced in Sec.\ref{sec:introduction}. Recall that (\textbf{P4}) Plug and play is satisfied by construction of Data-IQ. 

\textbf{Datasets.} We conduct experiments on four real-world medical datasets, with diverse characteristics (different sizes, binary/multiclass, varying degrees of task difficulty etc) and highlight real-world applicability with heterogeneous patient outcomes: (1)  Covid-19 dataset of Brazilian patients \cite{baqui2020ethnic}, (2) Prostate cancer datasets from both the US \cite{duggan2016surveillance} and UK \cite{prostate}, (3) Support dataset of seriously ill hospitalized adults \cite{knaus1995support}, (4) Fetal state dataset of cardiotocography \cite{ayres2000sisporto}. 
We describe the datasets in greater detail in Appendix \ref{sec:appendixB}, along with further experimental details. \emph{We observe similar performance across different datasets, but given the space limitations, we typically show pertinent results for a single dataset, and include results for the other datasets in Appendix \ref{sec:appendixC}.}

\subsection{(\textbf{P1}) Robust data characterization}\label{p1-exp}

\paragraph{Robustness to variation.}\label{robustness-sec}

\begin{wrapfigure}{r}{0.55\textwidth}
\vspace{-15mm}
  \centering
    \includegraphics[width=0.5\textwidth]{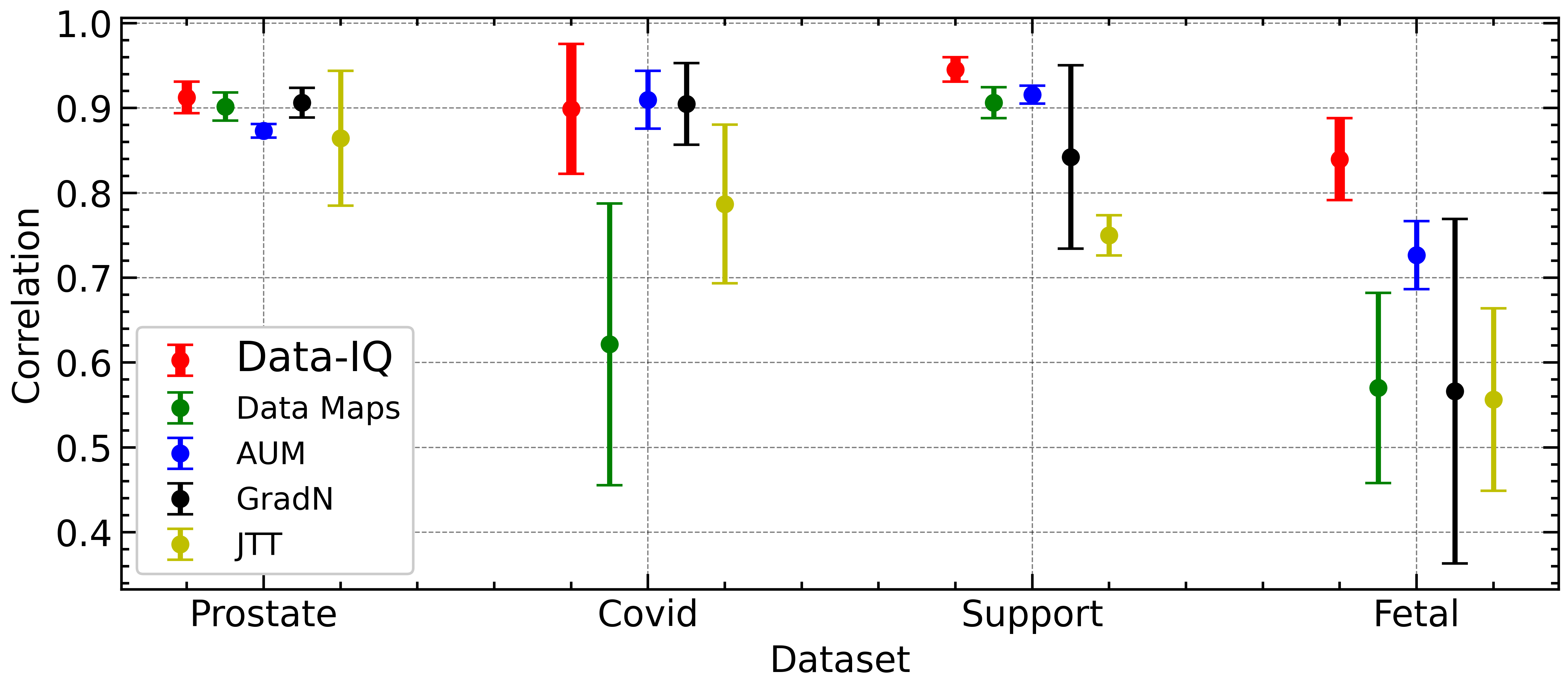}
   \vspace{-3mm}
   \caption{\footnotesize{Robustness to variation across models based on Spearman correlation, where Data-IQ has the highest correlation (i.e. consistency) across all datasets.}}
    \vspace{-3mm}
    \rule{\linewidth}{.5pt}
   \vspace{-8mm}
    \label{fig:comparison}
\end{wrapfigure}

As per \emph{P1}, we desire that Data-IQ identifies subgroups in a manner robust to variation across different models. This would allow a practitioner to obtain consistent insights about their data even when using different model architectures/parameterizations. 
When comparing the different methods from Sec. \ref{sec:related}, we note that each method has its own specific metric used to characterize examples (see Appendix \ref{sec:appendixB}).
To assess robustness to variation, we compare the consistency of the different characterization metrics, evaluated on models with different architectures/parameterizations. All models are trained to convergence, with early stopping on a validation set.

\begin{wrapfigure}{r}{0.25\textwidth}
\vspace{-1mm}
  \centering
  \subfigure[Data-IQ]{\includegraphics[width=0.23\textwidth]{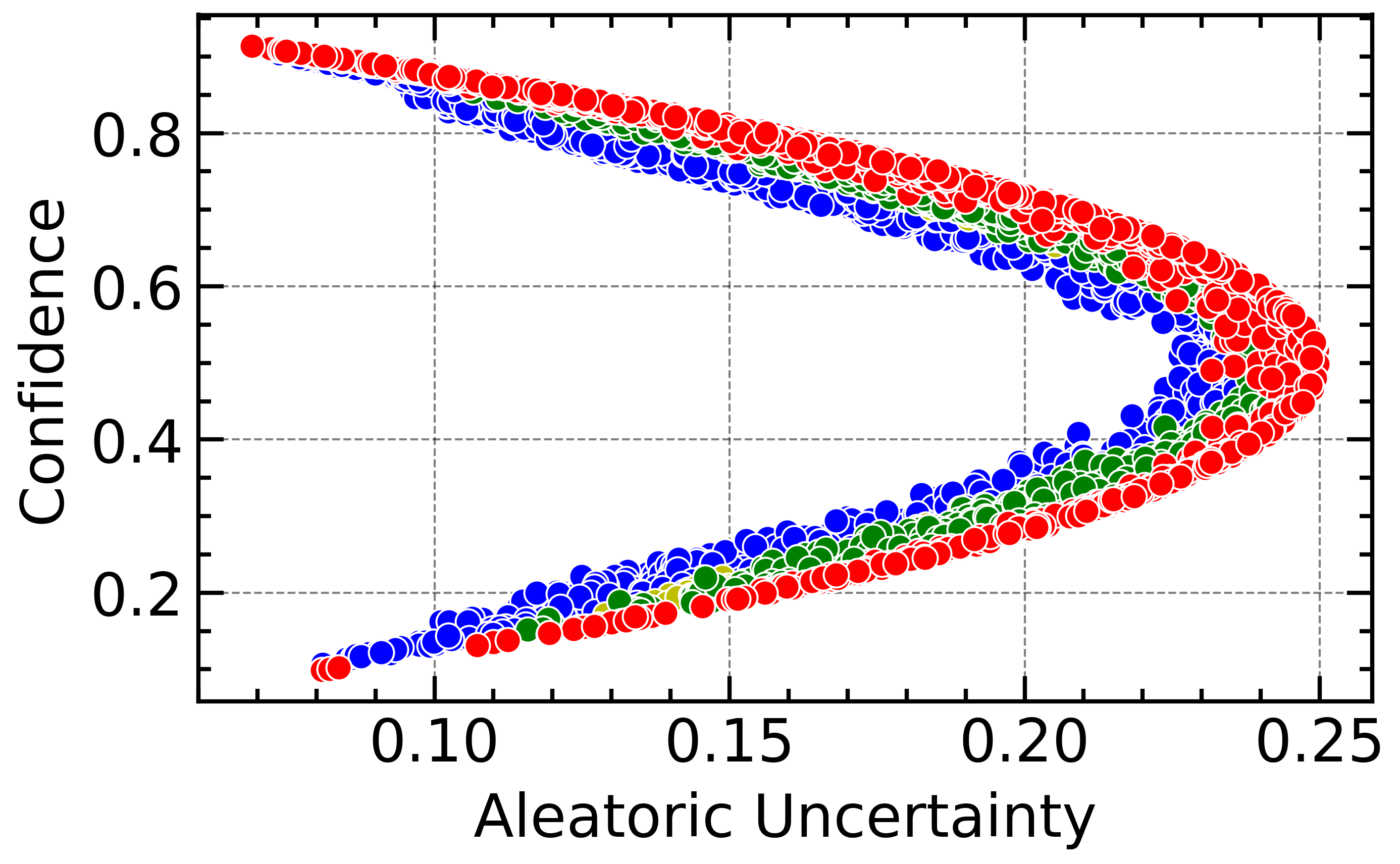}}\quad
  \subfigure[Data Maps \cite{swayamdipta2020dataset}]{\includegraphics[width=0.23\textwidth]{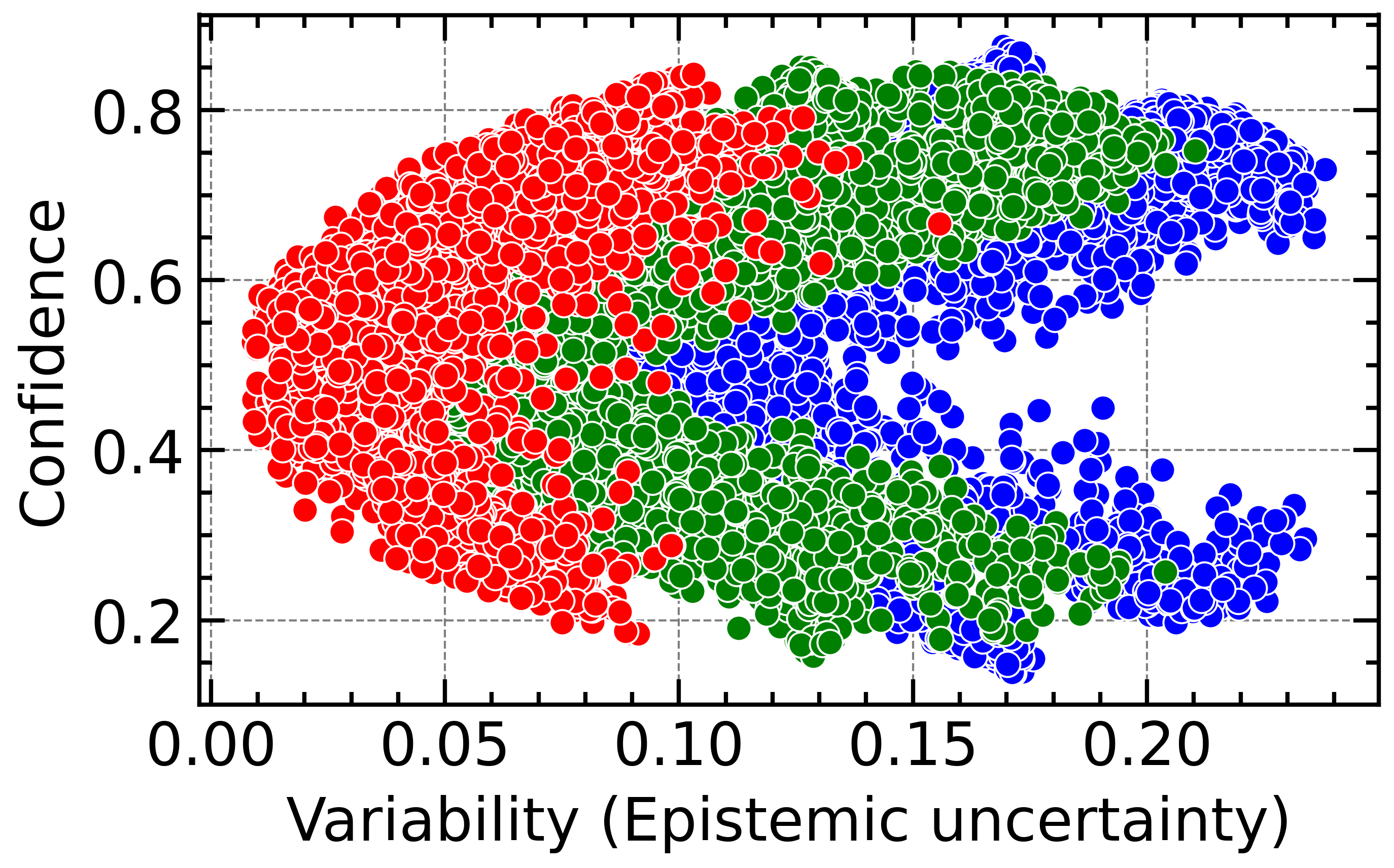}}\quad
  \vspace{-2.5mm}
  \caption{\footnotesize{Data-IQ's robustness to variation across diff. models (i.e.colors)}}
   \label{fig:compare}
   \vspace{-6mm}
\end{wrapfigure}

Quantitatively, we compute the Spearman rank correlation between all model combinations, see Fig.~ \ref{fig:comparison}. We observe that Data-IQ is the most consistent and robust to variation across different models, having the highest score on all datasets, satisfying \emph{P1}. Further, the baseline methods themselves are also not consistent in performance ordering across datasets, which is undesirable.
Ultimately, the robustness means practitioners can feel confident in the consistency of data insights, derived using Data-IQ.

To further compare Data-IQ and Data Maps\cite{swayamdipta2020dataset}, we examine 3 distinct models that achieve similar performance on the Covid-19 \cite{baqui2020ethnic} tabular dataset, and we produce a characterization of the training set using each model in Fig.~\ref{fig:compare}. We note that Data Maps groups can be recovered from \eqref{eq:groups_training} by replacing the aleatoric uncertainty $\val$ from Data-IQ with its epistemic counterpart $\vep$.  The y-axis is the same for both methods and corresponds to $\Bar{\P}(x)$. The x-axis corresponds to $\val(x)$ for Data-IQ and to $\vep(x)$ for Data Maps. Each model is assigned a color in Fig.~\ref{fig:compare}. We note three things \textcircled{1} Data-IQ's characterization of the data is significantly more stable across models.  \textcircled{2} Linked to the points in Sec.\ref{sec:formulation},  Data Map's high and low confidence examples in fact have high epistemic uncertainty $\vep$, which can lead to incorrect conclusions when attempting to use Data Maps to characterize data. \textcircled{3} Data-IQ always distributes the data around a bell shape, which standardizes its interpretation. We provide a theoretical analysis to explain this bell shape observation in Appendix \ref{sec:appendixA}.  

\noindent\textbf{Data-IQ: Neural Networks vs Other model classes. } Data-IQ can be used with \emph{any} ML model trained in stages linked to \emph{P4}: Plug and Play. Methods such XGBoost, LightGBM and CatBoost methods are widely used by practitioners on tabular data, often more so than neural networks \cite{borisov2021deep}. Ideally, based on \emph{P1}, we desire that the characterization of examples be consistent for similar performing models, irrespective of whether the model is a neural network or an XGBoost model.

\begin{wrapfigure}{r}{0.66\textwidth}
    \vspace{-5mm}
  \centering 
  \subfigure[Data-IQ ]{\includegraphics[width=0.3\textwidth]{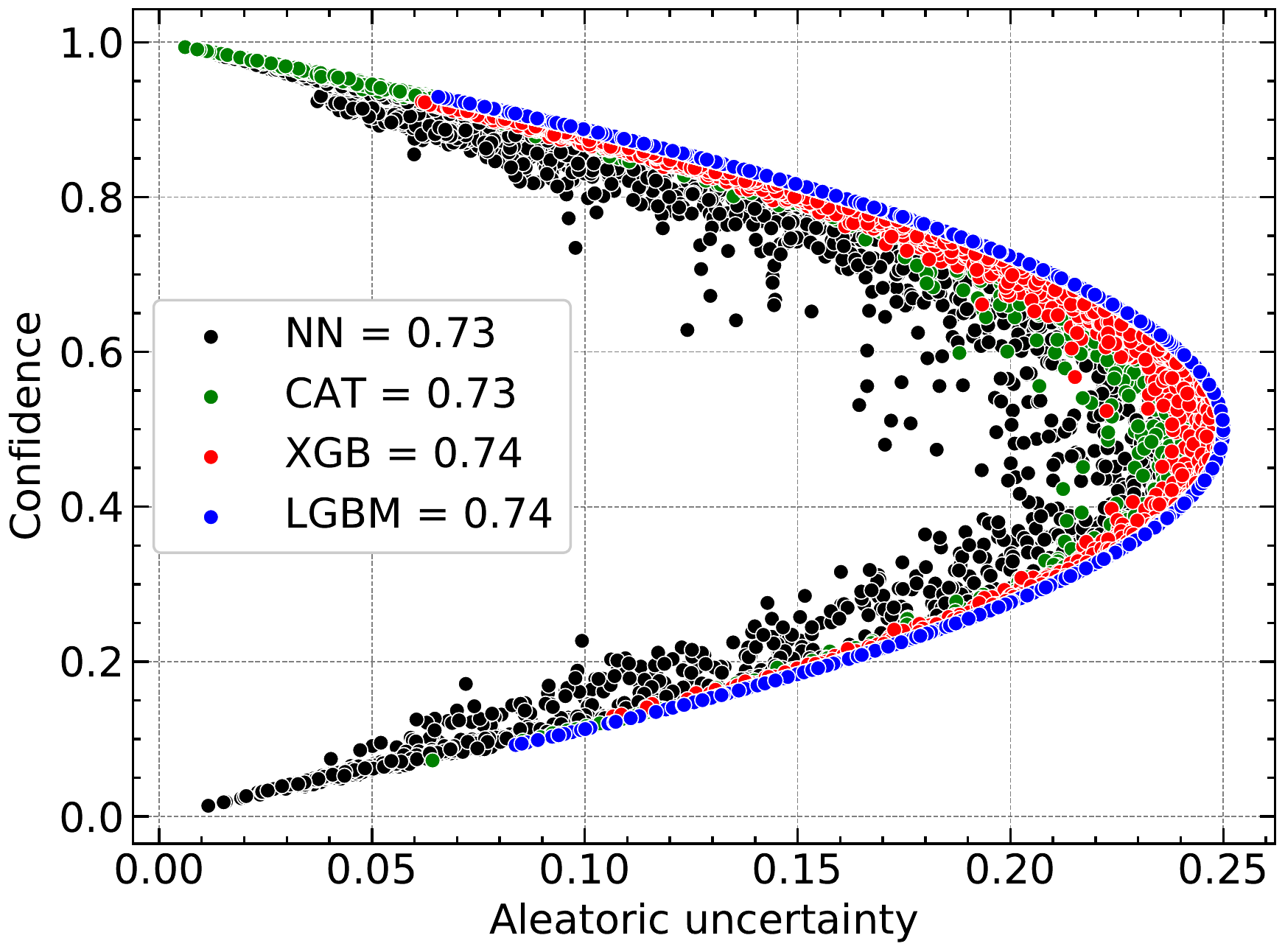}}\quad\quad
  \subfigure[Data Maps]{\includegraphics[width=0.3\textwidth]{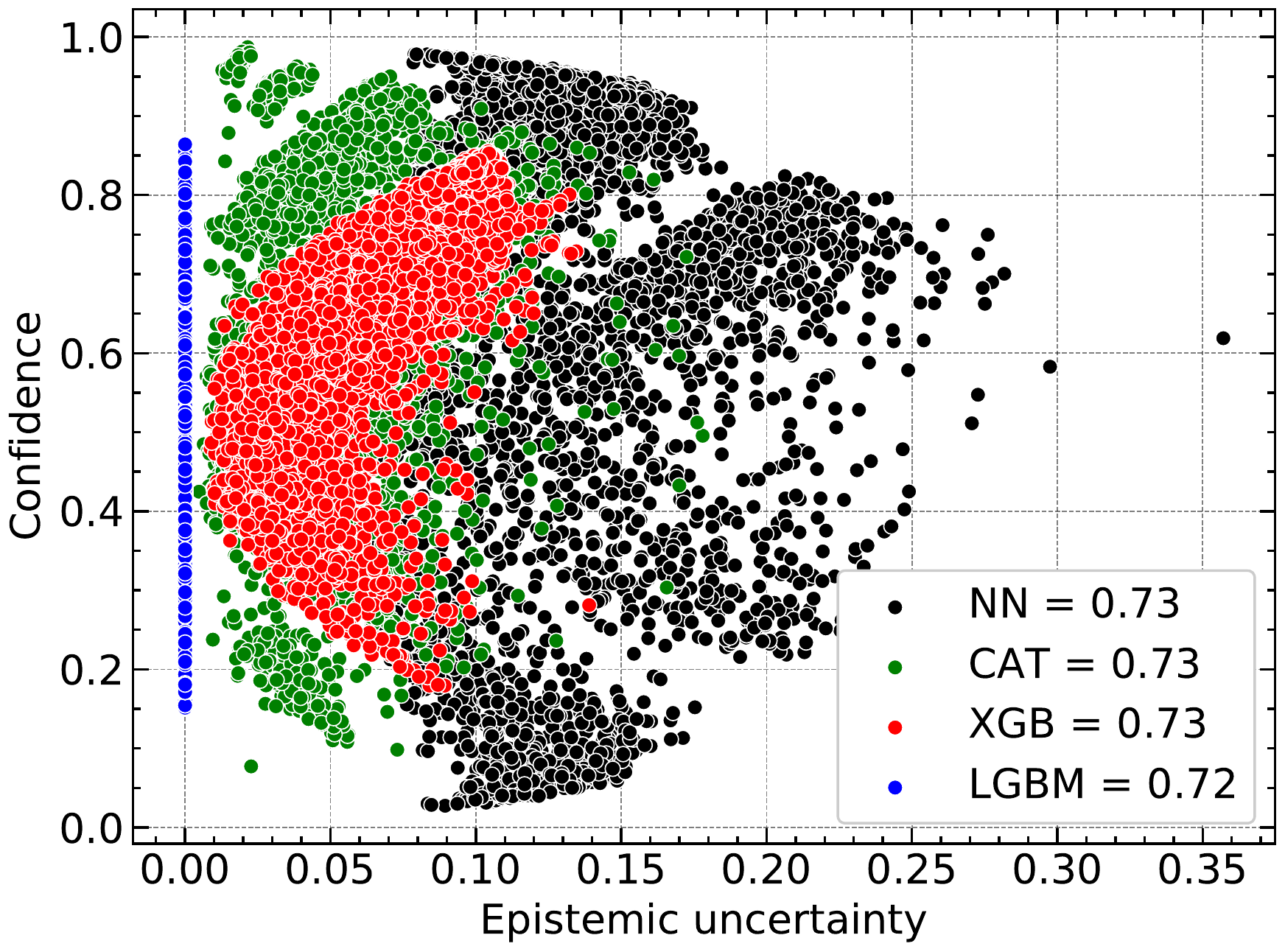}}
  \caption{NN vs XGBoost: Data-IQ is more consistent}
  \label{fig:model-compare}
  \vspace{-5mm}
\end{wrapfigure}

To assess the robustness of  both Data-IQ and Data Maps, we train a neural network, XGBoost, LightGBM and CatBoost models to achieve the same performance and then perform the characterization for all models. We can clearly see in Fig.~\ref{fig:model-compare} (Support) that Data-IQ has a similar characterization across all four models. Contrastingly, for Data Maps, the characterizations are significantly different for the different model classes. The implication of this result is that by Data-IQ capturing the uncertainty inherent to the data (aleatoric uncertainty), it leads to a more consistent and stable characterizations of the data itself. 
Especially, this highlights that Data-IQ characterizes the data in a manner that is not as sensitive to the choice of model when compared to Data Maps.  For more, see Appendix \ref{sec:appendixC}.

\noindent\textbf{Data insights from subgroups.}\label{insights}
Given the distinct differences between the subgroups, we seek to understand what factors make these subgroups different and how they can provide insight into the dataset. Such insights are especially useful in clinical settings.
Results for the prostate cancer dataset are illustrated in Fig. \ref{fig:insights} (with other datasets in Appendix \ref{sec:appendixC}). To visualize the different groups of patients within each subgroup, we cluster each subgroup ($\easy$, $\ambiguous$ and $\hard$) using a Gaussian Mixture Model (GMM) similar to \cite{sohoni2020no}, selecting the optimal clusters based on the Silhouette score. We assess cluster quality vs alternatives in Appendix \ref{sec:appendixC}.

In general, across datasets, the subgroups are: (1) $\easy$: Severe patients with a death outcome, and less severe patients with a survival outcome. (2) $\ambiguous$: Patients with similar features, but different outcomes. This could suggest that the features, we have at hand are insufficient to separate the differences in outcomes. (3) $\hard$: Severe patients with a survival outcome, and less severe patients with a death outcome. i.e. opposite outcomes as expected due to randomness in the outcomes.

\begin{figure}[!h]
    \vspace{-9mm}
    \centering
    \includegraphics[width=0.75\textwidth]{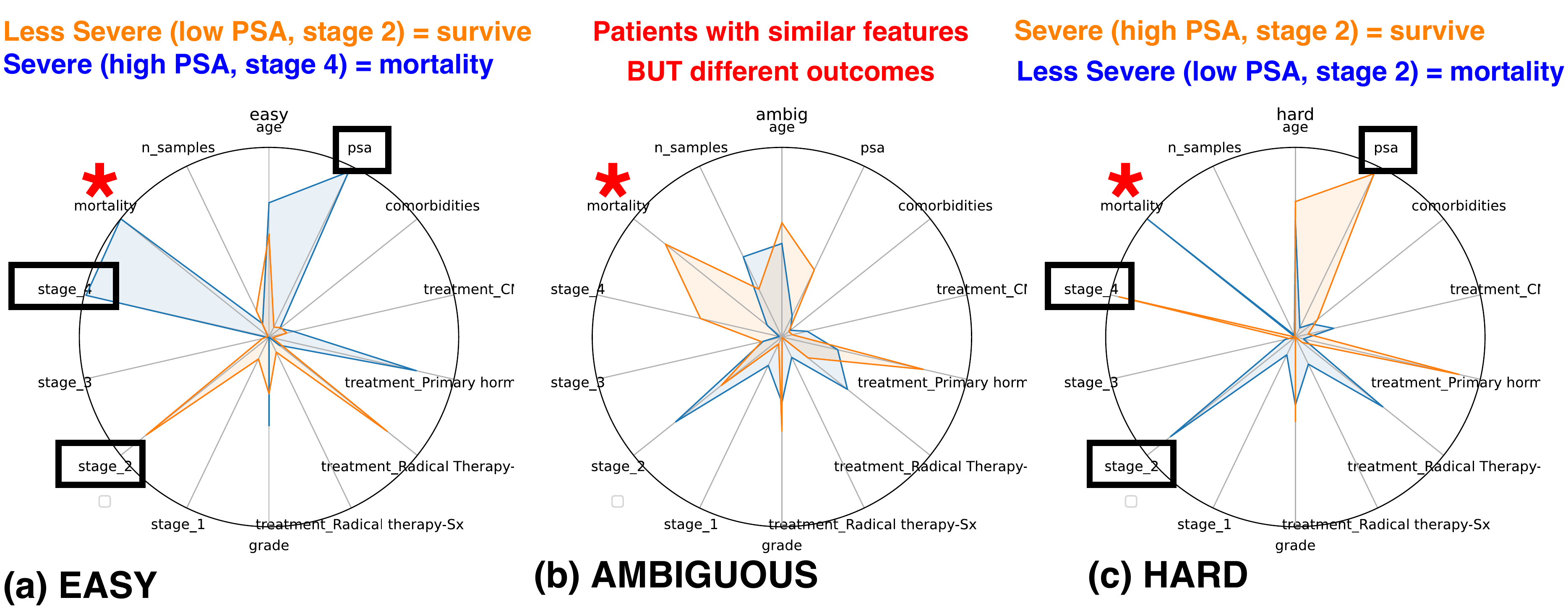}
    \vspace{-2mm}
    \caption{\footnotesize{Comparing subgroups identified by Data-IQ (descriptions above). Colors represent the GMM clusters.}}
    \label{fig:insights}
     \vspace{-3mm}
    \rule{\linewidth}{.75pt}
    \vspace{-9mm}
\end{figure}

\subsection{(\textbf{P2}) Principled data collection}\label{p2-exp}

\paragraph{Principled feature acquisition.}

\begin{wrapfigure}{r}{0.6\textwidth}
\vspace{-15mm}
  \centering
  \subfigure[
\scriptsize{Data-IQ subgroup ambiguity proportion is reduced as more informative features are acquired.}]{\includegraphics[width=0.28\textwidth]{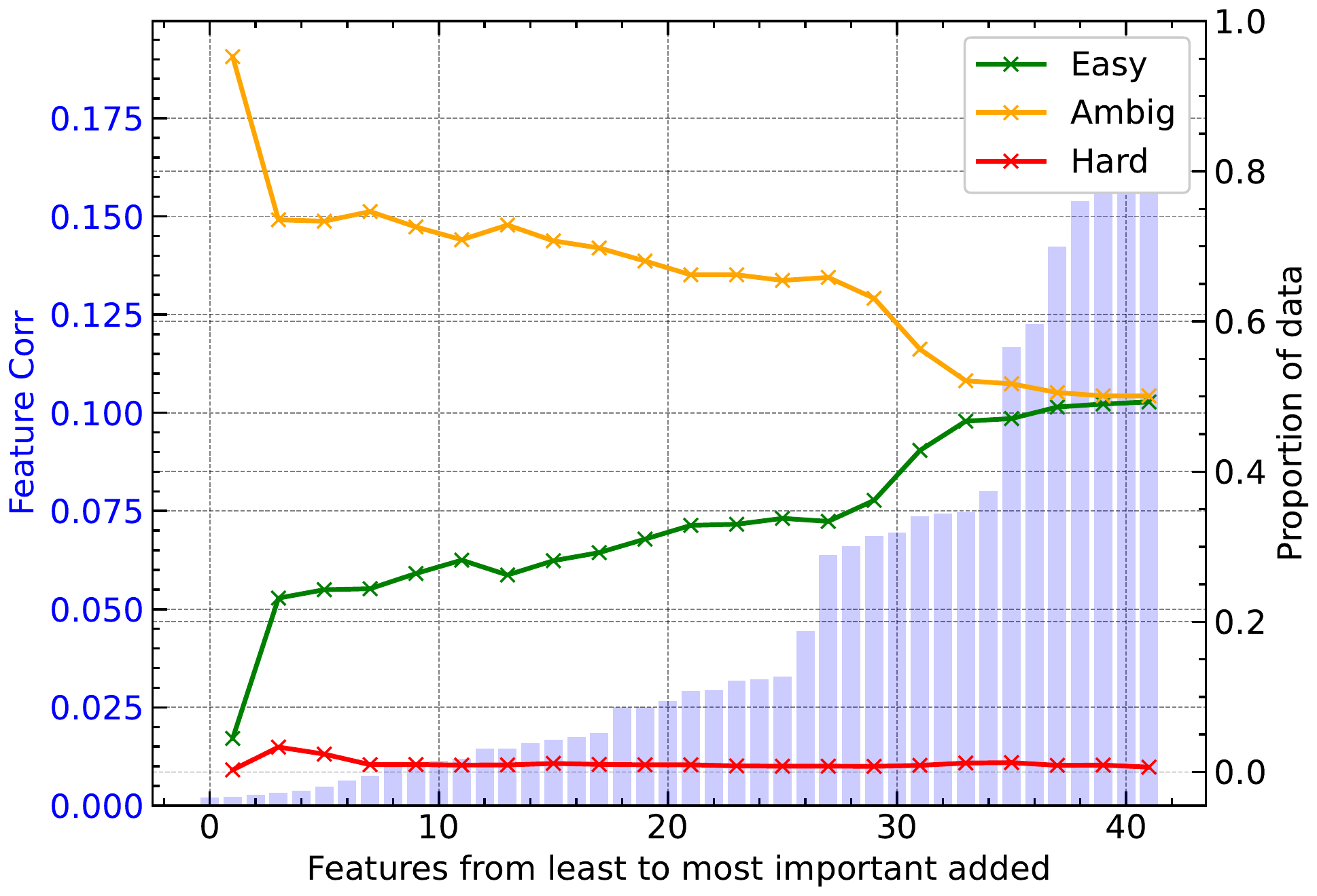}}\quad
  \subfigure[\scriptsize{Data-IQ aleatoric uncertainty remains stable for Ambiguous, reduces for others as features are acquired.} ]{\includegraphics[width=0.28\textwidth]{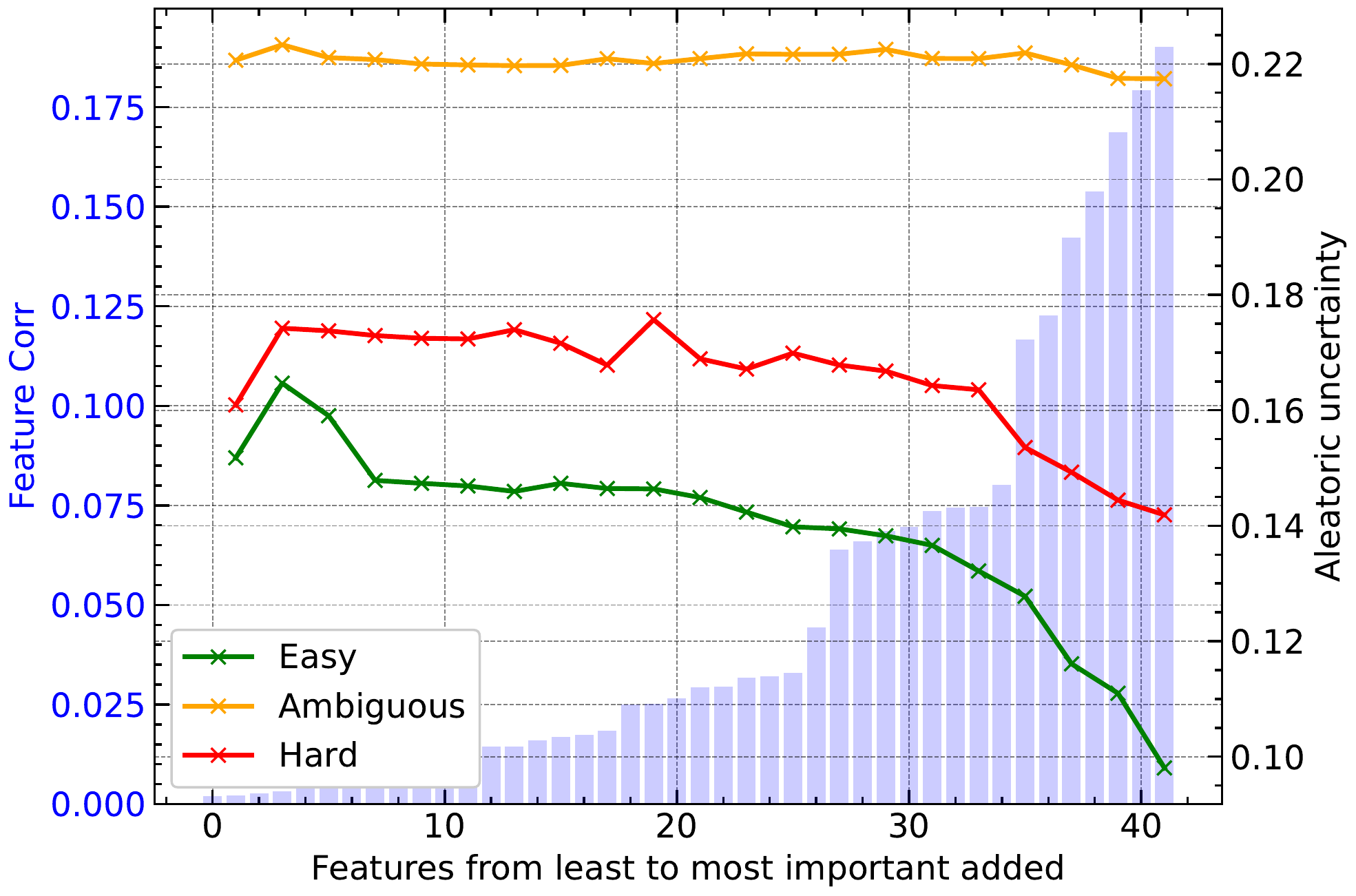}}\quad\\
  \subfigure[\scriptsize{Data Maps subgroup proportions largely unaffected as features acquired.}]{\includegraphics[width=0.28\textwidth]{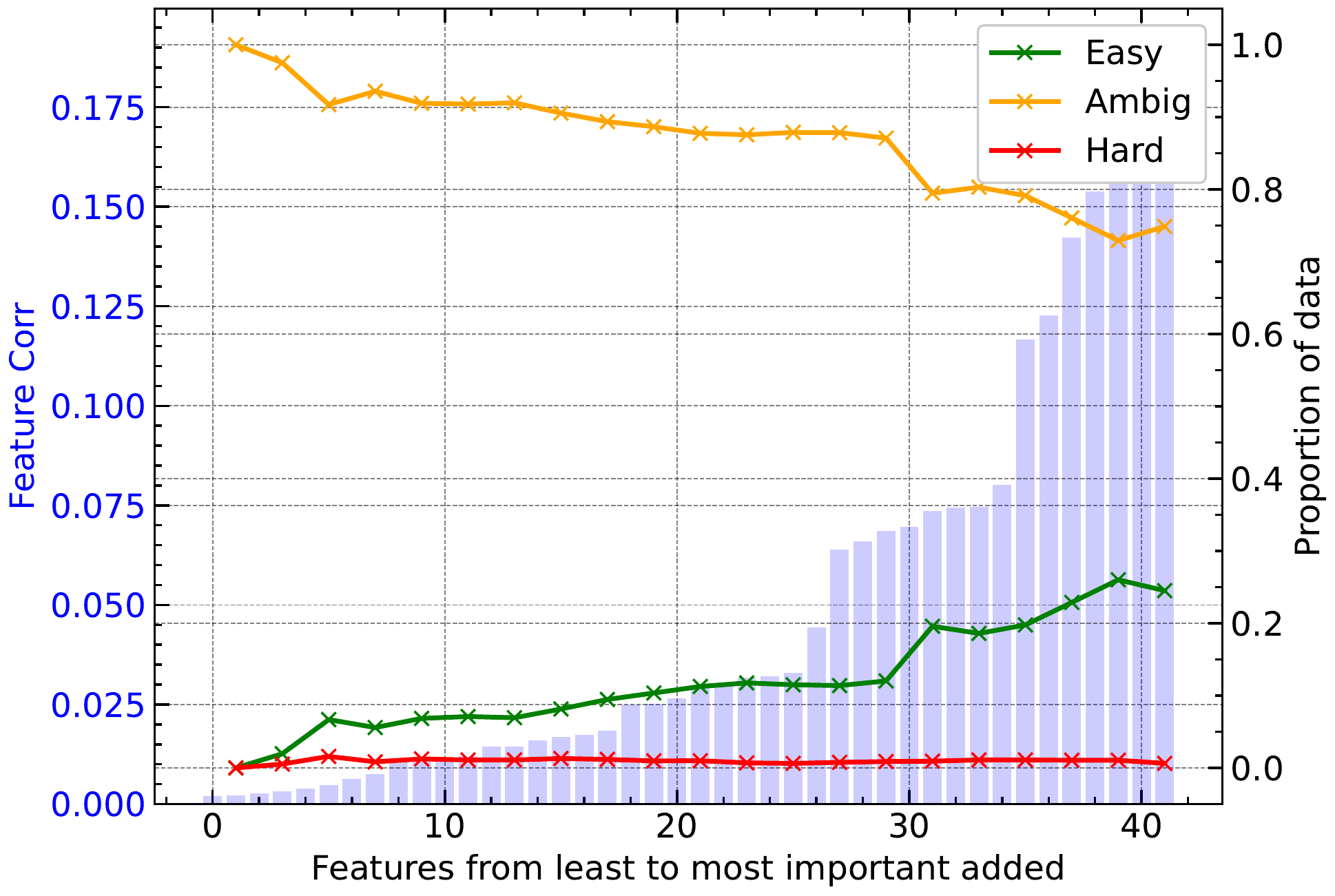}}\quad
  \subfigure[\scriptsize{Data Maps variability increases across subgroups as features acquired.} ]{\includegraphics[width=0.28\textwidth]{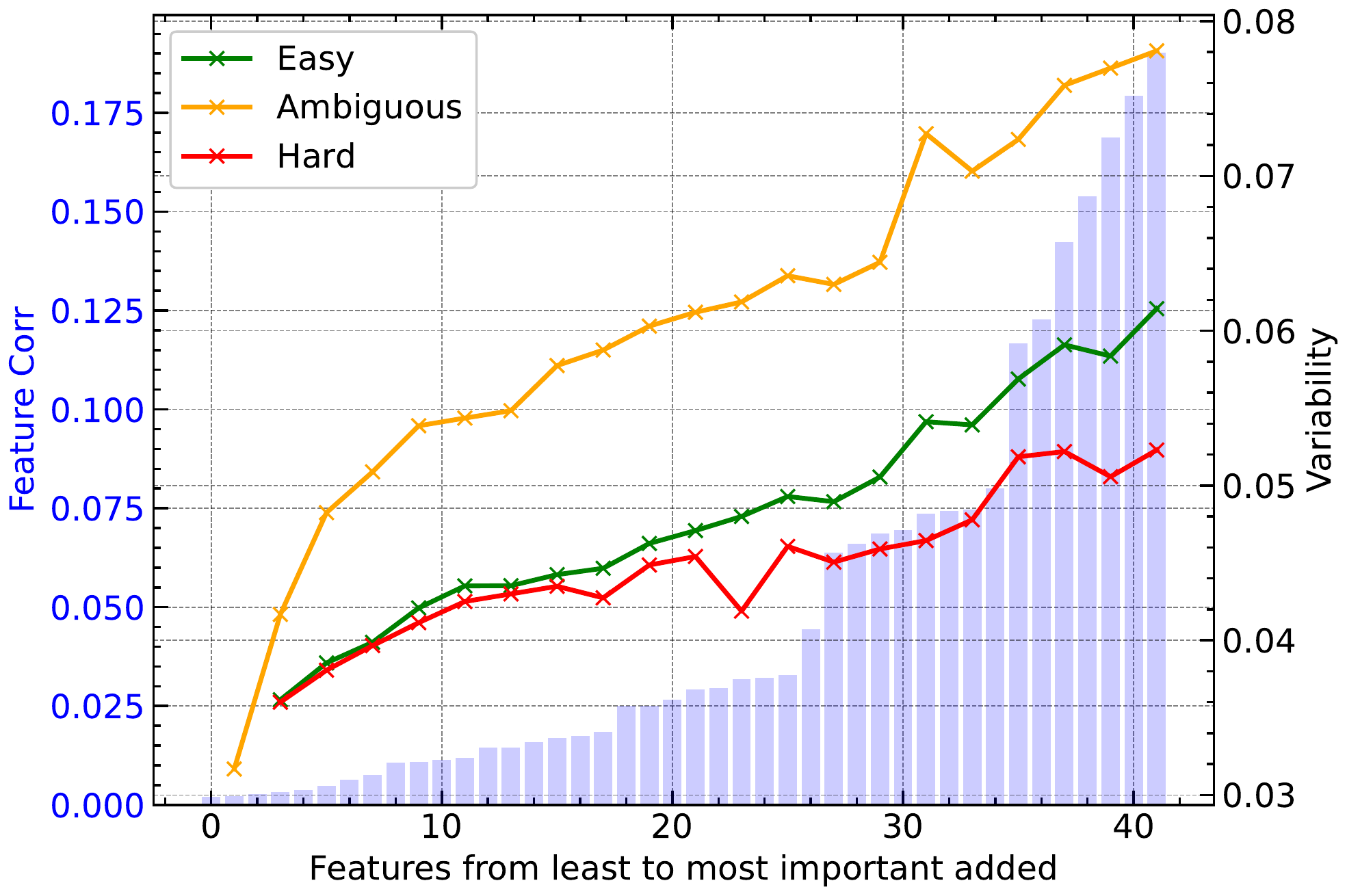}}\quad
  \vspace{-3mm}
  \caption{Quantifying the value of feature acquisition based on change in ambiguity. Only Data-IQ captures this.}
   \label{fig:support_feats}
 \vspace{-3mm}
    \rule{\linewidth}{.45pt}
\vspace{-8mm}
\end{wrapfigure}

As per Sec.\ref{insights}, the $\ambiguous$ subgroup has examples with similar features, yet different outcomes. Recall that this case of ambiguity in the tabular setting is very different from ambiguity in other modalities, such as images. The ambiguity is due to insufficient features to adequately separate the examples.  We link this to the concept that the $\ambiguous$ subgroup has a high aleatoric uncertainty that is irreducible, even if we collect more data examples. Rather, aleatoric uncertainty can only be reduced by acquiring better features \cite{gavves}. 
We leverage this idea and show that Data-IQ's example characterization provides a principled approach to assessing the benefit of acquiring a specific feature. This is different from feature selection, where all features are present and we select the most ``important feature''. Additionally, this is different from active learning which quantifies the value of acquiring examples, not features. 

With the above in mind, a valuable feature should decrease the example ambiguity (i.e. aleatoric uncertainty). Hence, a decrease in the proportion of $\ambiguous$ examples can serve as a proxy for the feature's potential value to the dataset. Understanding the value of features is useful in settings such as healthcare, where feature acquisition comes at a cost.  To showcase the potential, we construct a semi-synthetic experiment, where we rank sort the features based on correlation with the target. 

We then train different models, where we sequentially ``acquire'' features of increasing value (based on correlation). Fig. \ref{fig:support_feats}. shows results for the Support dataset. 
For Data-IQ, Fig \ref{fig:support_feats} (a) shows that as we acquire ``valuable'' features, the proportion of the $\ambiguous$ subgroup drops, whilst the $\easy$ subgroup increases, with significant changes for the important features. This shows that Data-IQ's subgroup characterization can be used to quantify a feature's value, by its ability to decrease ambiguity. In contrast, Data Maps, Fig \ref{fig:support_feats} (c), shows minimal response to feature acquisition, suggesting it may not be sensitive enough to capture the feature's value.

Further, for Data-IQ we see that the examples that remain as $\ambiguous$, after features are collected, maintain a consistent aleatoric uncertainty. This is desired as it demonstrates for those examples which remain $\ambiguous$, that indeed the features collected are not informative enough to reduce their inherent (aleatoric) uncertainty, i.e. those remaining still need better features  (see Fig \ref{fig:support_feats} (b)). While for Data Maps the added features, in fact increase the variability for \emph{all} subgroups making it harder to stratify (see Fig \ref{fig:support_feats} (d)). This links to the fact that Data Maps subgroups can't capture the value of the acquired features. We further show experimentally in Appendix \ref{sec:appendixC} that for $\ambiguous$ examples, it is not simply a case of increasing the size of the dataset (i.e. more examples). In fact, this can increase the proportion of $\ambiguous$ examples due to the increased probability of feature collisions as the dataset size increases. Ultimately, this motivates the usefulness of principled feature acquisition (which Data-IQ can guide), as a way to decrease dataset ambiguity.

\paragraph{Principled dataset comparison.}
Extending beyond feature-level, an understudied scenario involves systematically selecting between datasets in two cases: (1) purchasing data from data markets \cite{koutroumpis2020markets,raskar2019data,rasouli2021data} and (2) organizations where the data is siloed, with lengthy access processes \cite{el2020practical, goncalves2020generation}. In both cases, synthetic versions of the real dataset has begun to be used \cite{el2020practical}. For now, we ignore privacy concerns, and focus on data fidelity and quality, which compares the real and synthetic datasets using statistical measures. However, as per \cite{alaa2022how}, the conclusions can vary across different metrics. In practice, competing ``synthetic'' datasets can be generated by different ML models or vendors. Thus, while they model the same underlying distribution, depending on the process used, one version might be superior. We now ask whether Data-IQ could permit us to systematically select between synthetic datasets? We consider the setting where the real data is \emph{not accessible}. Hence, we can't use existing evaluation metrics, yet still wish to compare the synthetic datasets (e.g. comparing vendors).

\begin{wraptable}{r}{6.5cm}
\vspace{-3mm}
 \centering
\centering
\captionsetup{font=footnotesize}
\caption{Comparison of accuracy performance rank and  (\textcolor{ForestGreen}{dataset quality}). Synthetic dataset w/ better quality ($\uparrow$  easy)produces the best real data test performance. }
\scalebox{0.75}{
\begin{tabular}{ccc}
\toprule
Dataset &  (V1) CTGAN & (V2) Gaussian Copula    \\ \hline
\midrule
Prostate & \textbf{Rank 1 (\textcolor{ForestGreen}{63\% Easy})}  & Rank 2 (30\% Easy) \\ \hline
Covid & \textbf{Rank 1 (\textcolor{ForestGreen}{70\% Easy})}  & Rank 2 (63\% Easy)\\ \hline
Support & \textbf{Rank 1 (\textcolor{ForestGreen}{59\% Easy})}  & Rank 2 (38\% Easy)\\ \hline
Fetal & Rank 2 (40\% Easy)  & \textbf{Rank 1 (\textcolor{ForestGreen}{51\% Easy})}
\\ \hline
\bottomrule
\end{tabular}}
\label{tab:datasets_quality}
\vspace{-3mm}
\end{wraptable}

We simulate this scenario by generating synthetic data using 2 different models, representing 2 synthetic data vendors: (V1) CTGAN \cite{xu2019modeling} and (V2) Gaussian Copula \cite{li2020sync}. We then characterize the dataset subgroup proportions using Data-IQ. We hypothesize that datasets with greater proportions of $\easy$ examples generalize better. As is common, we validate fidelity by training with synthetic data and testing with real data \cite{esteban2017real,platzer2021holdout}, where the best fidelity data produces the best model performance on real data (test set).  Table \ref{tab:datasets_quality} shows that datasets with the highest quality as measured by Data-IQ indeed produces the best performance on real test data (i.e. Rank 1). Further, it shows that the same ``vendor'' does not always produce the best dataset, highlighting the value of comparative assessment. Ultimately, these two aspects demonstrate that when the real data is unavailable, Data-IQ is a useful tool in the hands of practitioners wishing to assess data quality, especially when selecting between different datasets.

\subsection{(\textbf{P3}) Reliable Model Deployment}\label{p3-exp}

\paragraph{Less is more: data sculpting based on subgroups.}

\begin{wrapfigure}{r}{0.3\textwidth}
\captionsetup{font=footnotesize}
\vspace{-12mm}
  \centering
    \includegraphics[width=0.3\textwidth]{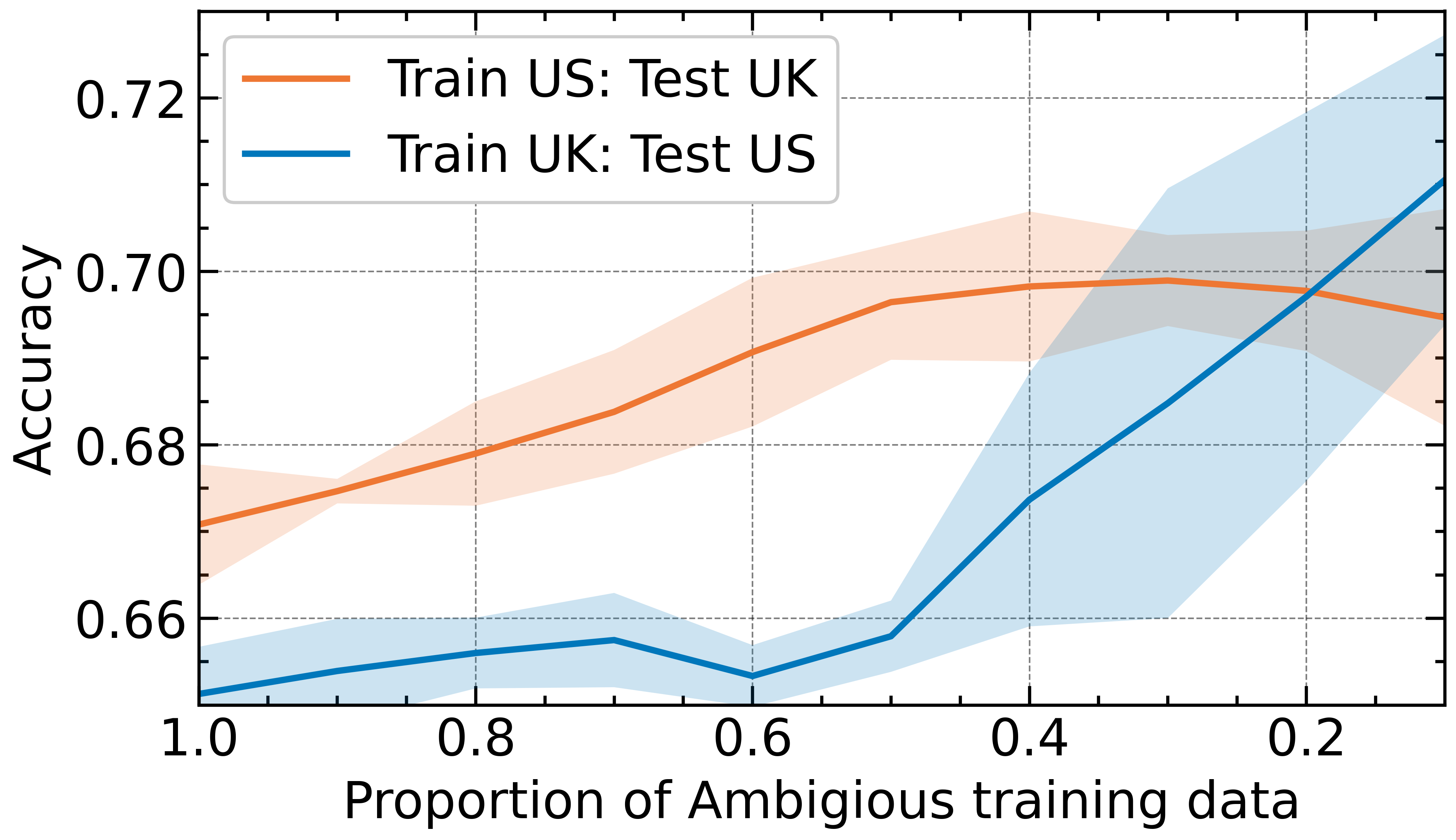}
   \vspace{-5mm}
   \caption{\footnotesize{Performance improvement via data sculpting based on subgroups}}
    \vspace{-3mm}
    \rule{\linewidth}{.25pt}
   \vspace{-6mm}
    \label{fig:sculpt}
\end{wrapfigure}

What role do the data subgroups, specifically $\ambiguous$ examples, play in ensuring model generalization? Using multi-country prostate cancer data, we train a baseline model on US data (SEER) and assess generalization when deployed on patients in the UK (CUTRACT) and vice versa.  In Fig. \ref{fig:sculpt}, we see that test time generalization performance, monotonically increases as we decrease the proportion of $\ambiguous$ training data (see Appendix \ref{sculpt-nums} for absolute numbers). Ultimately, this illustrates the value of sculpting the training dataset, by removing ambiguous examples, as a way to improve the reliability of a deployed model.

\paragraph{Group-DRO: Not a silver bullet when used in tabular settings.}

\begin{wraptable}{r}{7cm}
\vspace{-1mm}
 \centering
 \captionsetup{font=footnotesize}
\caption{Comparison of different model improvement/robustness techniques}
\vspace{-0mm}
\scalebox{0.6}{
\begin{tabular}{ l|l|l|ccc }

\hline
Dataset & Group & Baseline & \makecell{Group-DRO\\ (\textbf{Data-IQ})}  & \makecell{Group-DRO \\(George)} & JTT \\ \hline\hline
\multirow{3}{*}{Prostate} & Overall & 0.837 & \textbf{0.840} \textcolor{ForestGreen}{$\uparrow$} & 0.565 \textcolor{red}{$\downarrow$} & 0.723  \textcolor{red}{$\downarrow$} \\
 & Ambiguous  & 0.740 & \textbf{0.741} \textcolor{ForestGreen}{$\uparrow$} & 0.598  \textcolor{red}{$\downarrow$}& 0.538 \textcolor{red}{$\downarrow$} \\ 
  & The Rest & \textbf{0.935} & \textbf{0.935} & 0.535 \textcolor{red}{$\downarrow$} & 0.896 \textcolor{red}{$\downarrow$} \\ \hline\hline
 
 \multirow{3}{*}{Covid} & Overall & 0.729 & \textbf{0.732} \textcolor{ForestGreen}{$\uparrow$} & 0.687 \textcolor{red}{$\downarrow$} & 0.455 \textcolor{red}{$\downarrow$} \\
 & Ambiguous  & 0.629 & \textbf{0.633} \textcolor{ForestGreen}{$\uparrow$}  & 0.609 \textcolor{red}{$\downarrow$} & 0.477 \textcolor{red}{$\downarrow$}  \\ 
   & The Rest  & \textbf{0.832} & \textbf{0.832} & 0.766 \textcolor{red}{$\downarrow$} & 0.438 \textcolor{red}{$\downarrow$} \\ \hline\hline
 
  \multirow{3}{*}{Support} & Overall & \textbf{0.734} &  \textbf{0.734} & 0.660 \textcolor{red}{$\downarrow$} & 0.611 \textcolor{red}{$\downarrow$} \\
 
 & Ambiguous  & 0.621 & \textbf{0.622} \textcolor{ForestGreen}{$\uparrow$} & 0.576 \textcolor{red}{$\downarrow$} &  0.521 \textcolor{red}{$\downarrow$} \\
  & The Rest & \textbf{0.858} & \textbf{0.858} & 0.754 \textcolor{red}{$\downarrow$} & 0.711 \textcolor{red}{$\downarrow$} \\\hline\hline

  \multirow{3}{*}{Fetal} & Overall & 0.768 & \textbf{0.833} \textcolor{ForestGreen}{$\uparrow$} & 0.829 \textcolor{ForestGreen}{$\uparrow$} & 0.829 \textcolor{ForestGreen}{$\uparrow$}  \\
 & Ambiguous  & 0.575 & \textbf{0.701} \textcolor{ForestGreen}{$\uparrow$} & 0.695 \textcolor{ForestGreen}{$\uparrow$} & 0.698 \textcolor{ForestGreen}{$\uparrow$} \\ 
  & The Rest & 0.970 & \textbf{0.974} \textcolor{ForestGreen}{$\uparrow$} & 0.970 & 0.970 \\\hline\hline
\end{tabular}}
\label{robustness}
\vspace{-4mm}
\end{wraptable}

Once subgroups of underperformance are identified, it is generally assumed that methods such as Group Distributionally Robust Optimization (DRO) \cite{sagawa2019distributionally} can be applied to improve model performance and robustness. We compare Group-DRO with groups identified by Data-IQ, and as baselines: George \cite{sohoni2020no} and Just-Train-Twice (JTT) \cite{liu2021just}.  As per the previous experiments, the largest underperforming group (in proportion) is the $\ambiguous$ examples.   Similar to the literature, we evaluate the performance change from a baseline model, after Group-DRO is applied.

The results in Table \ref{robustness} show that Group-DRO using Data-IQ's groups both improves overall performance and improves performance on the $\ambiguous$ group. Whilst, for the other baselines, the performance actually degrades.

Nevertheless, while using Data-IQ can boost performance, it is evident that simply applying Group-DRO is not a silver bullet to equalize subgroup performance, given the sometimes small improvement. The rationale is our tabular setting is different, from the spurious correlation setting in computer vision where Group-DRO typically shines. Ultimately, we believe, based on the feature acquisition results, that in tabular settings, practitioners would be well served acquiring better features to improve performance and reduce ambiguity.

\paragraph{Subgroup-informed usage of uncertainty estimation.}
Uncertainty estimation methods are essential in safety-critical areas such as healthcare \cite{seedat2020mcu}, yet are typically assessed on average. We ask the question: since subgroups have different performance properties, are uncertainty estimates equally reliable for each subgroup? As done in the literature \cite{leibig2017leveraging, xia2021uncertainty}, if an uncertainty estimate is reliable and informative of predictive performance, it can be used to defer ``uncertain examples''. This is done by rank sorting examples based on uncertainty and thresholding proportions of tolerated uncertainty \cite{leibig2017leveraging,xia2021uncertainty,krishnan2020improving,mukhoti2018evaluating}. Ideally, as the threshold proportion of examples increases (i.e. inclusion of more uncertain examples), we should see a monotonic decrease of accuracy. We assess this by training a Bayesian Neural Network (BNN)\cite{ghosh2018structured} to obtain uncertainty estimates. We then compute the performance across different threshold proportions $\tau \in \{0.1,0.2, \dots ,1\}$ for the $\ambiguous$ subgroup specifically, and as is commonly done, across the entire dataset (i.e. \emph{average}).

\begin{wrapfigure}{r}{0.3\textwidth}
\vspace{-10mm}
  \centering
    \includegraphics[width=0.3\textwidth]{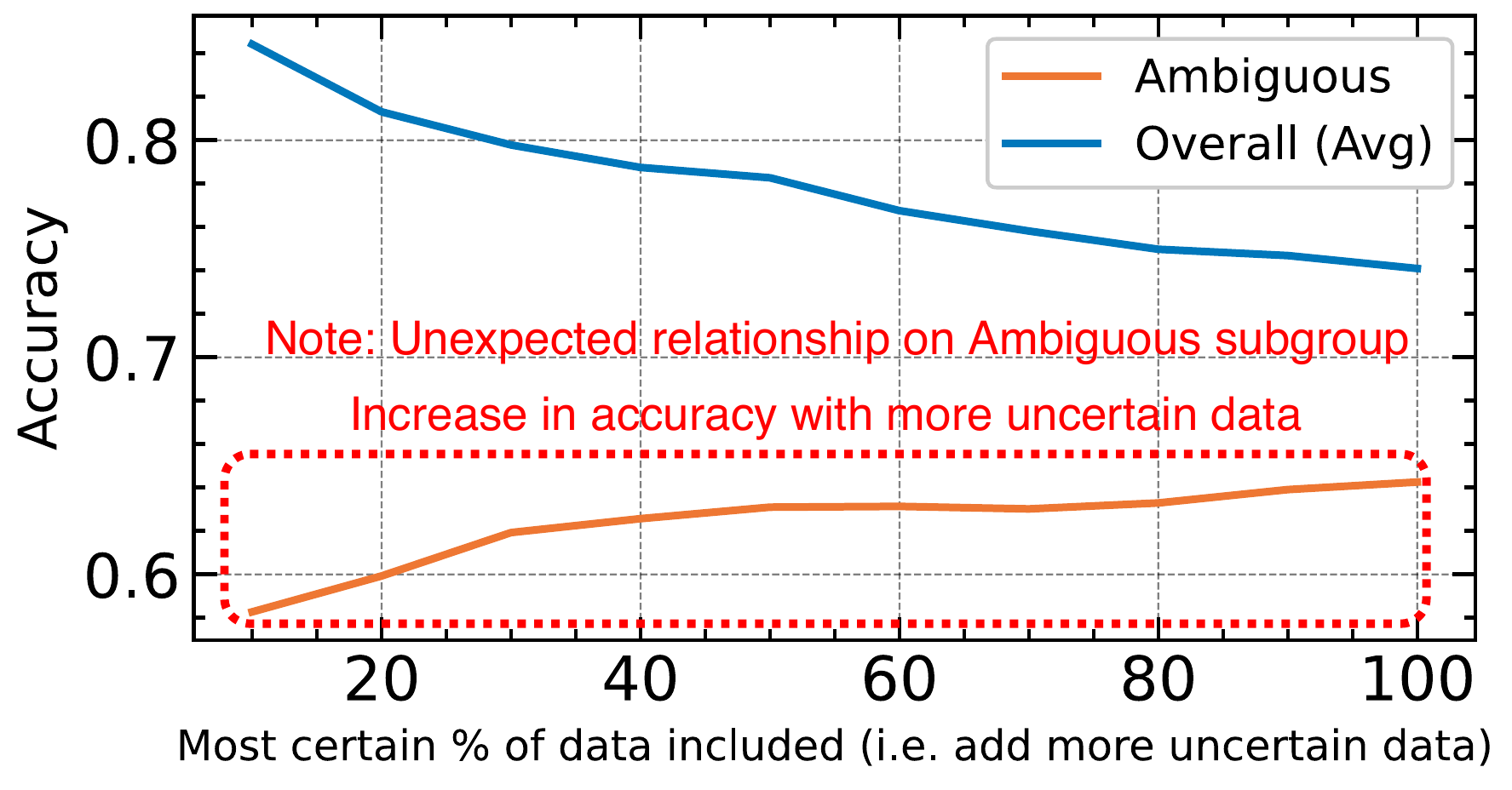}
   \vspace{-5mm}
   \caption{\footnotesize{Data-IQ subgroups can unmask unreliable prediction deferral by uncertainty methods}}
    \vspace{-3mm}
    \rule{\linewidth}{.4pt}
   \vspace{-7mm}
    \label{fig:uncert}
\end{wrapfigure}

Fig.~\ref{fig:uncert} shows a \emph{specific} example, wherein \emph{average} examples exhibit the monotonic decreasing relationship, as expected. However, the $\ambiguous$ examples categorized by Data-IQ , contrary to expectations, show an increase in accuracy as more uncertain examples are included. This suggests that uncertainty estimates in this case are not as informative of predictive performance for the $\ambiguous$ examples. This shows the potential for practitioners to use Data-IQ at deployment time to understand which examples require auditing before deferring, as the ``average'' monotonic decreasing behavior, based on the uncertainty estimates, may not always hold. Ultimately, the result further highlights how subgroup characterization via Data-IQ could assist practitioners in unmasking unreliable performance, not evident on average.

\section{Discussion} \label{sec:discussion}
In this paper, we introduce Data-IQ, a systematic framework that can be used with \emph{any} ML model with checkpoints, to characterize examples into subgroups with respect to the outcome. Through several experiments, we demonstrate that the usage of aleatoric uncertainty, which captures properties more inherent to the data, is indeed more principled, being more robust to variation across models and/or parameterizations. Data-IQ's consistency is unmatched by any compared baseline.  Data-IQ should not automate and replace the intuition of a data scientist. Rather, as we have demonstrated, Data-IQ should serve as a systematic ``data-centric ML'' tool that assists and empowers data scientists with the ``data'' work at training time, whilst also guiding reliable model usage at deployment time.

\textbf{Data-IQ beyond tabular settings.} The main paper has primarily assessed the utility of Data-IQ in the tabular setting. That said, in Appendices \ref{nlp} and \ref{cv}, we evaluate the utility of Data-IQ on text data (\emph{NLP}) and images (\emph{computer vision}) respectively.

\textbf{Limitations and future opportunities.} \textcircled{1} While Data-IQ characterizes examples; the current formulation does not allow us to understand which attributes are responsible for the characterization per example. This would be an interesting extension around dataset explainability, allowing practitioners to better probe their data. \textcircled{2} In high-stakes settings such as healthcare, to mitigate possible adverse effects (e.g. difficulty of $\easy$ vs $\hard$), Data-IQ should be used with a ``human-in-the-loop'', allowing experts to complement and validate findings with domain knowledge.

\section*{Acknowledgments}
The authors are grateful to Zhaozhi Qian, Yuchao Qin, Evgeny Saveliev and
the anonymous NeurIPS reviewers for their useful comments
\& feedback.  Nabeel Seedat is supported by the Cystic Fibrosis Trust, Jonathan
Crabbe by Aviva, Ioana Bica by the Alan Turing Institute, EPSRC grant EP/N510129/1  and Mihaela van der Schaar by the Office of Naval Research (ONR), NSF 1722516. 

\newpage

\bibliographystyle{unsrt}
\bibliography{refs}

\newpage
\section*{Checklist}

\begin{enumerate}

\item For all authors...
\begin{enumerate}
  \item Do the main claims made in the abstract and introduction accurately reflect the paper's contributions and scope?
    \answerYes{Please refer to our “contributions” paragraph in Section 1. Furthermore, all claims made about Data-IQ are verified in Section 4 and in the supplementary material.}
  \item Did you describe the limitations of your work?
    \answerYes{See Section 3.3 and Section 5.}
  \item Did you discuss any potential negative societal impacts of your work?
    \answerYes{Data-IQ provides practioners with a tool to help them with their ``data'' work and to better understand the characteristics of their data. In fact, Data-IQ may help reveal the negative impacts of ML models. However, as discussed in Section 5, expert knowledge should still be used to validate our characterization.}
  \item Have you read the ethics review guidelines and ensured that your paper conforms to them?
    \answerYes{We have carefully read the ethics review guideline and confirm that our paper respects the guidelines.}
\end{enumerate}

\item If you are including theoretical results...
\begin{enumerate}
  \item Did you state the full set of assumptions of all theoretical results?
    \answerYes{See Section 3 and Appendix A of the supplementary material}
        \item Did you include complete proofs of all theoretical results?
    \answerYes{See Section 3 and Appendix A of the supplementary material}
\end{enumerate}

\item If you ran experiments...
\begin{enumerate}
  \item Did you include the code, data, and instructions needed to reproduce the main experimental results (either in the supplemental material or as a URL)?
    \answerYes{See footnotes 2 and 3}
  \item Did you specify all the training details (e.g., data splits, hyperparameters, how they were chosen)?
    \answerYes{See Appendix B, detailing all relevant information}
        \item Did you report error bars (e.g., with respect to the random seed after running experiments multiple times)?
    \answerYes{Included as relevant to the experiments in Section 4 and Appendix C}
        \item Did you include the total amount of compute and the type of resources used (e.g., type of GPUs, internal cluster, or cloud provider)?
    \answerYes{See Appendix B}
\end{enumerate}

\item If you are using existing assets (e.g., code, data, models) or curating/releasing new assets...
\begin{enumerate}
  \item If your work uses existing assets, did you cite the creators?
    \answerYes{See citations \cite{baqui2020ethnic}, \cite{duggan2016surveillance}. \cite{prostate}, \cite{knaus1995support},\cite{ayres2000sisporto}. We cite these in the main paper and supplementary material}
  \item Did you mention the license of the assets?
    \answerYes{See Appendix B of the supplementary material.}
  \item Did you include any new assets either in the supplemental material or as a URL?
    \answerNA{We only use existing assets}
  \item Did you discuss whether and how consent was obtained from people whose data you're using/curating?
   \answerYes{It is discussed in detail in the publications by the dataset creators, which we have cited}
  \item Did you discuss whether the data you are using/curating contains personally identifiable information or offensive content?
   \answerYes{See Appendix B of the supplementary material.}
\end{enumerate}

\item If you used crowdsourcing or conducted research with human subjects...
\begin{enumerate}
  \item Did you include the full text of instructions given to participants and screenshots, if applicable?
    \answerNA{}
  \item Did you describe any potential participant risks, with links to Institutional Review Board (IRB) approvals, if applicable?
   \answerNA{}
  \item Did you include the estimated hourly wage paid to participants and the total amount spent on participant compensation?
    \answerNA{}
\end{enumerate}

\end{enumerate}

\clearpage
\appendix

\addcontentsline{toc}{section}{Appendix}
\part{Appendix: Data-IQ: Characterizing subgroups with
heterogeneous outcomes in tabular data}

\parttoc

\clearpage

\section{Data-IQ details \& related work} \label{sec:appendixA}
\subsection{Extended Related Work}\label{extended_related}
We present a comparison of our framework Data-IQ, and provide further contrast to related work. Table \ref{related_work}, highlights that the related methods do not satisfy all the properties (P1-P4). Furthermore even for cases where it is satisfied, the properties are not naturally satisfied; rather they require our framework to enable the use-case. \emph{P2} around data collection and selection, has been largely unaddressed by prior methods. In the case of \emph{P3}, the related methods focus on training time, without taking into account end-to-end deployment scenarios. Finally, these prior methods are typically applicable to only neural networks, which limits the broad applicability related to \emph{P4}, where in high-stakes settings such as healthcare and finance, which often have tabular data, practitioners typically use non-neural methods, such as Gradient Boosting or XGBoost.

\begin{table}[!h]
\centering
\caption{Comparison of related work, highlights that \emph{ONLY} Data-IQ naturally address all desiderata.\\ (\textcolor{ForestGreen}{$\heartsuit$}): satisfied naturally by construction, (\textcolor{Red}{$\diamondsuit$}): needs our framework}
\scalebox{0.7}{
\begin{tabular}{@{}l|c|cccc@{}}
\toprule
      & \makecell{Assessment \\ metric} 
      & \makecell{(P1) Data characterization \\ consistent across models}
      & \makecell{(P2) Principled data \\ collection \& selection} 
      & \makecell{(P3) Applicable \\at deployment time}  
      &  \makecell{(P4) Applicable to \emph{any}  \\ ML model} 
                        \\ \midrule
Data-IQ (Ours) & Aleatoric uncertainty & V. High & \cmark   & \cmark (\textcolor{ForestGreen}{$\heartsuit$})  & \cmark (\textcolor{ForestGreen}{$\heartsuit$})  \\
Data Maps \cite{swayamdipta2020dataset}  & Training variability   & Medium & \xmark   & \cmark (\textcolor{Red}{$\diamondsuit$})   & \cmark (\textcolor{Red}{$\diamondsuit$}) \\
AUM \cite{pleiss2020identifying} & Logit difference & High & \xmark   & \xmark  & \cmark  \\
GraNd \cite{paul2021deep}            & Gradient Norm  & Medium & \xmark  & \xmark  & \xmark   \\
JTT \cite{liu2021just}   & Training errors   & Low & \xmark   & \cmark (\textcolor{Red}{$\diamondsuit$})  & \cmark   \\
\bottomrule
\end{tabular}}
\label{related_work}
\end{table}

The properties of \emph{P1} were evaluated in Section \ref{robustness}. We delve into each subsequent property below and show that the related works have different goals and hence do not satisfy the properties:

\begin{itemize}
    \item \emph{P2}: Data Maps \cite{swayamdipta2020dataset} does not fulfill this property as was shown experimentally in Section \ref{p2-exp}. The method is shown to be insufficiently sensitive for the task. AUM \cite{pleiss2020identifying} uses the differences between logits to flag mislabelled instances. By virtue, AUM solves a different task of label errors, whereas the task of improving the data by collecting better features, etc., is focussed on the inputs rather than labels. GraNd \cite{paul2021deep} looks at the gradient norm to identify important examples. Their task, rather than feature acquisition, is to assist in the pruning of a dataset.  JTT \cite{liu2021just} looks at training errors to detect samples with spurious correlations. Not only is the task different, but flagged errors could also link to both ambiguity (needing more features) or hardness (example being difficult). Therefore, it is not sufficiently fine-grained to be leveraged for feature acquisition. In the best case, we might want to augment the dataset with such flagged examples. 
    \item \emph{P3}: Data Maps naturally operates only at training time (as per the original paper). However, if it leverages Data-IQ's representation space, then it could be extended to be used at test time. AUM operates to flag mislabelled data (i.e. outcome labels). These do not exist at test time, hence AUM is a training-time ONLY method. GraNd uses gradients to flag examples. Hence, it is only applicable at training time and not at test time, since we do not compute gradients at test time as we do not have access to labels. JTT surfaces training errors and hence we would not know test time errors as we do not have labels. However, if JTT leverages Data-IQ's representation space idea, then it could also be extended to be used at test time by flagging examples relative to a training time embedding.
    \item \emph{P4}: Data Maps was only studied for usage with neural networks. For use with \emph{any} ML model, it can be extended using our framework (see Section \ref{p4-formulation} and Appendix \ref{adapt}). AUM could of course be used with any ML model with output logits. GraNd does not facilitate this, as only ML models where gradients can be computed are usable. Finally, since JTT, simply captures model errors it is easily usable with any ML model.
\end{itemize}

\textbf{Tangential related work.}
We wish to highlight that while methods such as Influence functions \cite{koh2017understanding} and Data Shapley \cite{ghorbani2019data} may seem related, firstly they have fundamentally different goals around capturing the contributing role of examples to the models decision boundary and predictions. Hence, they are considered as tangential to our work of characterizing data into subgroups. Additionally, these methods assign an instance importance, related to its contribution to the decision boundary of a specific model. This is different from our task of characterizing the properties inherent in the data and then using it to stratify the examples into subgroups.

Another key difference lies in the computational requirements, where both Influence functions and Data Shapley are highly compute intensive, on top of the model training. By contrast, Data-IQ's characterization can be obtained simply by virtue of training an ML model.

\subsection{Data-IQ adapted to any model}\label{adapt}

\begin{figure}[!h]
    \centering
    \includegraphics[width=0.65\textwidth]{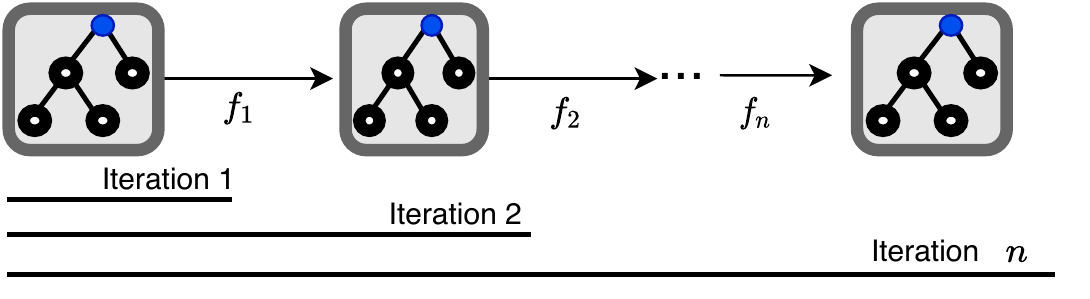}
    \caption{Pseudo-ensemble for GBDTs for usage with Data-IQ, where each sub-model is a checkpoint}
    \label{fig:xgboost_fig}
\end{figure}
Data-IQ's formulation allows us to use it with \emph{any} ML model training in stages (i.e. iterative learning) such as gradient boosting decision trees (GBDTs) methods - such as XGBoost, LightGBM etc. We simply need a set of checkpoints through training. 

This property is incredibly important for practical utility, since methods such as GBDTs or XGBoost are widely used by practitioners due to their performant nature on tabular data (often outperforming neural networks). GBDTs iteratively combines weak models and has shown great success on tabular data, often outperforming neural networks. Hence, having our method applicable for GBDTs in addition to neural networks adds to the broad utility.  

Before outlining why methods such as GBDT's fit the Data-IQ pardigm, we provide a brief overview.
Formally, given a dataset $\Dtrain$, GBDT iteratively constructs a model $F:X \rightarrow \mathbb{R}$ to minimize the empirical risk. At each iteration $e$ the model is updated as: 
\begin{equation}\label{eq:update}
F^{(e)}(\mathbf{}{x}) = F^{(e-1)}(\mathbf{x}) + \epsilon h^{(e)}(\mathbf{x}),
\end{equation}

We now provide guidance on how to apply Data-IQ to such methods. Naively, we could construct the checkpoints as an ensemble of multiple independent GBDT's. However, this is inefficient as the space and time complexity scales with $N$ models. To avoid increasing the overhead of a single model (from a practitioner perspective), we create a pseudo-ensemble using a single GBDT, see Figure \ref{fig:xgboost_fig}. 

Similar to neural networks, the iterative learning nature of a GBDT means that the sequential submodels can be considered as checkpoints. Formally, each sub-model has parameters $\mathbf{\theta}^{(i)}$, hence the ensemble of checkpoints can be described as $\Theta = \{\mathbf{\theta}^{(i)}, 1 \le i \le N\}$. 

We then apply Data-IQ as normal to the checkpoints ($\theta_{1}, \theta_{2}...\theta_{E}$). The flexibility of this approach is that it applies both to training a new model, but interestingly, we can also apply this to an ALREADY trained model by looping through the structure to create the pseudo-ensemble.

\subsection{Why does Data-IQ produce a bell-shape?}
We now theoretically analyze why Data-IQ has a bell-shaped curve. We show if the x-axis is aleatoric uncertainty and if the y-axis is predictive confidence, that we always obtain a bell-shaped curve.

Recall two terms from our formulation: (1) $\P(x,\theta)$: the model's predictive confidence for an input $x$ applied to a model with parameters $\theta$, (2) $\Bar{\P}(x)$: the model's average confidence for the ground-truth class. Now assume $\P(x,\theta)$ is peaked around $\Bar{\P}(x)$, as shown in Figure \ref{fig:bell} below.
\begin{figure}[!h]
    \centering
    \vspace{-3mm}
    \includegraphics[width=0.3\textwidth]{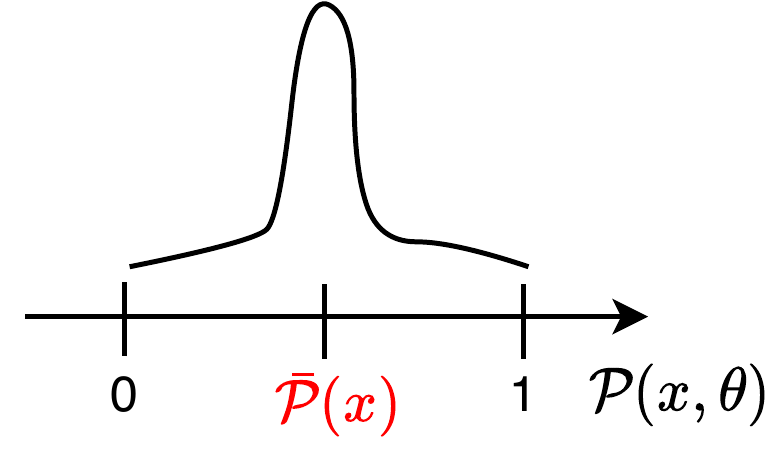}
    \caption{$\P(x,\theta)$ peaked around $\Bar{\P}(x)$}
    \label{fig:bell}
    \vspace{-3mm}
\end{figure}

\begin{figure*}[!h]
  \centering
  \subfigure[Theoretical bell-shape]{\includegraphics[width=0.3\textwidth]{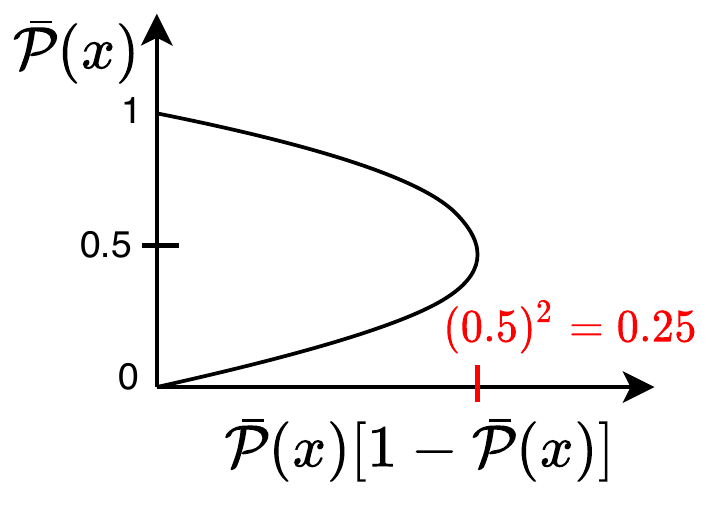}}\quad\quad
  \subfigure[Real bell-shape]{\includegraphics[width=0.25\textwidth]{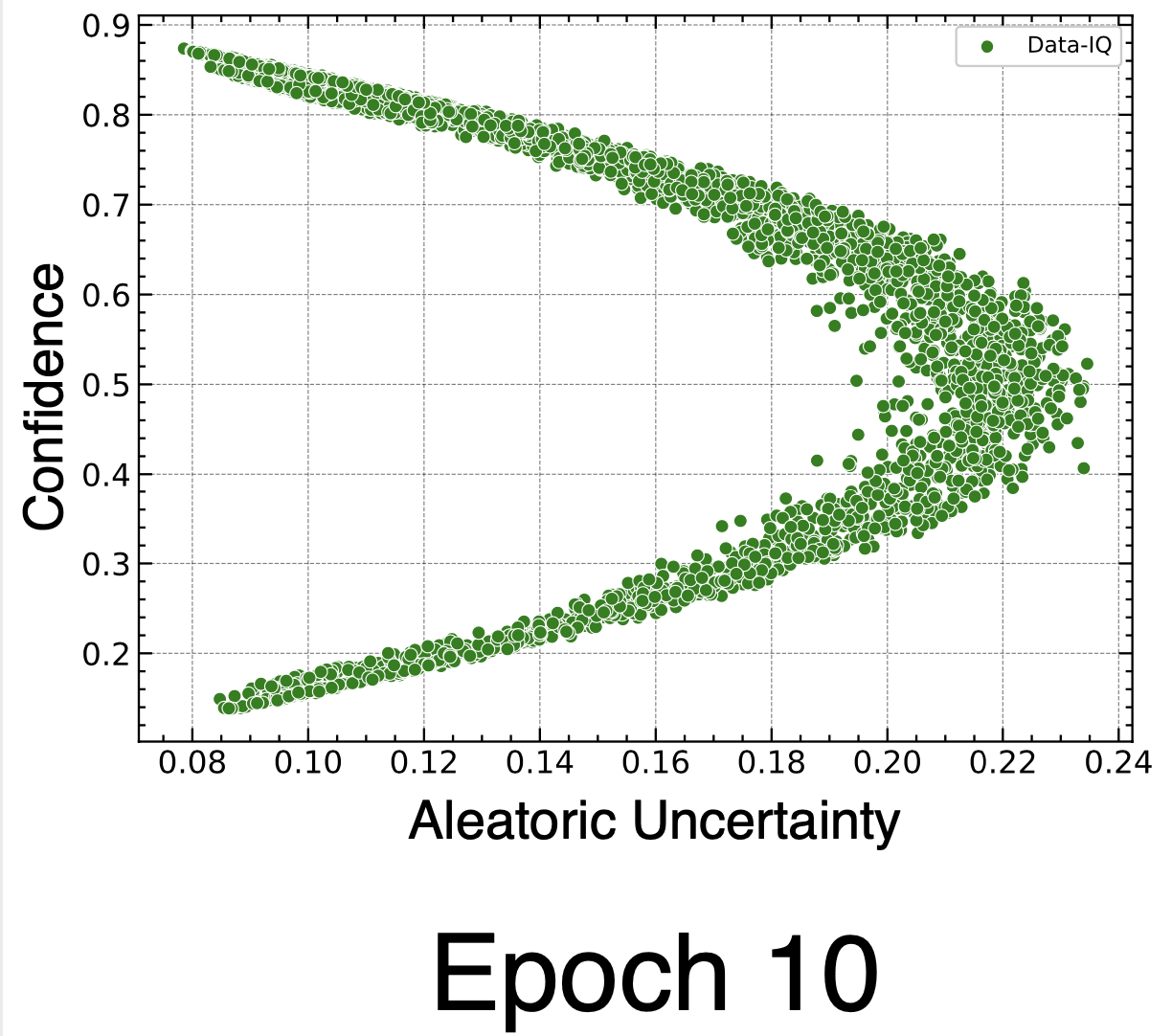}}
  \vspace{-2mm}
  \caption{The bell-shapes derived theoretically and practically match.}
  \label{fig:bell_values}
\end{figure*}

Based on this, when we compute Data-IQ and obtain the following axes. This results in Figure \ref{fig:bell_values} (a) with the bell-shape, which matches what we observe in practice (Figure \ref{fig:bell_values} (b))
\begin{itemize}
    \item y-axis: $ \nicefrac{1}{E} \sum_{e=1}^E \P(x, \theta_e) \approx \bar{\mathcal{P}}(x)$
    \item x-axis: $\frac{1}{E} \sum_{e=1}^{E} \mathcal{P}(x,\theta_e)(1-\mathcal{P}(x,\theta_e))\approx \bar{\mathcal{P}}(x)[1-\bar{\mathcal{P}}(x)]$
\end{itemize}

\subsection{Evaluating Data-IQ and Data Maps over stages}

\paragraph{Goal.} Recall that we utilize \emph{any} ML model trained in stages (e.g. epochs or iterations), for both Data-IQ and Data Maps. In the main paper, we train our models to convergence, with early stopping on a validation set. As is common practice, early stopping is done to prevent the model from overfitting. Once early stopping kicks in and we stop model training, we then compute the Data-IQ and Data Maps, averaged over all previously checkpointed models.

This brings two questions to mind, that we seek to answer:
\begin{enumerate}
    \item How do Data-IQ and Data Maps differ at various stages (epochs)?
    \item What happens if we don't do early stopping and let the model overfit?
\end{enumerate}

\begin{figure}[!h]
    \centering
    \includegraphics[width=0.925\textwidth]{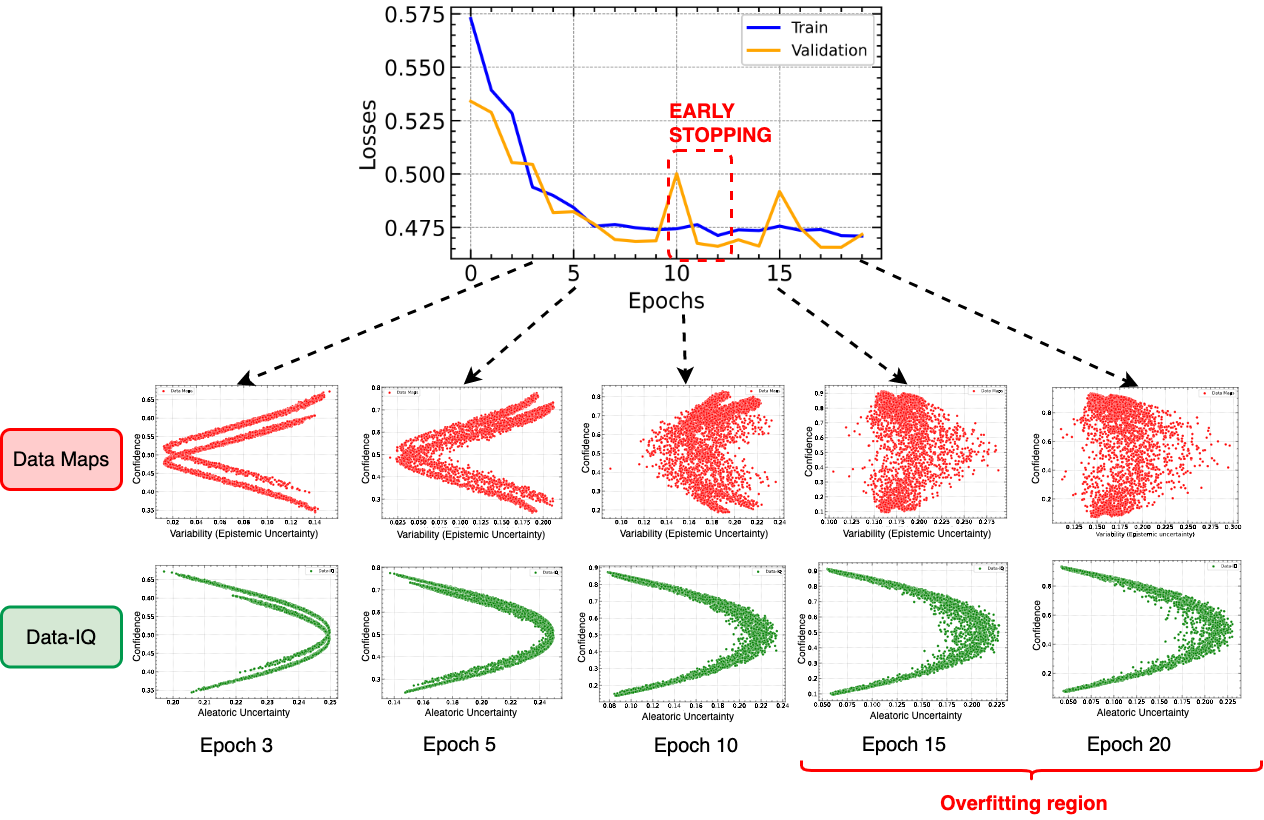}
    \vspace{-1mm}
    \caption{Evaluation of Data-IQ and Data Maps over various epochs (stages), even letting the model overfit. Data-IQ is more stable, whereas Data Maps changes shape after overfitting which is undesirable.}
    \label{fig:evolution_stages}
\end{figure}

Using Figure \ref{fig:evolution_stages}, we answer both questions. For this we train a neural network and around 10 epochs early stopping would normally kick in. However, instead of using early stopping, we continue training the neural network until 20 epochs, thereby letting the model overfit.

\paragraph{Takeaway.}
\begin{enumerate}
    \item \textbf{Differences across epochs:} Data Maps changes significantly over the course of training, This is due to its sensitivity as the model parameters are changed/updated through training. This sensitivity is reflected in the variability (epistemic uncertainty) on the x-axis. In contrast, Data-IQ is less sensitive and rapidly converges. The rationale is that Data-IQ models the inherent uncertainty in the data (aleatoric), thereby being more stable and less sensitive.
    \item \textbf{Model overfits:} An interesting phenomenon occurs after the early stopping phase and into the overfitting stage. DataIQ remains consistent/stable, as expected. It is during this overfitting stage that the Data Maps shape changes, and begins to look similar to Data-IQ. The reason for this is that after $\approx$10 epochs (when the model begins to overfit), the model parameters do not get updated much (the loss doesn't change). The small changes of model parameters are then reflected by the epistemic/model uncertainty not changing much (x-axis of Data Maps). The minimal change means that when we average at the end of a longer training - the effects get washed out. This washing out of the effect causes the change in shape by Epoch 20. We will delve into this point in greater detail in the next sub-section as it explains the difference of Data Maps in our setting as compared to other settings. 
\end{enumerate}

\subsection{Data Maps: difference in our setting}

\paragraph{Goal.} We now take a deep dive, based on the interesting findings at the model ovefitting stage, where the Data Maps shape changes. We believe that the difference of our setting from the original NLP setting of Data Maps provides a clear explanation for why the shape we observe for Data Maps is different from the shape observed in the original paper \cite{swayamdipta2020dataset}.

In typical NLP settings, the models are larger and highly parameterized. Hence, the models are trained for a large number of epochs (sometimes hundreds). However, in the tabular setting, models need to train for significantly fewer (as is evident by the early stopping around $\approx 10$ epochs. Further, our models are much shallower compared to highly parameterized NLP models, motivating why our setting requires early stopping to prevent overfitting.

\paragraph{Takeaway.} We now begin our deep dive. As we saw in Figure \ref{fig:evolution_stages}, after around 10 epochs, the training loss plateaus. This means that subsequently the model parameters do not get updated much over training. The result of this is the epistemic uncertainty would have much less variability after 10 epochs. Thus, when we average the epistemic uncertainty for longer epochs these minimal changes would wash out effects from earlier in training.

This effect over longer training, explains why the shape of Data Maps is different in the NLP settings with large number of epochs and, assuming, we let the model overfit in our setting. 

What implications does this have? For this we assess the evolution of three examples where we know their ground truth subgroup ($\easy, \ambiguous, \hard$).

\begin{figure}[!h]
    \centering
    \includegraphics[width=0.75\textwidth]{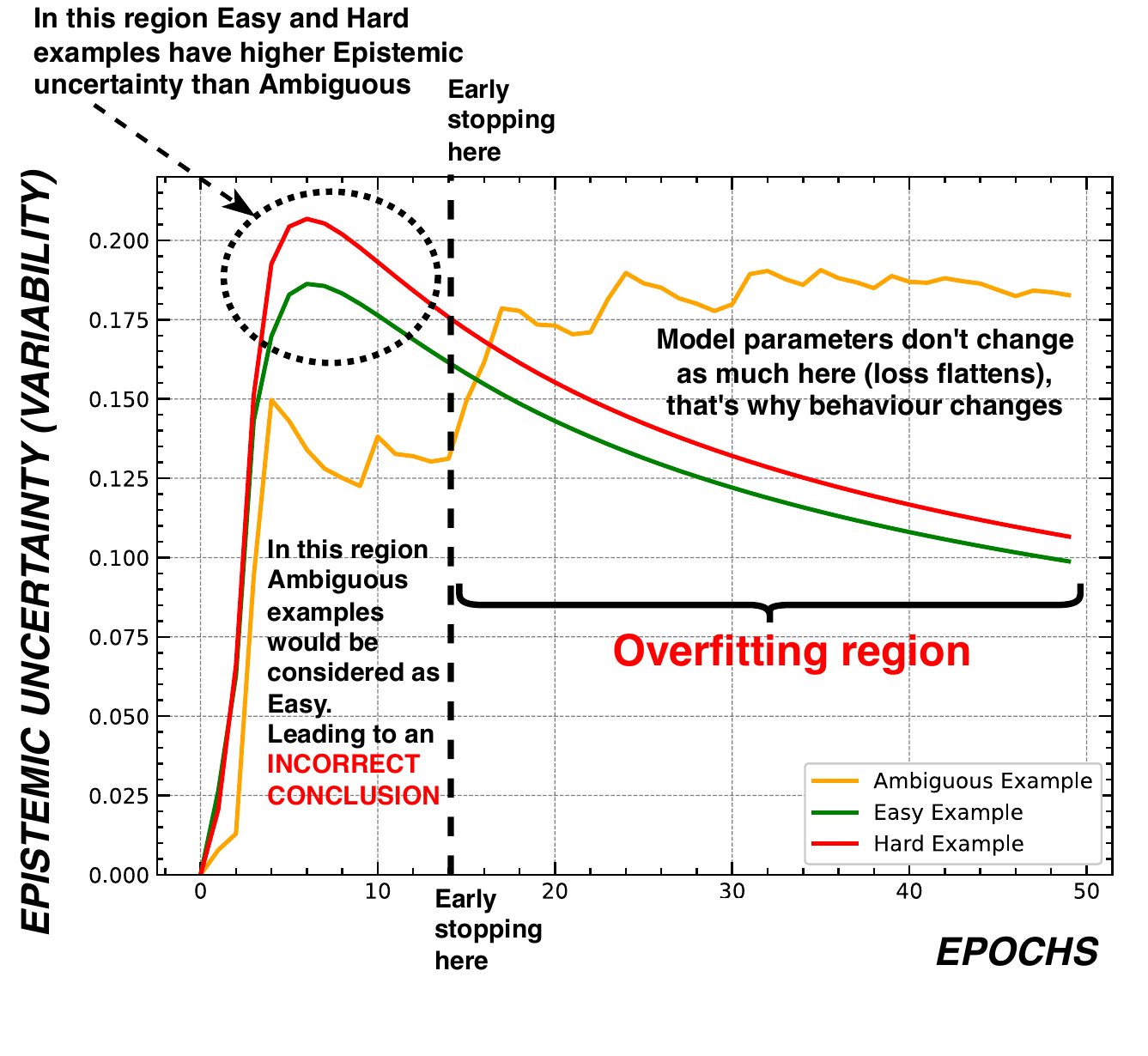}
    \vspace{-5mm}
    \caption{Data Maps: Epistemic Uncertainty (variability) evolution over epochs per subgroup}
    \label{fig:epistemic_evolution}
\end{figure}

First, we assess Data Maps which uses epistemic uncertainty. Figure \ref{fig:epistemic_evolution} shows the evolution of epistemic uncertainty (Data Maps) for the three subgroup examples during training. We see that during model training (before early stopping), that the $\easy$ and $\hard$ examples, in fact, have higher epistemic uncertainty than the $\ambiguous$ example. This would lead us to the incorrect conclusion that the $\ambiguous$ example is Easy.

In practice, we do not want to let our model knowingly overfit. However, we can only obtain the correct characterization for Data Maps if we let the model overfit and wash out the earlier effects. This highlights a limitation of Data Maps vs Data-IQ.

We contrast this with Data-IQ which uses the Aleatoric uncertainty. We not only have consistent behavior for the $\ambiguous$ examples, but also irrespective of whether we use early stopping or let the model overfit, we can obtain the correct characterization of the examples.

\begin{figure}[!h]
    \centering
    \includegraphics[width=0.75\textwidth]{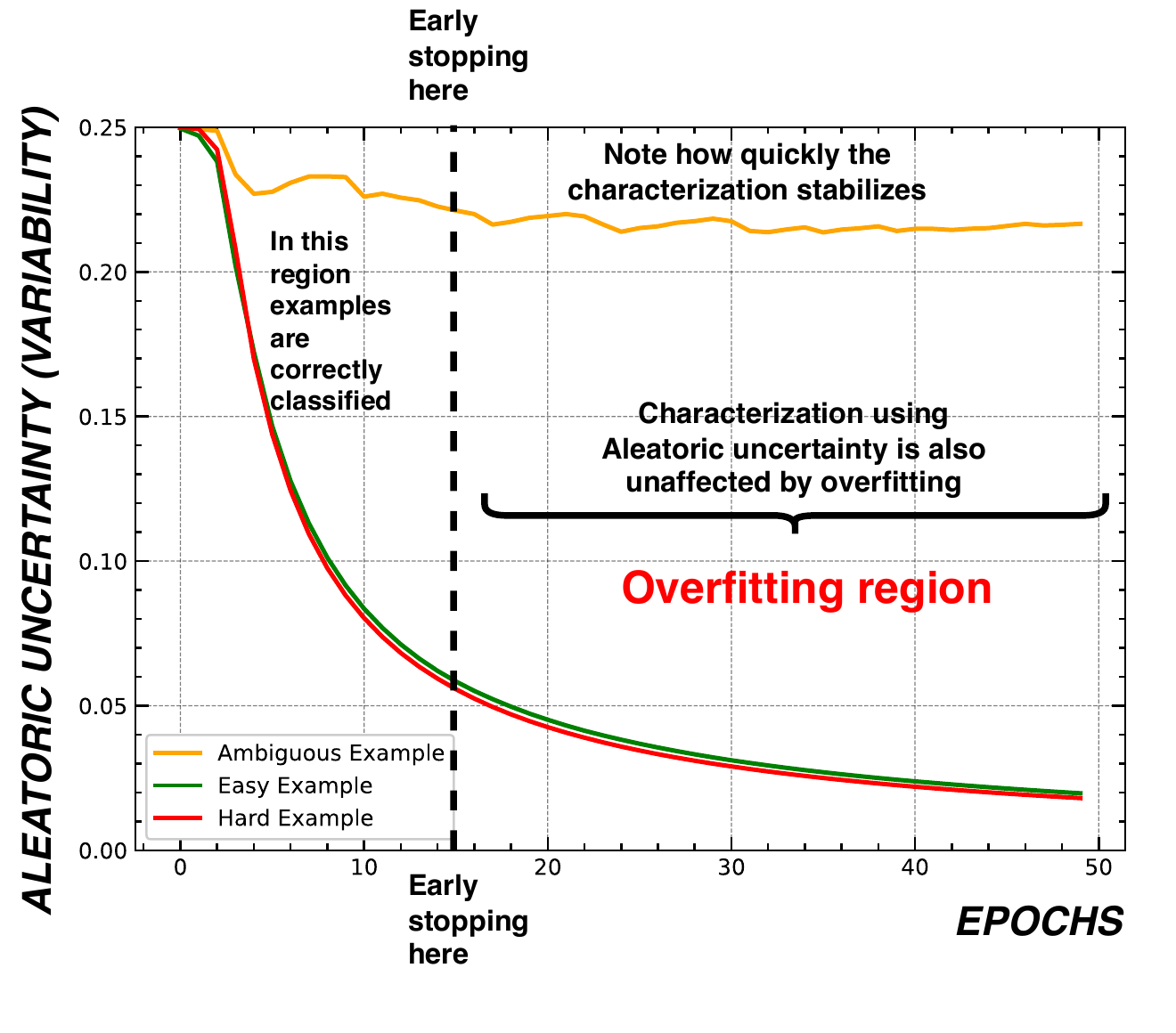}
    \caption{Data-IQ: Aleatoric Uncertainty evolution over epochs per subgroup}
    \label{fig:alaeatoric_evolution}
\end{figure}

\begin{figure*}[!h]
  \centering
  \subfigure[Data Maps @ 10 epochs (Correct model)]{\includegraphics[width=0.45\textwidth]{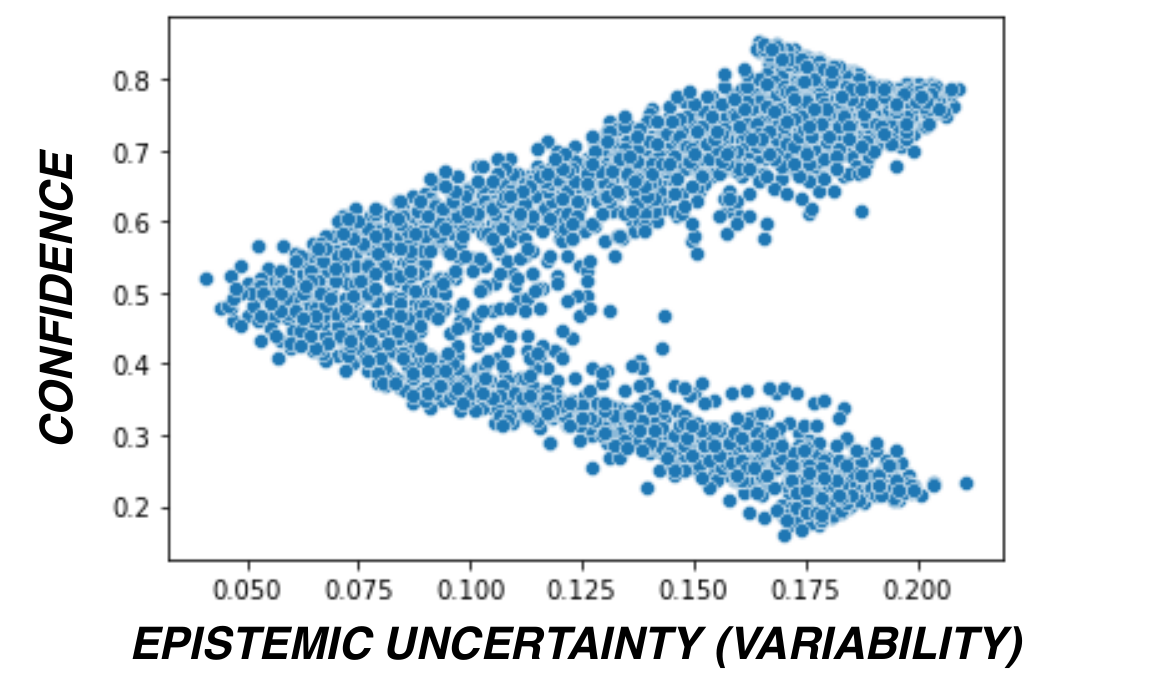}}\quad\quad
  \subfigure[Data Maps @ 30 epochs (OVERFIT model)]{\includegraphics[width=0.45\textwidth]{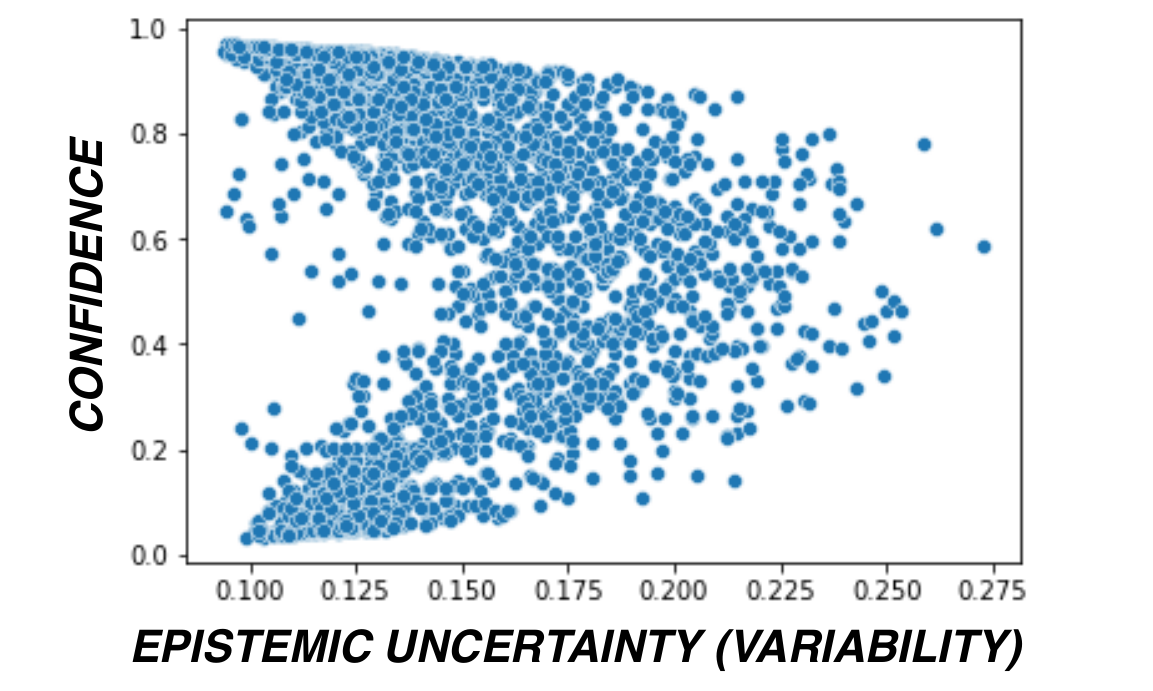}}
  \caption{Comparison of Data Maps when model is just right and overfits, illustrating how the shape changes}
  \label{fig:shapes}
\end{figure*}

Finally, we look at the Data Maps shapes at the point where we perform early stopping (Figure \ref{fig:shapes}(a)) and at the end of training if we let the model severely overfit (Figure \ref{fig:shapes}(b)). 

Figure \ref{fig:shapes} shows we can only obtain the expected shape from the original paper \cite{swayamdipta2020dataset}, if we let the model overfit. This is of course not an ideal solution, as in practice we do not want to overfit our models by training for extra epochs. Especially in our tabular setting with less parameterized models, on smaller tabular datasets.

This motivates further why indeed we do require Data-IQ and why it's characterization of examples based on aleatoric uncertainty is more useful and principled.

\subsection{Data-IQ thresholds}
As outlined in Section \ref{formulation:stratification}, we stratify samples based on the Aleatoric uncertainty and predictive confidence. 

Typically, for aleatoric uncertainty we set the percentile $P_{n}=50$th percentile. The rationale for this is any example where the aleatoric uncertainty is greater than the 50th percentile should be considered $\ambiguous$. This is based on the definition of $\ambiguous$ examples. Of course, this threshold can be updated by practitioners.

This means that the examples below $P_{50}$ of the aleatoric uncertainty are then $\easy$ or $\hard$. To differentiate the two, we look at the predictive confidence. We consider examples with high confidence, on the true label as $\easy$ and those with low confidence, on the true label as $\hard$. This means that we need to practically set $\Cupper$ and $\Clower$. 

We define a threshold $thresh \in \{0,0.5\}$ such that $\Cupper=1-thresh$ and $\Clower=thresh$. An obvious threshold=0.25 such that $\Cupper=0.75$ and $\Clower=0.25$. However, we present a practical method that could be applied to determine this threshold. We note that from our characterizations, examples with high aleatoric uncertainty are often uncertain about their predicitive confidence. Hence, they never have very high or low uncertainty, rather the uncertainty is around 0.5. However, since it is never exactly 0.5 this motivates a band namely $\Cupper$ and $\Clower$.

Now assume for any dataset, we train a model and apply Data-IQ. We can then sweep $thresh \in \{0,0.5\}$ and assess the proportion of examples in each group.

\begin{figure}[!h]
  \centering
  \subfigure[
\scriptsize{Data-IQ subgroup ambiguity proportion is reduced as more informative features are acquired.}]{\includegraphics[width=0.4\textwidth]{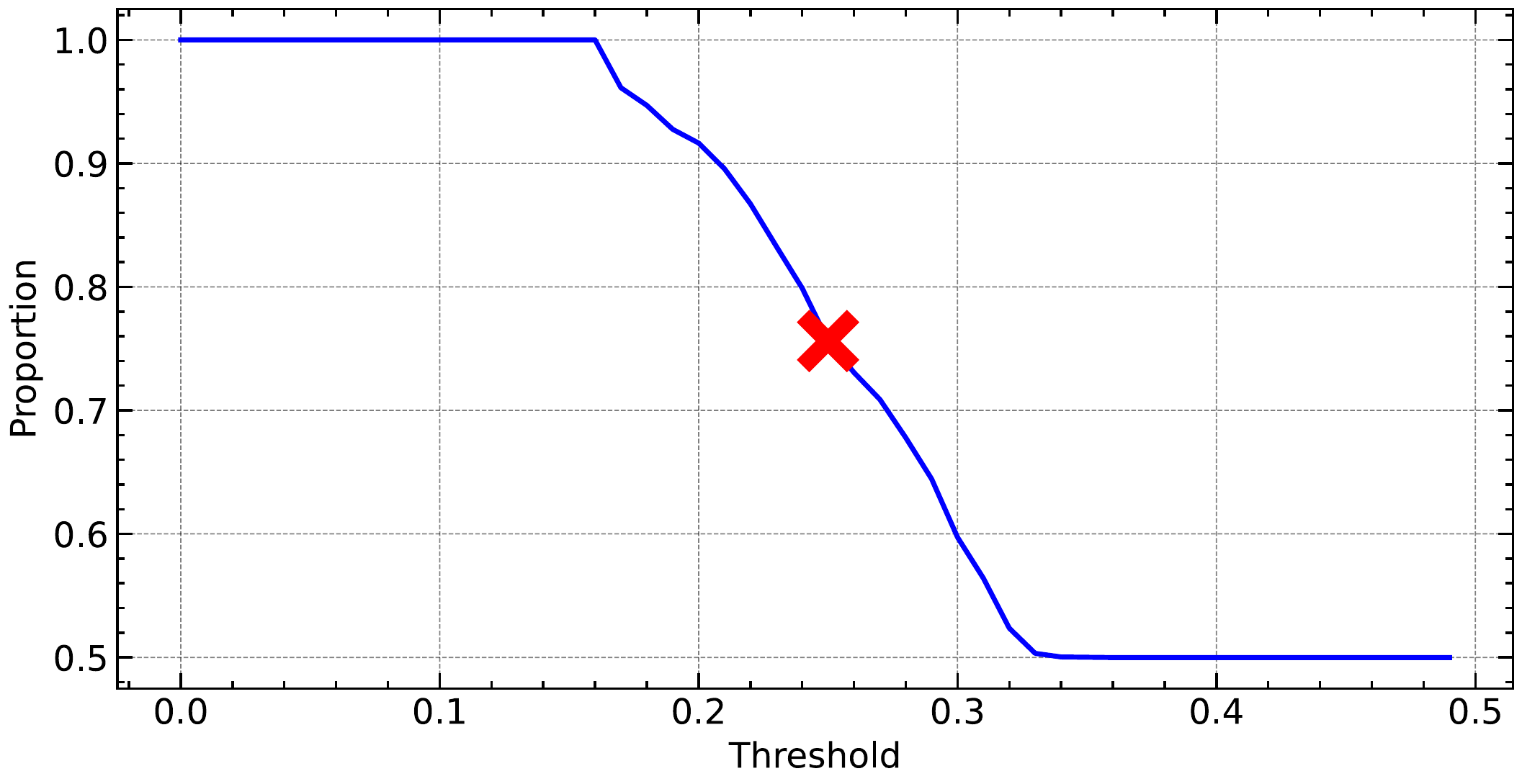}}\quad
  \subfigure[\scriptsize{Data-IQ aleatoric uncertainty remains stable for Ambiguous, reduces for others as features are acquired.} ]{\includegraphics[width=0.4\textwidth]{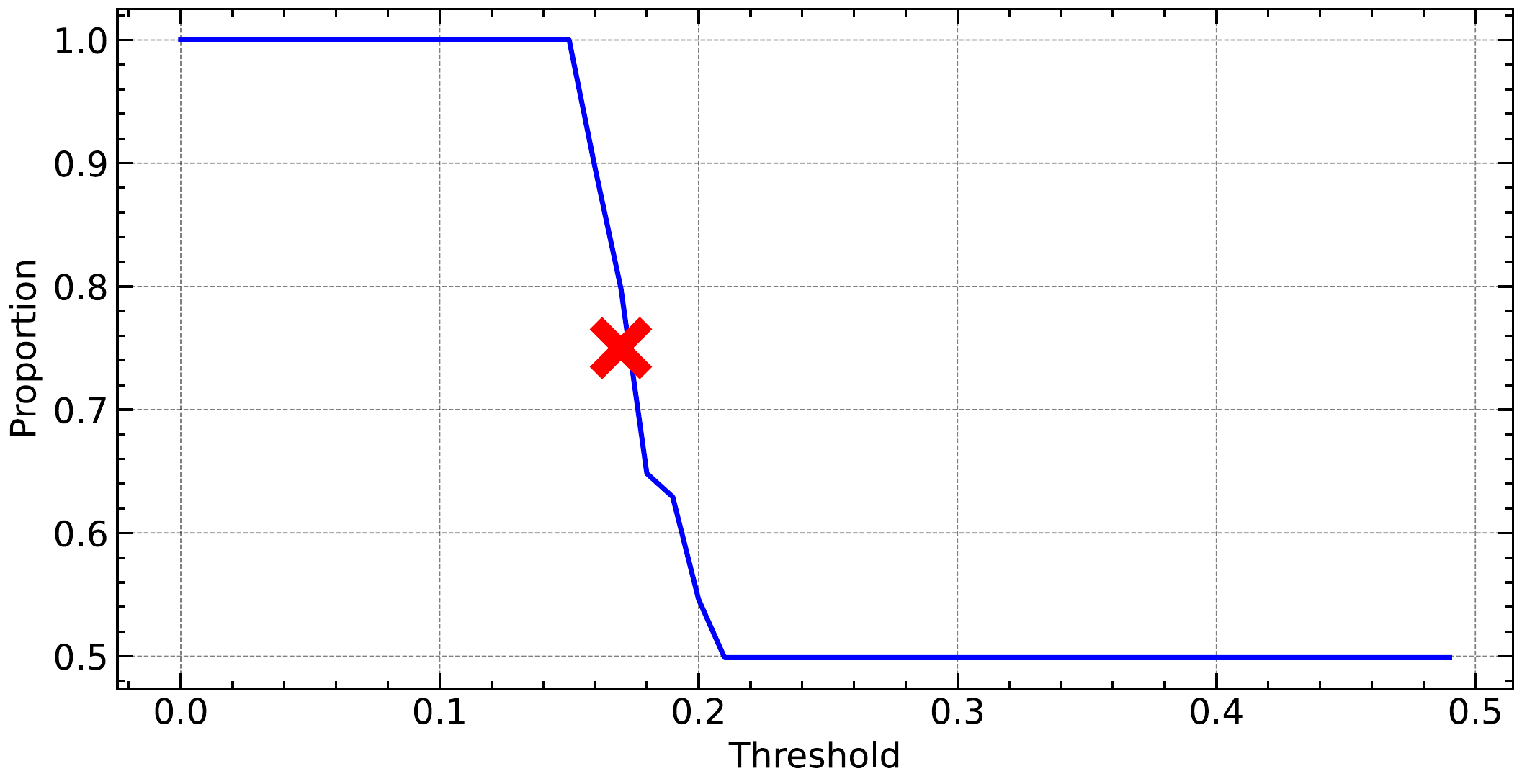}}\quad\\
  \subfigure[\scriptsize{Data Maps subgroup proportions largely unaffected as features acquired.}]{\includegraphics[width=0.4\textwidth]{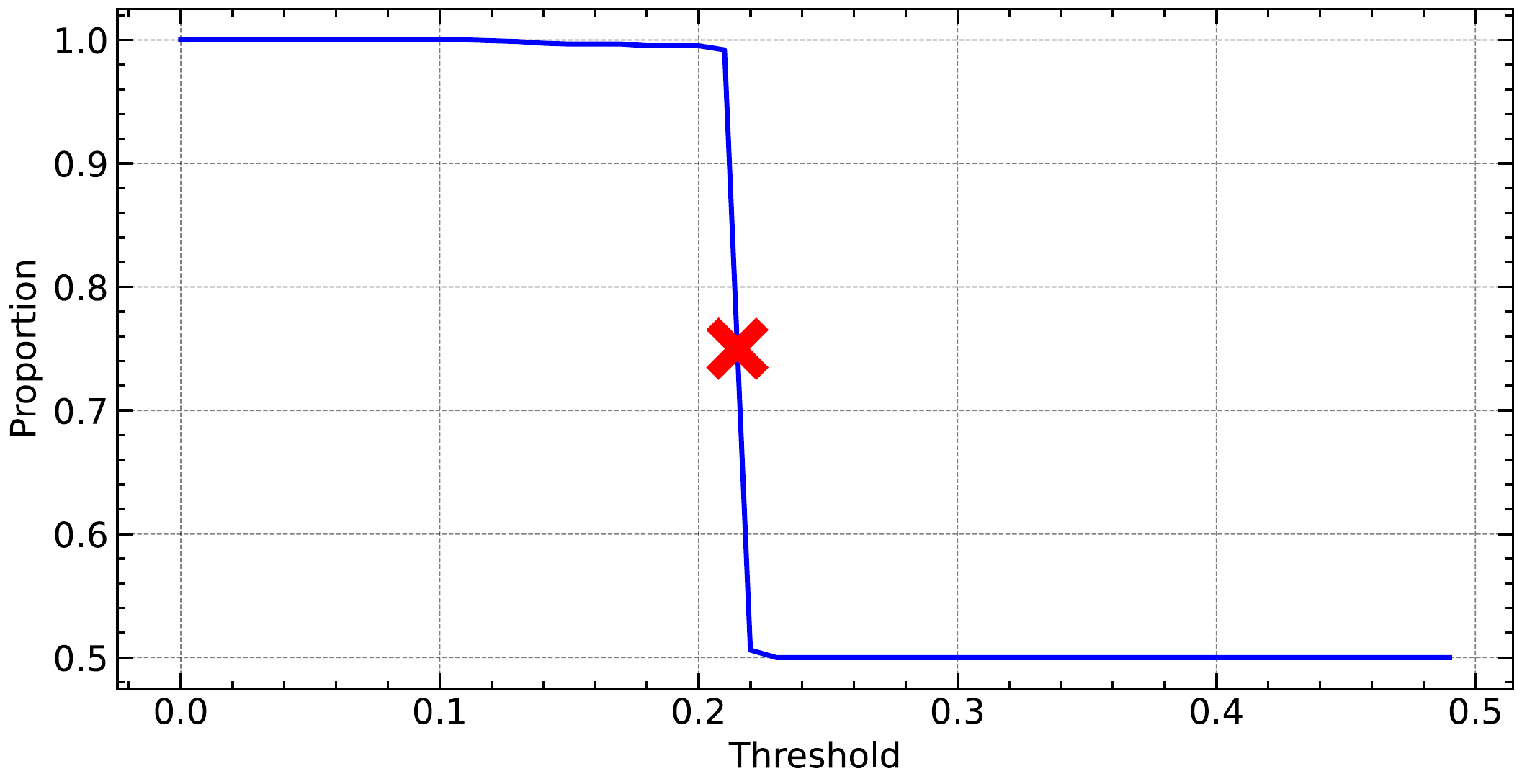}}\quad
  \subfigure[\scriptsize{Data Maps variability increases across subgroups as features acquired.} ]{\includegraphics[width=0.4\textwidth]{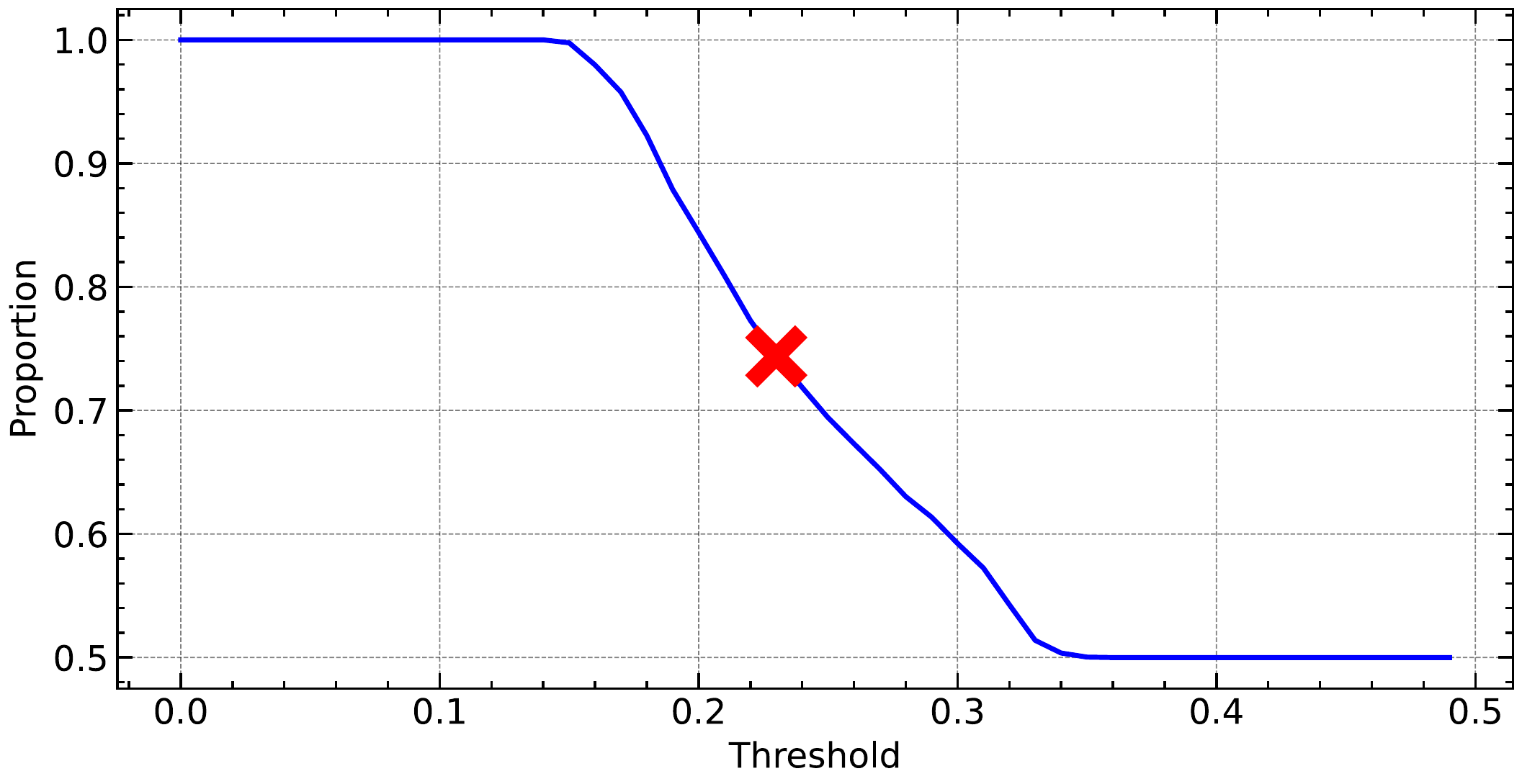}}\quad
  \caption{SUPPORT Quantifying the value of feature acquisition based on change in ambiguity. Only Data-IQ captures this relationship.}
   \label{fig:thresh}
\end{figure}

We show results in Figure \ref{fig:thresh} below as we sweep the threshold. For low threshold values, the proportions are, of course, stable (selecting the most confident samples). However, there is a threshold value beyond which where the curve rises until a threshold knick point. After that knick-point there is stability once again. We propose that this knick-point threshold of stability be used as the threshold in practice. The rationale is that this will allow for better consistency of data characterization.

\newpage

\subsection{Data-IQ subgroup characterization}

With Data-IQ, we can obtain a plot of the data where the x-axis represents aleatoric uncertainty and the y-axis predicitive confidence for the true label. As discussed this means we obtain a bell-shaped characterization. Our goal is highlight where on the plot each subgroup lies. i.e. illustrate the location of  $\easy, \ambiguous$ and $\hard$ on the plot.

\begin{figure}[!h]
    \centering
    \vspace{-3mm}
    \includegraphics[width=0.75\textwidth]{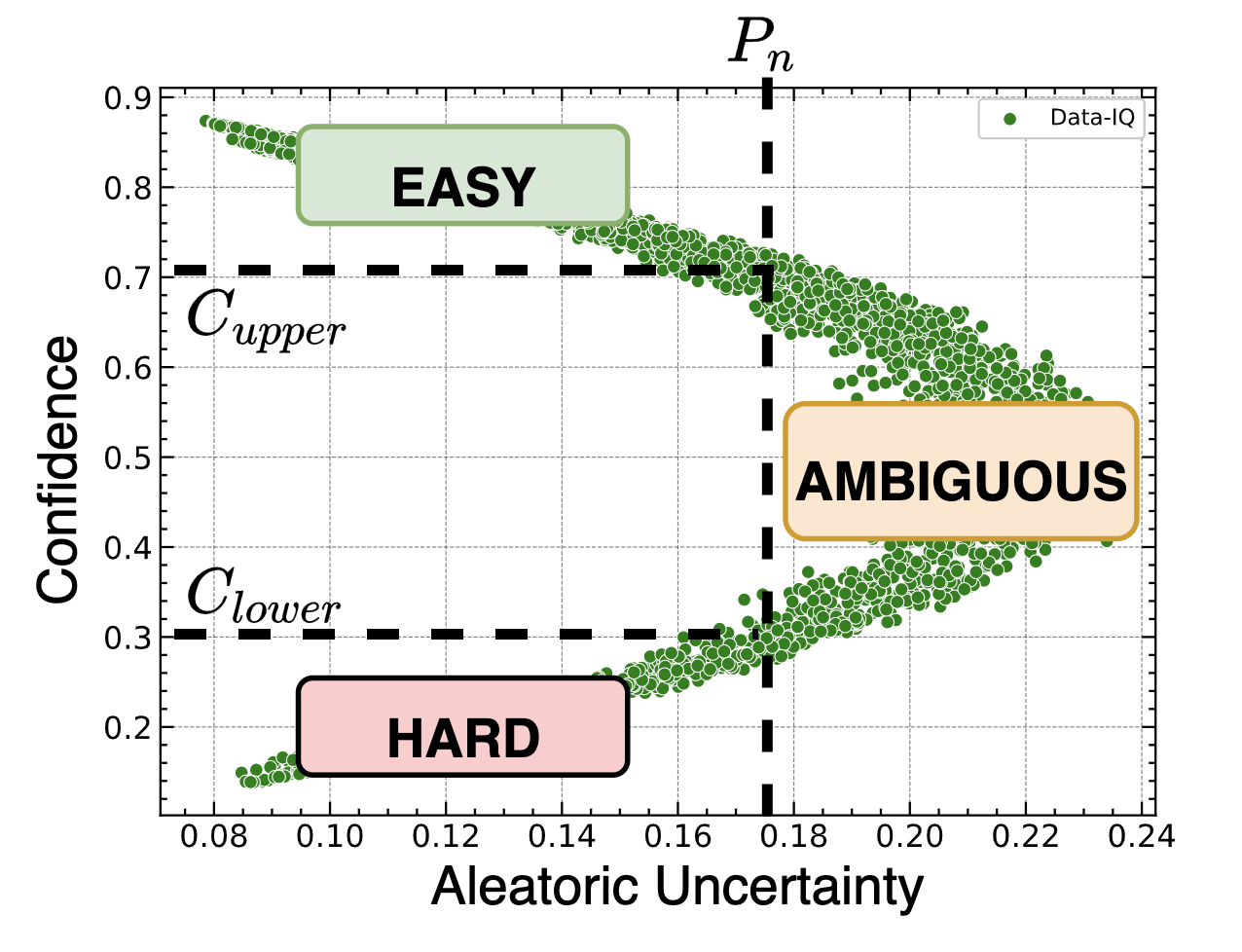}
    \caption{$\P(x,\theta)$ peaked around $\Bar{\P}(x)$}
    \label{fig:diq_shape}
\end{figure}

We highlight in Figure \ref{fig:diq_shape}, that $\ambiguous$ examples more likely to lie on the right side with higher aleatoric uncertainty. We note that based on our characterization this region of the plot never has high and low predictive uncertainty. Rather, it typically only has uncertainty approximately around 0.5.

\clearpage

\section{Benchmarks \& Experimental Details} \label{sec:appendixB}
\subsection{Benchmarks, Datasets and Implementation details}\label{appendix:benchmarks}

\subsubsection{Data-IQ}
The description of Data-IQ is detailed in the main paper, where we stratify examples based on aleatoric uncertainty and predictive confidence.
\paragraph{Implementation details.} We implement Data-IQ via a class that is versatile and can plug into any framework including Pytorch, Tensorflow/Keras, Skorch and Scikit-learn style APIs. The rationale is that since Data-IQ can be used with any model, the functionality should be equally plug and play from a usability perspective. Note that for libraries including Pytorch, Tensorflow/Keras, Skorch, we can easily include Data-IQ simply by adding it to the training callbacks of the respective training loops. 

\subsubsection{Data-IQ algorithm}
Algorithm 1 below presents the Data-IQ algorithm in more detail.

\begin{algorithm}[H]
 \KwData{Dataset $\mathcal{D}$}
 \KwResult{Subgroup assignment $g$ per sample $x$, i.e. $g(x,\Dtrain)$}
  \textbf{Inputs:} $\Cupper = 0.75$, $\Clower = 0.25$, $P_{50}$ = 50th percentile \\
  \textbf{Training:} Train model ($f$) for $E$ epochs. At each epoch $e$, compute $\P(x, \theta_e)$ for datapoint $x$, where $\theta_e$ is the model parameters at epoch $e$.  \\
  \textbf{Characterize samples into subgroups ($g(x,\Dtrain)$):} 
  \For{$(x_i, y_i)$ in $\Dtrain$}
{\textbf{Metrics:}
  $\Bar{\P}(x_i) = \nicefrac{1}{E} \sum_{e=1}^E \P(x_i, \theta_e)$;\\
  $\val(x_i) = \frac{1}{E} \sum_{e=1}^{E} \mathcal{P}(x_i,\theta_e)(1-\mathcal{P}(x_i,\theta_e)) $\\
  
 \uIf{$ \Bar{\P}(x_i) \ge \Cupper \ \wedge \ \val(x_i) < P_{50}\left[ \val(\Dtrain) \right] $}{
    Assign $\easy$ subgroup 
  }
  \uElseIf{$\Bar{\P}(x_i) \le \Clower \ \wedge \ \val(x_i) < P_{50}\left[ \val(\Dtrain) \right] $}{
    Assign $\hard$ subgroup 
  }
  \Else{
    Assign $\ambiguous$ subgroup
  }
   }

 \caption{Data-IQ Algorithm}
\end{algorithm}

\begin{align} \label{eq:groups_training_new}
\vspace{-10mm}
    g(x,\Dtrain) = \begin{cases}
$\easy$ & \mathrm{if} \ \Bar{\P}(x) \ge \Cupper \ \wedge \ \val(x) < P_{50}\left[ \val(\Dtrain) \right]  \\
$\hard$ & \mathrm{if} \ \Bar{\P}(x) \le \Clower \ \wedge \ \val(x) < P_{50}\left[ \val(\Dtrain) \right]  \\
$\ambiguous$ & \mathrm{otherwise}  \\
\end{cases}
\vspace{-5mm}
\end{align}

\subsubsection{Data Maps}
Similar to Data-IQ; Data Maps has been detailed in the main paper.        
\paragraph{Implementation details.}  Our implementation of Data Maps can be used with the flexible approach we built for Data-IQ. We adapt the implementation of Data Maps from \footnote{https://github.com/allenai/cartography}.

\subsubsection{AUM}
The Area-under-the-Margin (AUM) metric first computes the margin for data $x,y$ at epoch $t$.

        \begin{equation}
  M^{(t)}(x,y) =
  \overbrace{z^{(t)}_{y}(x)}^{\text{assigned logit}} -
  \overbrace{\textstyle \max_{i \ne y} z^{(t)}_i (x)}^{\text{largest other logit}}.
  \label{eqn:margin}
\end{equation}

Over all epochs the AUM is then computed as the average of these margin calculations.
\begin{equation}
  AUM(x,y)
  = \textstyle{ \frac{1}{T} \sum_{t=1}^{T}} \: M^{(t)}(x,y),
  \label{eqn:aum}
\end{equation}
\paragraph{Implementation details.}  The benchmark is based on \cite{pleiss2020identifying} and we use the implementation from \footnote{https://github.com/asappresearch/aum}.

\subsubsection{GraNd}

The gradient norm at epoch $t$ for an input $x,y$ is computed as :\\
$\chi_{t}(x, y)=\mathbb{E}\left\|\sum_{k=1}^{K} \grad_{f^{(k)}} \ell\left(f_{t}(x), y\right)^{T} \psi_{t}^{(k)}(x)\right\|_{2}$\\
where $\psi_{t}^{(k)}(x)=\grad_{\mathbf{w}_{t}} f_{t}^{(k)}(x)$.

\paragraph{Implementation details.}  The benchmark is based on \citep{paul2021deep}. 
We adapt the Jax implementation from \footnote{https://github.com/mansheej/data\_diet}
to Pytorch with the help of \footnote{https://github.com/cybertronai/autograd-lib}. 

\subsubsection{JTT}
The Just-Train-Twice (JTT) method is a two-stage method. For a model, trained via ERM, it first identifies errors in the training examples during training $i.e. y \neq \hat{y}$.
 
The second step is to train a final model, where the error samples are increased via weighting.

\begin{align}
\label{eqn:steptwo}
     L_{\text{up-ERM}}(\theta, E) = ( \lambda \sum \limits_{(x, y) \in E} 
     \ell(x, y; \theta) 
     +  \sum \limits_{(x, y) \not\in E} \!\!\ell(x, y; \theta)),
\end{align}
where $\lambda$ is the weighting factor and $E$ errors.

The goal is that the weighting factor will upsample the samples with errors and by training on this augmented dataset it will aid to improve worst-group performance.

\paragraph{Implementation details.} Our benchmark is based on \citep{liu2021just} and we adapt the implementation from \footnote{https://github.com/anniesch/jtt}.

\subsection{Datasets}
We describe the four real-world medical datasets in greater detail.
\paragraph{SEER Dataset} 
The SEER dataset is a publicly available dataset consisting of 240,486 patients enrolled in the American SEER program~\cite{duggan2016surveillance}. The dataset consists of features used to characterize prostate cancer: including age, PSA (severity score), Gleason score, clinical stage, treatments etc. A summary of the covariate features can be found in Table \ref{tab:seer_features}. The classification task is to predict patient mortality, which is binary label $\in \{ 0 , 1 \}$.

The dataset is highly imbalanced, where $~94\%$ of patients survive. Hence, we extract a balanced subset of of 20,000 patients (i.e. 10,000 with label=0 and 10,000 with label=1).

\begin{table}[h]
\caption{Summary of features for the SEER Dataset \cite{duggan2016surveillance}}
\scalebox{0.8}{
\begin{tabular}{ll}
\toprule
Feature & Range \\ 
\midrule
Age & $37-95$ \\ 
PSA & $0-98$ \\ 
Comorbidities & $0, 1, 2, \geq 3$ \\ 
Treatment & Hormone Therapy (PHT), Radical Therapy - RDx (RT-RDx), \newline Radical Therapy -Sx (RT-Sx), CM \\ 
Grade & $1, 2, 3, 4, 5$ \\ 
Stage & $1, 2, 3, 4$ \\ 
Primary Gleason & $1, 2, 3, 4, 5$ \\ 
Secondary Gleason & $1, 2, 3, 4, 5$ \\ 
\bottomrule
\end{tabular}}
\label{tab:seer_features}
\end{table}

\paragraph{CUTRACT Dataset}

The CUTRACT dataset is a private dataset consisting of 10,086 patients enrolled in the British Prostate Cancer UK program~\citep{prostate}. Similar, to the SEER dataset, it consists of the same features to characterize prostate cancer. In addition, it has the same task to predict mortality. A summary of the covariate features can be found in Table \ref{tab:cutract_features}.

Once again, the dataset is highly imbalanced, hence we then choose extract a balanced subset of of 2,000 patients (i.e. 1000 with label=0 and 1000 with label=1).

\begin{table}[h]
\caption{Summary of features for the CUTRACT Dataset \cite{prostate}}
\scalebox{0.8}{
\begin{tabular}{ll}
\toprule
Feature & Range \\ 
\midrule
Age & $44-95$ \\ 
PSA & $1-100$ \\ 
Comorbidities & $0, 1, 2, \geq 3$ \\ 
Treatment & Hormone Therapy (PHT), Radical Therapy - RDx (RT-RDx), \newline Radical Therapy -Sx (RT-Sx), CM \\ 
Grade & $1, 2, 3, 4, 5$ \\ 
Stage & $1, 2, 3, 4$ \\ 
Primary Gleason & $1, 2, 3, 4, 5$ \\ 
Secondary Gleason & $1, 2, 3, 4, 5$ \\ 
\bottomrule
\end{tabular}}
\vspace{.5cm}
\label{tab:cutract_features}
\end{table}

\paragraph{Fetal Dataset}
The Fetal dataset \cite{ayres2000sisporto} is a publicly available data set consisting of 2126 patients who underwent fetal cardiotograms (CTG). The patients had specific diagnostic features measured and the CTGs were classified by 3 expert obstertricians (with a consensus label) to assign three fetal states (normal, suspect and pathologic). Our goal is to predict these three fetal states, hence the task is a multi-class classification problem. A summary of the covariate features can be found in Table \ref{tab:fetal_features}.

\begin{table}[h]
\centering
\caption{Summary of features for the Fetal Dataset \cite{ayres2000sisporto}}
\scalebox{0.8}{
\begin{tabular}{ll}
\toprule
Feature & Range \\ 
\midrule
LB (FHR baseline) & $106-160$ \\ 
AC (Acceleration/sec) & $0-260$ \\ 
FM (fetal movements/sec) & $0-564$ \\ 
UC (uterine contractions/sec) & $0-23$ \\ 
ASTV (\% of time with abnormal short term variability) & $12-87$ \\ 
ALTV (\% of time with abnormal long term variability) & $0-91$ \\ 
MLTV (mean value of long term variability) & $0-51$ \\ 
DS (severe deceleration/sec)& $0-1$ \\ 
DP (prolonged deceleration/sec) & $0-4$ \\ 
Nzeros (n histogram zeros) & $0-10$ \\ 
Variance (histogram variance) & $0-269$ \\ 
Tendency (histogram tendency) & $0-1$ \\ 
NSF (fetal class code) & $1 (Normal),2 (suspect),3 (pathologic)$ \\ 
\bottomrule
\end{tabular}}
\vspace{.5cm}
\label{tab:fetal_features}
\end{table}

\paragraph{Covid-19 Dataset}

The Covid-19 dataset \cite{baqui2020ethnic} consists of Covid patients from Brazil. The dataset is publicly available as it is based on SIVEP-Gripe data. The dataset consists of 6882 patients from Brazil recorded between Februrary 27-May 4 2020. The dataset captures risk factors including comorbidities, symptoms, and demographic characteristics. There is a mortality label from Covid-19 making it a binary classification task.  A summary of the characteristics of the covariates can be found in Table \ref{tab:covid_features}.

\begin{table}[h]
\centering
\caption{Summary of features for the Covid-19 Dataset \citep{baqui2020ethnic}}
\scalebox{0.8}{
\begin{tabular}{ll}
\toprule
Feature & Range \\ 
\midrule
Sex & 0 (Female), 1(Male) \\
Age & $1-104$ \\ 
Fever & $0,1$ \\ 
Cough & $0,1$ \\ 
Sore throad & $0,1$ \\ 
Shortness of breath & $0,1$ \\ 
Respiratory discomfort & $0,1$ \\ 
SPO2 & $0-1$ \\ 
Diharea & $0,1$ \\ 
Vomitting & $0,1$ \\ 
Cardiovascular & $0,1$ \\ 
Asthma & $0,1$ \\ 
Diabetes & $0,1$ \\ 
Pulmonary & $0,1$ \\ 
Immunosuppresion & $0,1$ \\ 
Obesity & $0,1$ \\ 
Liver & $0,1$ \\ 
Neurologic & $0,1$ \\ 
Branca (Region) & $0,1$ \\ 
Preta (Region) & $0,1$ \\ 
Amarela (Region) & $0,1$ \\ 
Parda (Region) & $0,1$ \\ 
Indigena (Region) & $0,1$ \\ 
\bottomrule
\end{tabular}}
\vspace{.5cm}
\label{tab:covid_features}
\end{table}

\paragraph{Support Dataset}

The Support  \cite{knaus1995support} is a private dataset which examines outcomes for seriously ill hospitalized patients consisting of 9105 patients. The features include demographics, medical history, clinical variables, and laboratory variables. Our goal is to predict patient mortality. A summary of the covariate features can be found in Table \ref{tab:support_features}.

\begin{table}[h]
\centering
\caption{Summary of features for the Support dataset \cite{knaus1995support}}
\scalebox{0.8}{
\begin{tabular}{ll}
\toprule
Feature & Range \\ 
\midrule
Sex & 0 (Female), 1(Male) \\
Age & $18-102$ \\ 
Number of comorbities & $0-9$\\
ARF/MOSF w/Sepsis  & $0,1$ \\ 
COPD  & $0,1$ \\ 
CHF  & $0,1$ \\ 
Cirrhosis  & $0,1$ \\ 
Coma  & $0,1$ \\ 
Colon Cancer  & $0,1$ \\ 
Lung cancer  & $0,1$ \\ 
ARF/MOSF w/Malig & $0,1$ \\ 
Cancer  & $0,1$ \\ 
under \$11k & $0,1$ \\ 
\$11k-\$25k  & $0,1$ \\ 
\$25k-\$50k & $0,1$ \\ 
>\$50k & $0,1$ \\ 
white  & $0,1$ \\ 
black & $0,1$ \\ 
asian & $0,1$ \\ 
hispanic & $0,1$ \\ 
yeas of education & $0-31$ \\ 
avg TISS & $1-83$ \\ 
diabetes & $0,1$ \\ 
dementia & $0,1$ \\ 
mean bp & $0-195$ \\ 
white blood cell count & $0-200$ \\ 
heart rate & $0-300$ \\ 
respiratory rate & $0-90$ \\ 
temperature & $31-42$ \\ 
pafi & $12-890$ \\ 
albumnin & $0.39-29$ \\ 
bilirubin & $0.09-63$ \\ 
creatinine & $0.09-63$ \\ 
sodium & $6-181$ \\ 
ph & $6.8-7.81$ \\ 
glucose & $0-1092$ \\
bun & $1-300$ \\
urine output & $0-9000$ \\
adlp & $0-7$ \\
adls & $0-7$ \\
\bottomrule
\end{tabular}}
\vspace{.5cm}
\label{tab:support_features}
\end{table}

\subsection{Additional experiment details}
We note that all experiments were performed using a single Nvidia Tesla P100 GPU.

\subsubsection{Section 4.1. Robustness to variation experiment details.}
We assess robustness to variation across different models and/or parameterizations. The rationale for this is that these are model changes or perturbations that different practitioners might easily make. However, we desire consistency and stability in the metrics such that the data insights remain consistent. For layers we evaluate a: 3 layer MLP, 4 layer MLP and 5 layer MLP. For number of hidden units we define two types: type 1 has 64 units in layer one, which reduce by half per layer and type 2 has 256 units in layer one, which reduce by half per layer. Finally, we evaluate these combinations with both Adam as the optimizer during training. These models perform at a similar level (i.e. accuracy). We then assess the consistency of the metrics for all examples using these different models/parameterizations based on Spearman rank correlation.

\subsubsection{Section 4.1. Data insights from subgroups experiment details.}
For the data insights experiment, we perform clustering on each subgroup identified by Data-IQ. We make use of Gaussian Mixture Models (GMM), similar to \cite{sohoni2020no}. The difference is the sub-space in which the clustering is applied. In Data-IQs case, we cluster within each subgroup identified by Data-IQ. For each supergroup, we search over $k \in 2,\ldots , 10$. We choose the $k$ value that yields the highest average Silhouette score. Our implementation is based on \footnote{https://scikit-learn.org}

\subsubsection{Section 4.2. Principled feature acquisition experiment details.}
We assess the potential for principled feature acquisition using a semi-synthetic experiment. We assess how valuable each feature is based on the Pearson correlation of the feature with the target. We then rank sort the correlations from lowest to highest. This ordering is then used to determine the order in which to ``acquire'' and assess the features.

\subsubsection{Section 4.2. Principled dataset comparison experiment details.}
As described in the main paperm we compare two synthetic datasets using two models representing two synthetic data vendors: namely CTGAN and Gaussian Copula. We use the implementations of \footnote{https://github.com/sdv-dev/SDV}.

\subsubsection{Section 4.3. Less is more experiment details.}
For this experiment we note that we train multiple baseline models to evaluate the generalization based on dataset sculpting. We evaluated three different XGBoost models with 100,150 and 200 estimators each. Our XGBoost implementation uses the Python version from \footnote{https://github.com/dmlc/xgboost}.

\subsubsection{Section 4.3. Group-DRO experiment details.}
Group DRO uses training group annotations to directly minimize the worst-group error on the training set. Hence, we compare different methods to obtain the group annotations. This experiment compares Group-DRO applied to groups identified by Data-IQ and as a baseline George \cite{sohoni2020no}. We also compare this to Just-Train-Twice (JTT) \cite{liu2021just}. We use the implementation of Group-DRO from \footnote{https://github.com/HazyResearch/hidden-stratification}, which is based on \cite{sagawa2019distributionally}. Our George benchmark is based on the implementation from \footnote{https://github.com/HazyResearch/hidden-stratification} and the JTT benchmark is based on the implementation from \footnote{https://github.com/anniesch/jtt} . All code is in Pytorch.

The group DRO objective can then be written as: 
\begin{align}\label{eqn:gdro_objective}
    L_{\text{group-DRO}}(\theta) = \max_{g \in G} 
    \frac{1}{n_g} \sum_{i \mid g_i = g} 
    \ell(x_i, y_i; \theta)
\end{align}
where $n_g$ is the number of training points with group $g_i = g$.

\subsubsection{Section 4.3. Subgroup informed usage of uncertainty estimation experiment details.}
As mentioned in the main paper we obtain uncertainty estimates using a Bayesian Neural Network based on \cite{ghosh2018structured}. Specifically, we train a 5-layer MLP model. A Gaussian prior is placed over the weights and we optimize the KL divergence during training. Our benchmark is based on \citep{ghosh2018structured} and we use the implementation from \footnote{https://github.com/IBM/UQ360}.
\clearpage

\section{Additional Experiments} \label{sec:appendixC}
\subsection{Understanding the representation space}

\paragraph{Goal.} At deployment/testing time, we stratify the examples into subgroups using the representation space. of the training examples. Our goal is to illustrate the points made in Section \ref{formulation:stratification} with respect to the representation space

\begin{figure}[!h]
    \centering
    \includegraphics[width=0.75\textwidth]{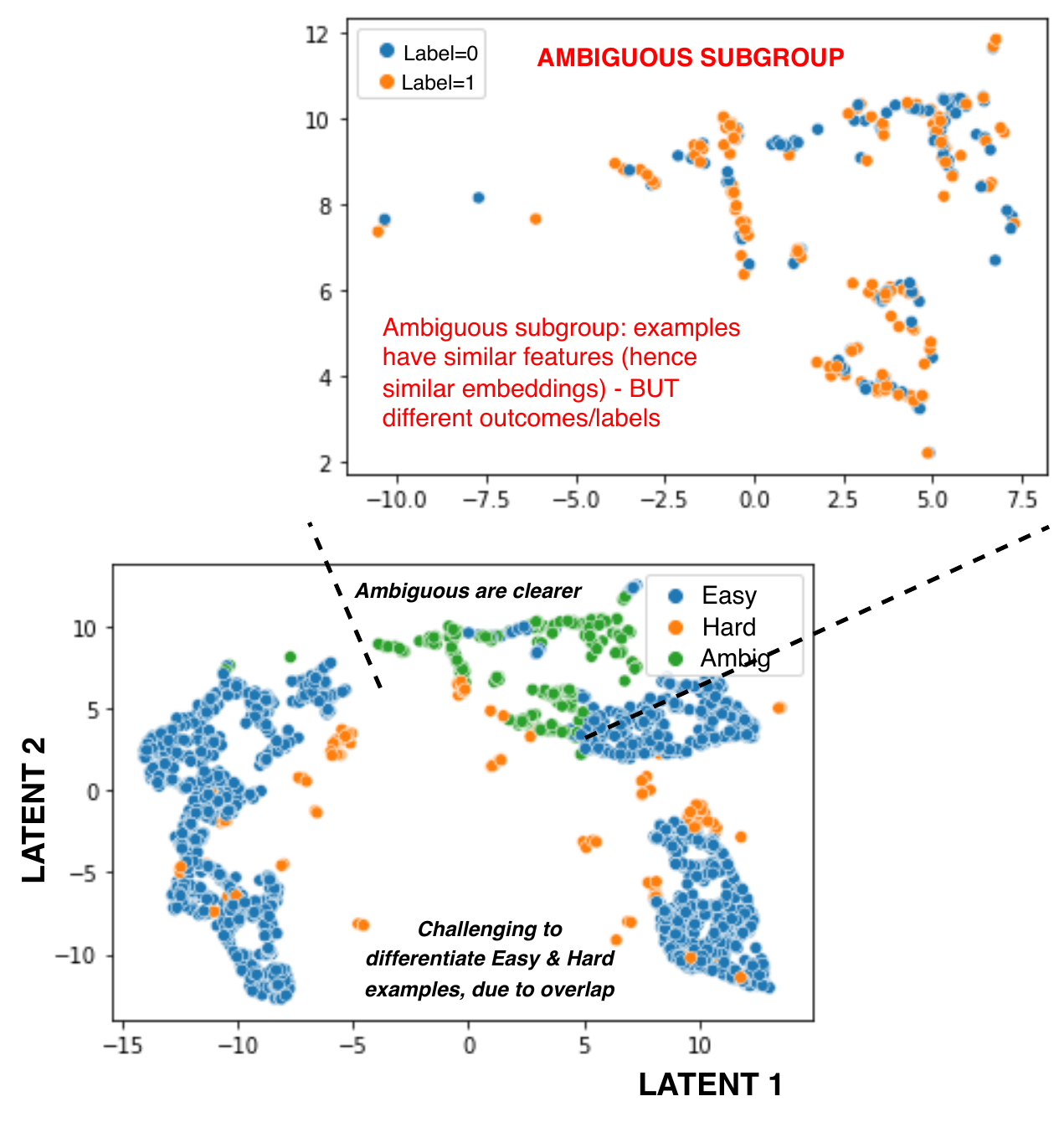}
    \caption{Representation space of training examples broken down by sub-group. We then zoom into the Ambiguous subgroup and illustrate the heterogeneous outcomes, where examples with similar features can have different outcomes.}
    \label{fig:embedding_plot}
\end{figure}

\paragraph{Takeaway.}
We recall the two main points about the representation space made in the main paper and illustrate them using the representation space (i.e. UMAP embedding) shown in Figure \ref{fig:embedding_plot}.
\begin{enumerate}
    \item $\ambiguous$ examples have distinctive features and are clustered in embedding space. Thus, we can distinguish the $\ambiguous$ examples using the embedding. We can clearly see this with the green points in Figure \ref{fig:embedding_plot}.  
    \item It is not possible to reliably distinguish $\easy$ examples from $\hard$ examples based on the embedding, because $\hard$ examples are a minority with outcome randomness that have similar features, as the $\easy$ examples. Naturally, since the Easy and Hard example embeddings are similar, they also have similar features.  We can clearly see this where the orange examples (Hard) are randomly scattered with the Blue (Easy) examples, in Figure \ref{fig:embedding_plot}.  This makes these two groups difficult to distinguish.
\end{enumerate}

We also zoom in to the $\ambiguous$ region (sub-figure of Figure \ref{fig:embedding_plot}). This illustrates the specific issue of outcome heterogeneity, where similar patients (with similar features) have different outcomes. This is illustrated by the different colors, which represent the different outcomes (i.e. label=0 is survive and label=1 is death).

\subsection{Data-IQ: Neural Networks vs Other model classes}
\paragraph{Goal.} As discussed in the main text and Appendix \ref{adapt}, Data-IQ can be used with \emph{any} ML model trained in stages. Methods such XGBoost, LightGBM and CatBoost methods are widely used by practitioners on tabular data, often more so than neural networks \cite{borisov2021deep}.  While Data-IQ can naturally handle any such model, as long as there are checkpoints (over iterations), Data Maps naturally cannot (or at least it was never formulated to do so). However, to enable comparison, we can use Data Maps within our framework, so that it can be applied to a more general class of models.

Ideally, based on \emph{P1}, we desire that the characterization of examples be consistent for similar performing models, irrespective of whether the model is a neural network or, for example, an XGBoost model.

\begin{figure*}[!h]

  \centering 
  \subfigure[Data-IQ ]{\includegraphics[width=0.40\textwidth]{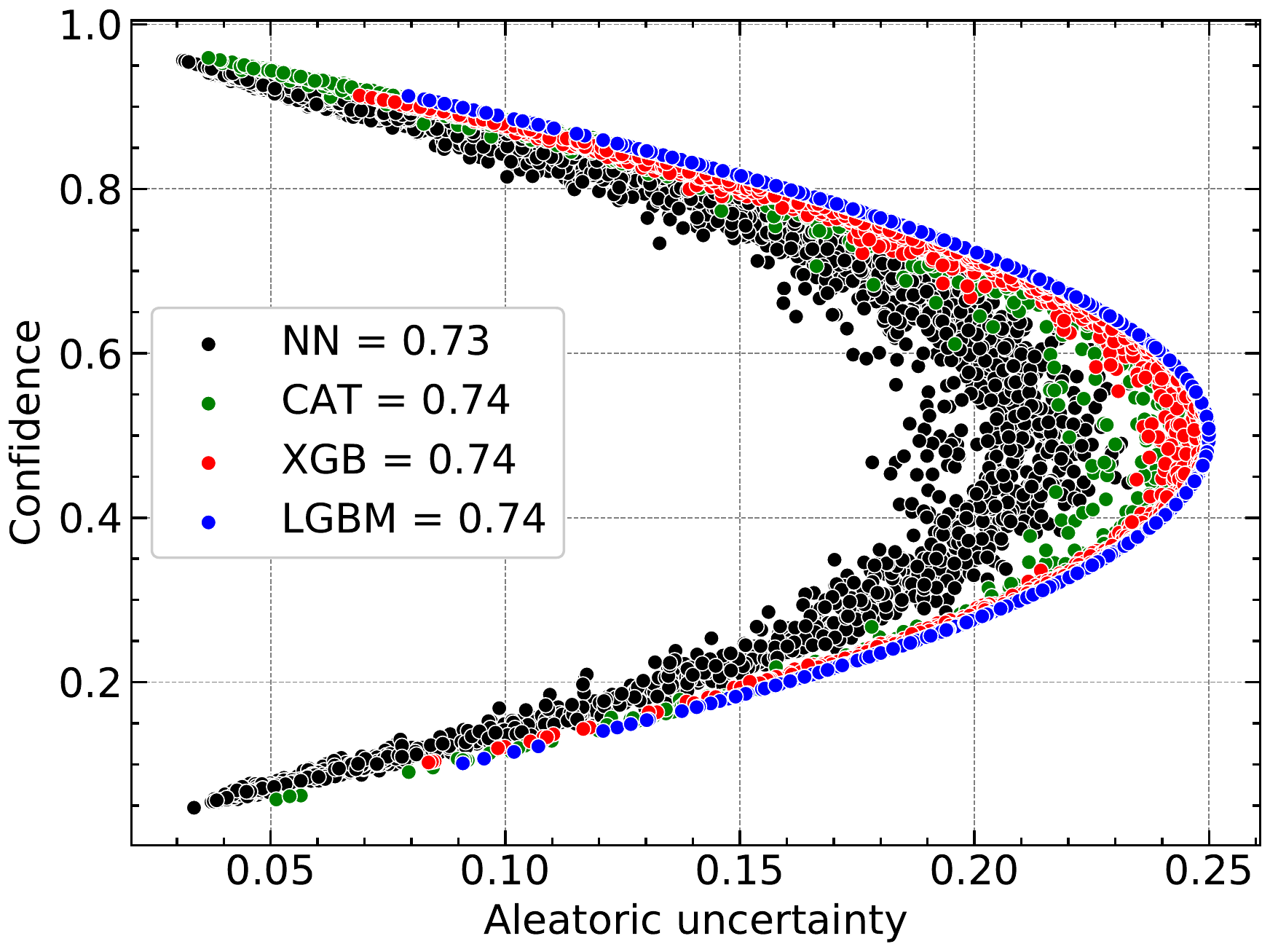}}\quad\quad
  \subfigure[Data Maps]{\includegraphics[width=0.40\textwidth]{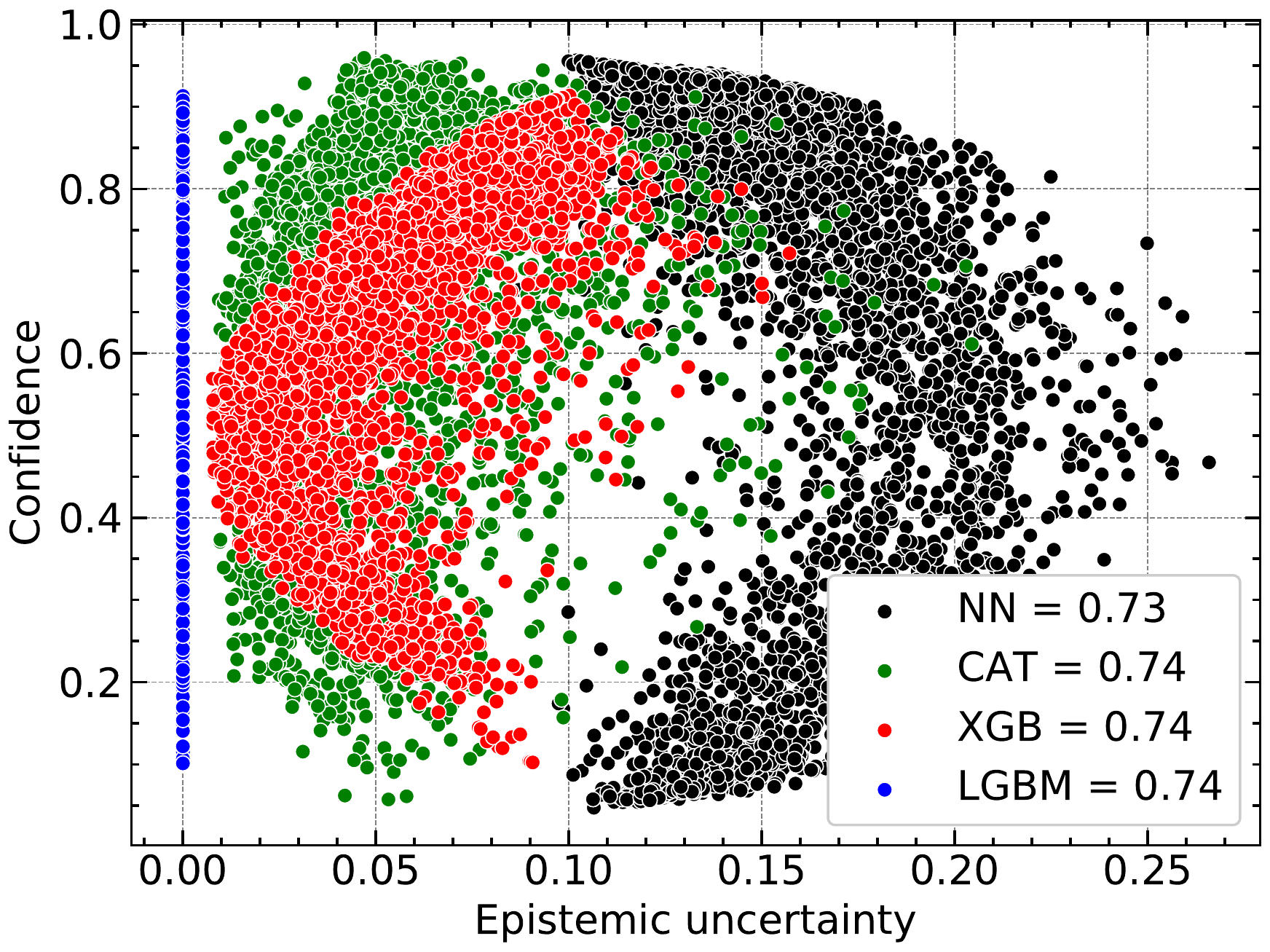}}
  \caption{NN vs XGBoost comparison on the Covid-19 dataset, showing Data-IQ is more consistent}
  \label{fig:covid_nnxgb}
\end{figure*}

\begin{figure*}[!h]

  \centering 
  \subfigure[Data-IQ ]{\includegraphics[width=0.40\textwidth]{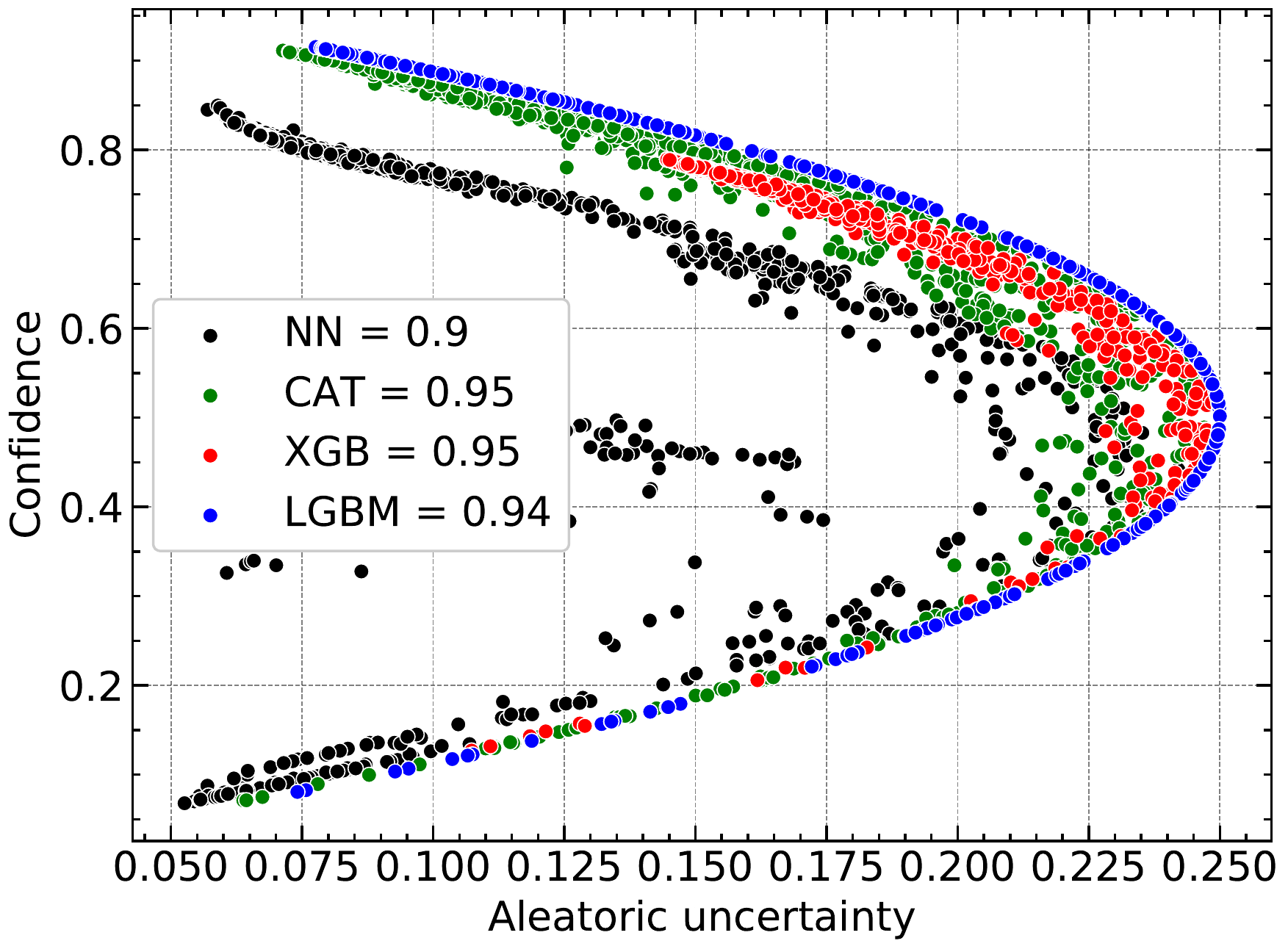}}\quad\quad
  \subfigure[Data Maps]{\includegraphics[width=0.40\textwidth]{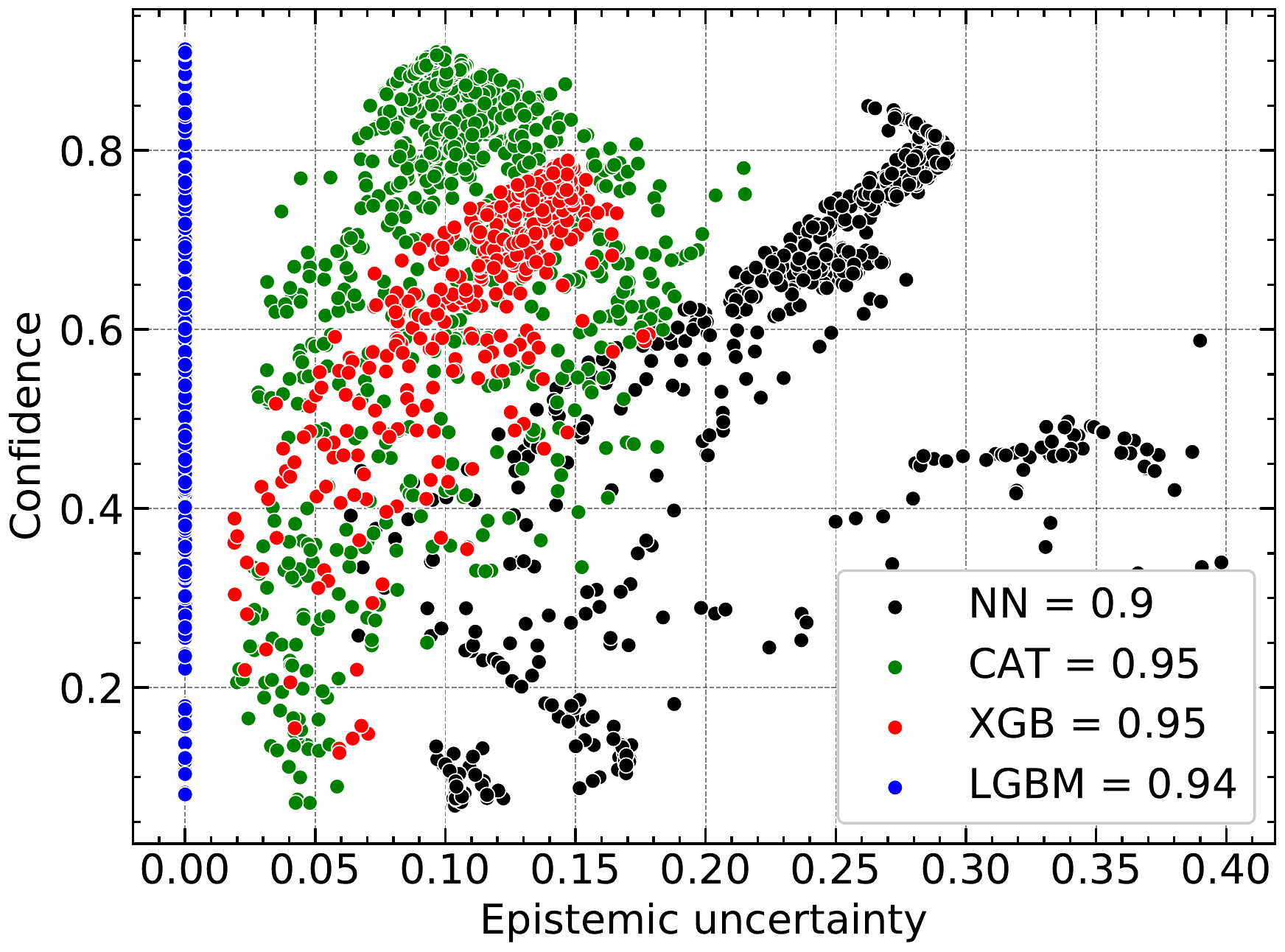}}
  \caption{NN vs XGBoost comparison on the Fetal dataset, showing Data-IQ is more consistent}
  \label{fig:fetal_nnxgb}
\end{figure*}

\begin{figure*}[!h]

  \centering 
  \subfigure[Data-IQ]{\includegraphics[width=0.40\textwidth]{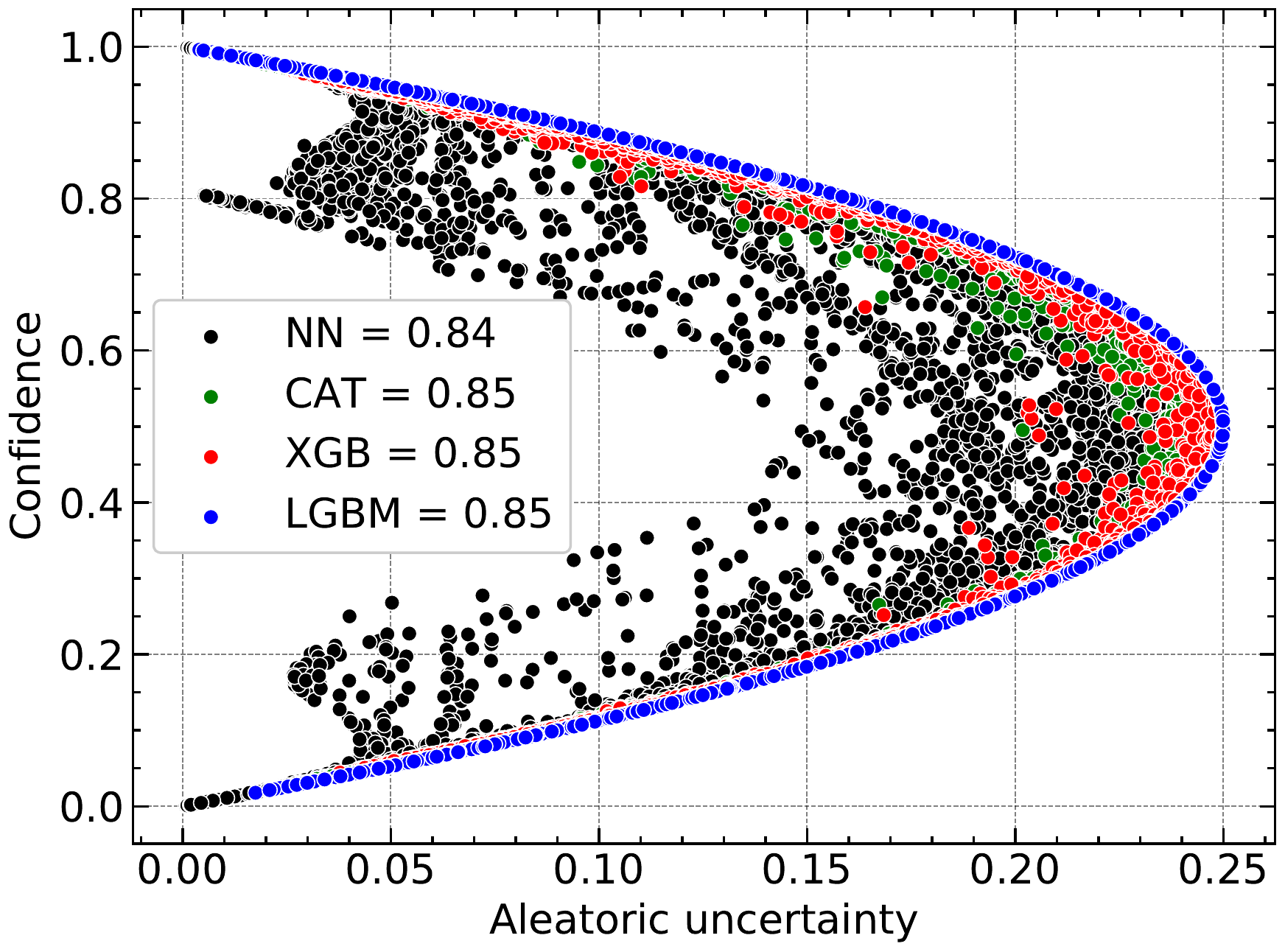}}\quad\quad
  \subfigure[Data Maps]{\includegraphics[width=0.40\textwidth]{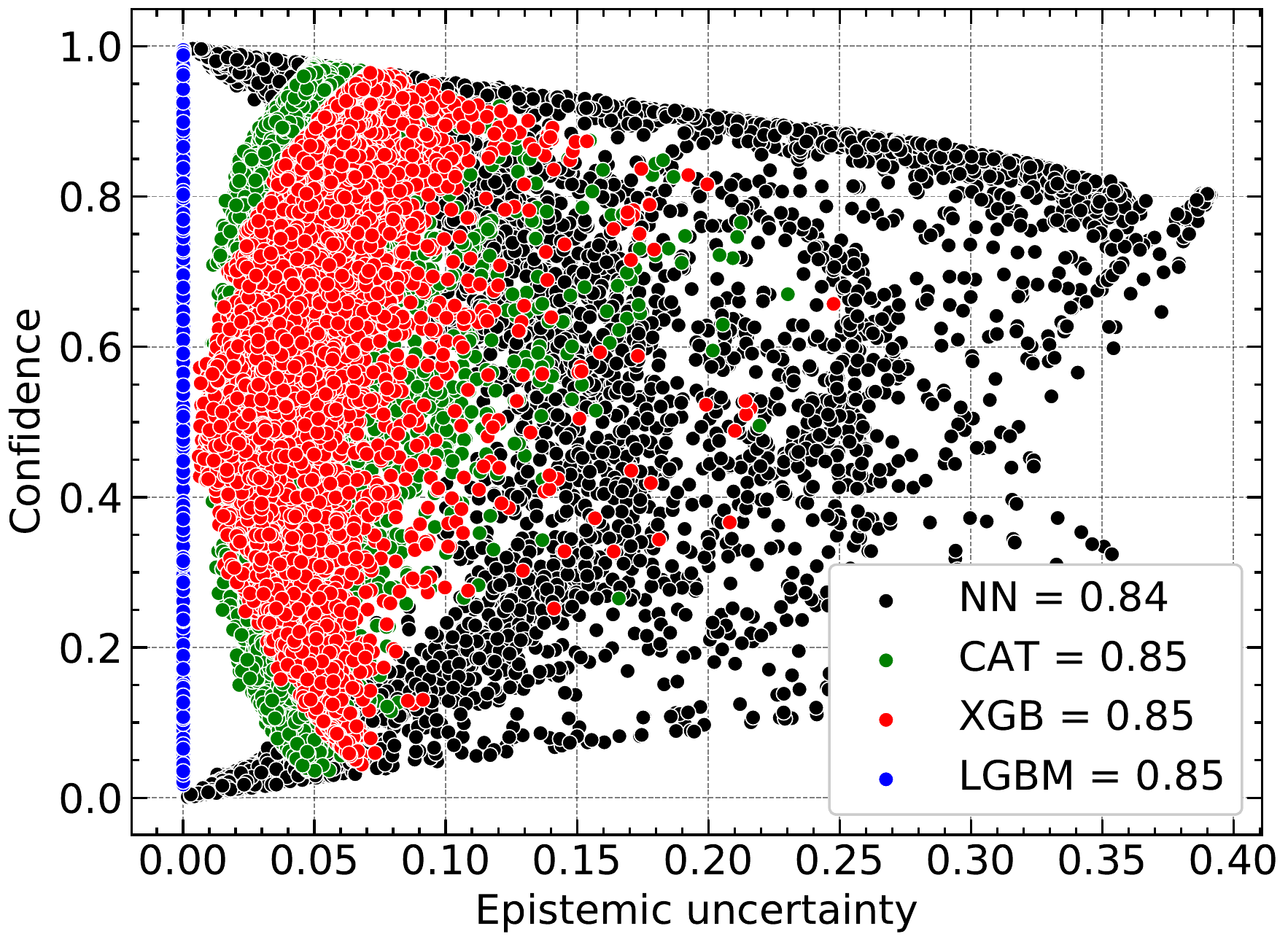}}
  \caption{NN vs XGBoost comparison on the Prostate dataset, showing Data-IQ is more consistent}
  \label{fig:prostate_nnxgb}
\end{figure*}

\begin{figure*}[!h]

  \centering 
  \subfigure[Data-IQ ]{\includegraphics[width=0.4\textwidth]{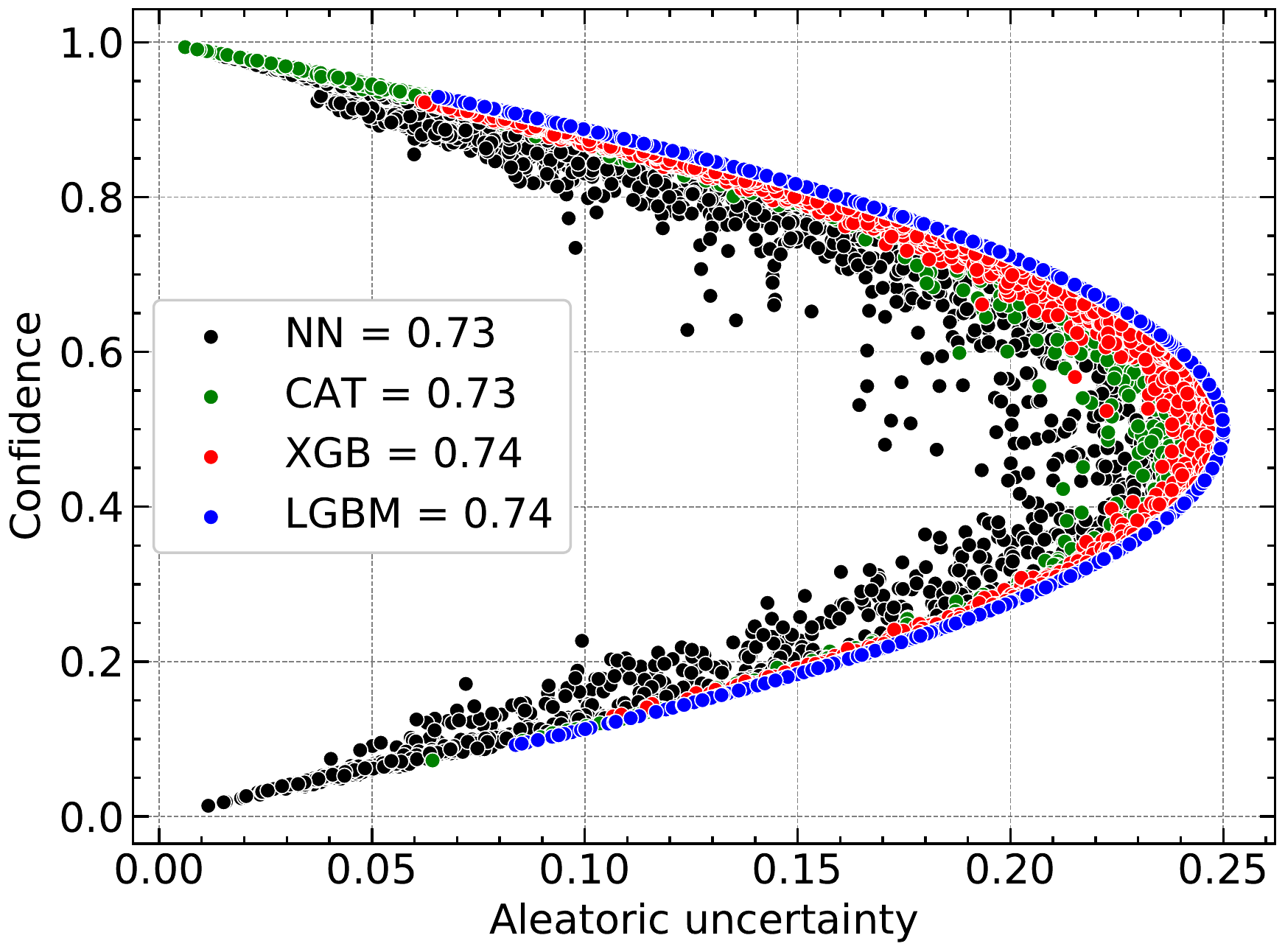}}\quad\quad
  \subfigure[Data Maps]{\includegraphics[width=0.4\textwidth]{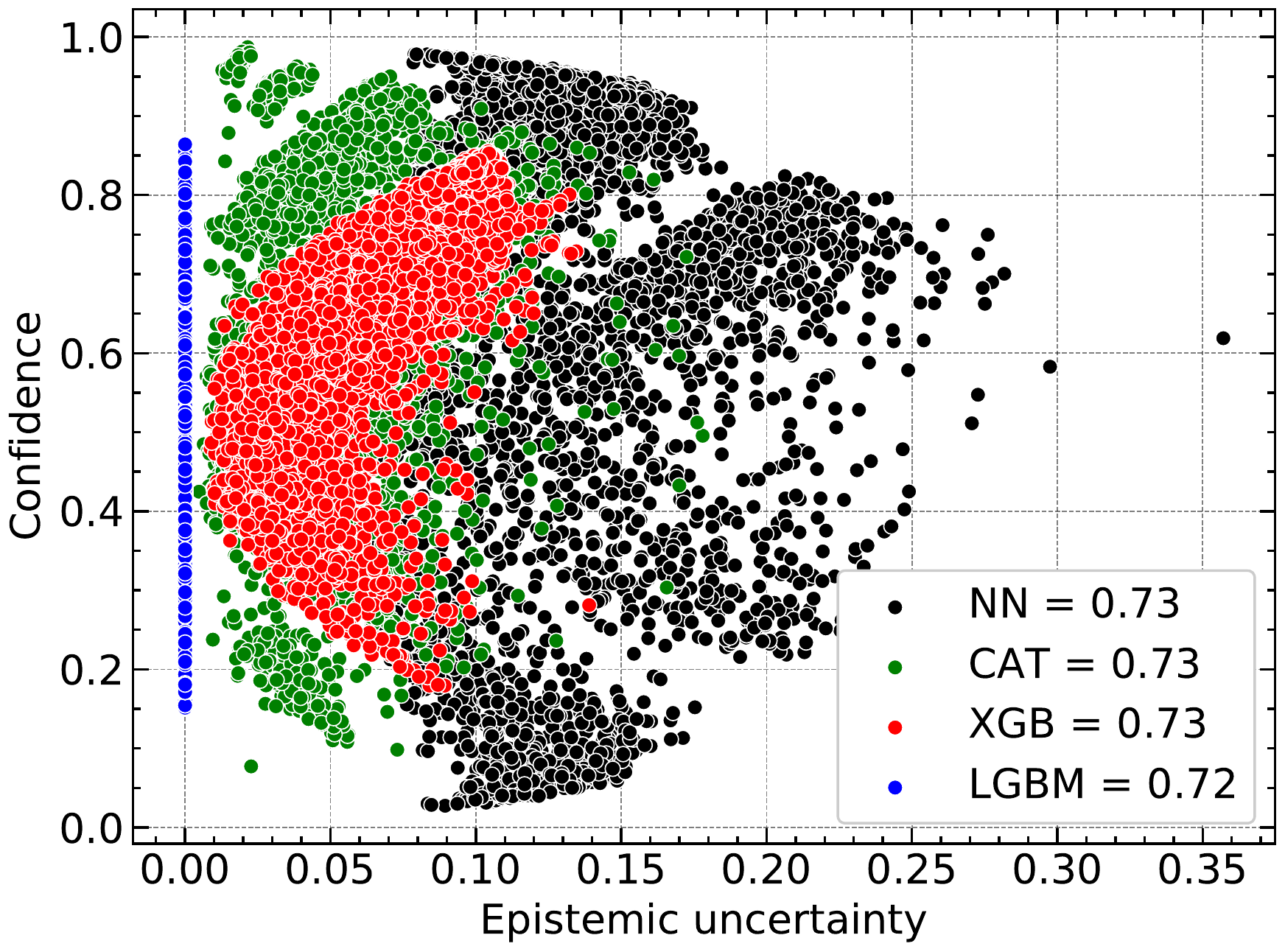}}
  \caption{NN vs XGBoost comparison on the Support dataset, showing Data-IQ is more consistent}
  \label{fig:support_nnxgb}
\end{figure*}

To assess this the robustness of  both Data-IQ and Data Maps, we train a neural network, XGBoost, LightGBM and CatBoost models to achieve the same performance and then perform the characterization for all models. The evaluation of all data sets is shown in Figures \ref{fig:covid_nnxgb}-\ref{fig:support_nnxgb}.

\paragraph{Takeaway.} We can clearly see that Data-IQ has a similar characterization for across all four models. Contrastingly, for Data Maps, the characterizations are significantly different for the different model classes. The key difference is Data-IQs usage of Aleatoric uncertainty, rather than the Data Maps usage of Epistemic uncertainty. The implication of this result is that by Data-IQ capturing the uncertainty inherent to the data (aleatoric uncertainty), it leads to a more consistent and stable characterizations of the data itself. Especially, this highlights that Data-IQ characterizes the data in a manner that is not as sensitive to the choice of model when compared to Data Maps.

\newpage
To further augment these results, we quantitatively assess the consistency across model types. Similar to Section 4.1, we have compared the Spearman rank correlation between the metrics of all model combinations (neural networks, XGBoost, LightGBM, CatBoost). Note that all four models are trained to perform similarly on a held-out dataset. 

Below, we present Table \ref{tab:robustness_models} showing the average and standard deviation. We see that Data-IQ has higher scores for all datasets, highlighting the better consistency compared to Data Maps.

\begin{table}[!h]
 \centering
\caption{Comparison of robustness/consistency across different models on the basis of the Spearman rank correlation}
\scalebox{1}{
\begin{tabular}{ccc}
\toprule
Dataset &  (Ours) Data-IQ & Data Maps    \\ \hline
\midrule
Covid & \bf $0.85 \pm 0.05$  & $0.53 \pm 0.07$\\ \hline
Support & \bf $0.84 \pm 0.06$  & $0.48 \pm 0.04$\\ \hline
Prostate & \bf $0.84 \pm 0.11$  &$0.44 \pm 0.17$ \\ \hline
Fetal & \bf $0.72 \pm 0.07$  & $0.45 \pm 0.06$
\\ \hline
\bottomrule
\end{tabular}}
\label{tab:robustness_models}
\end{table}

\newpage

\subsection{Collecting more data - it might make things worse}

\paragraph{Goal.} Typically it is assumed that simply by collecting more data that most problems will be mitigated. We aim to assess the influence of adding more data samples, in contrast to previous experiments in the main paper on acquiring features. Especially, we want to show how, in the case of heterogeneous outcomes, \textit{blindly} adding more data might not always be the solution.

\begin{figure*}[!h]

  \centering
  \subfigure[covid more samples ]{\includegraphics[width=0.4\textwidth]{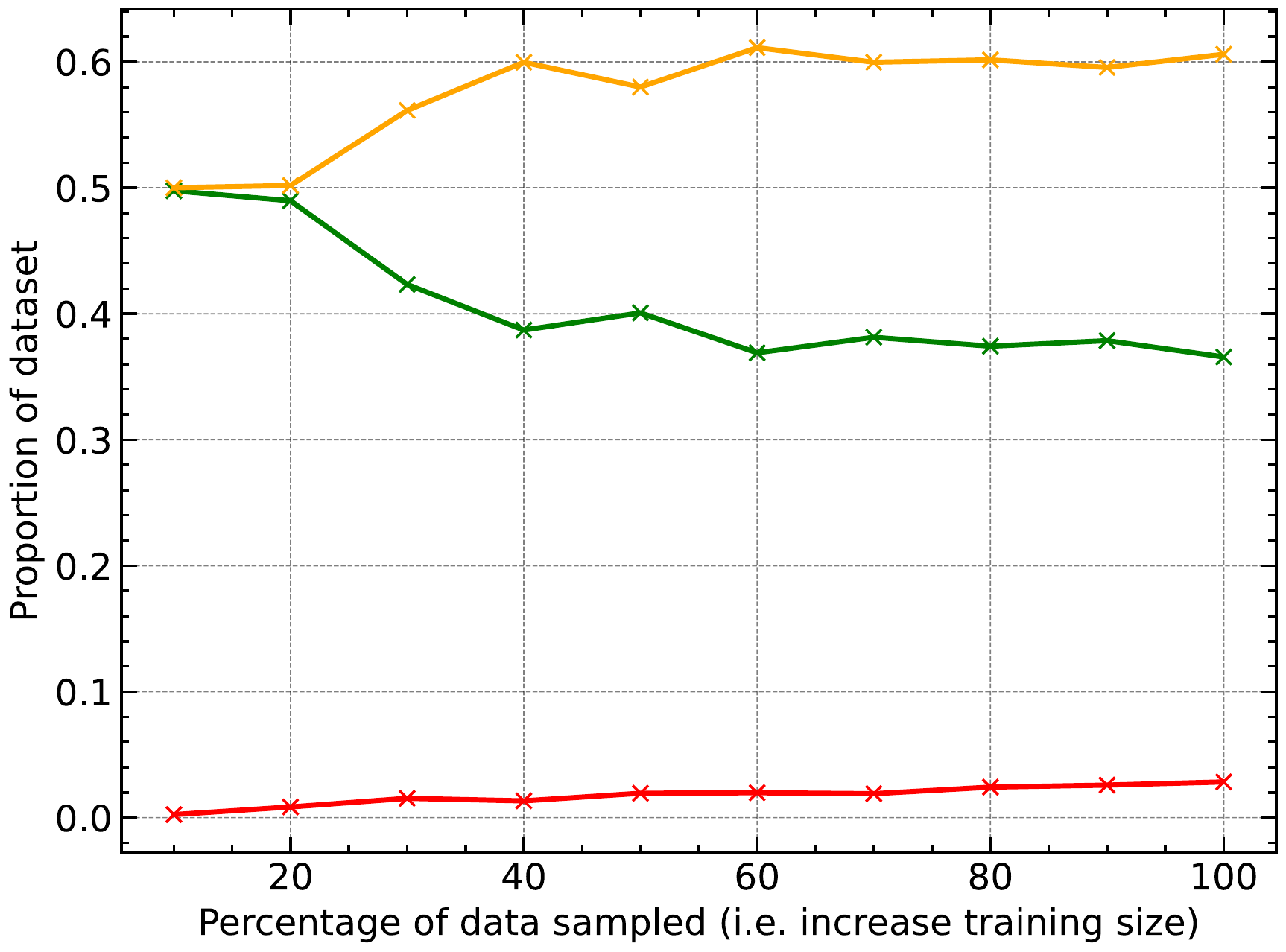}}\quad\quad
  \subfigure[support more samples]{\includegraphics[width=0.4\textwidth]{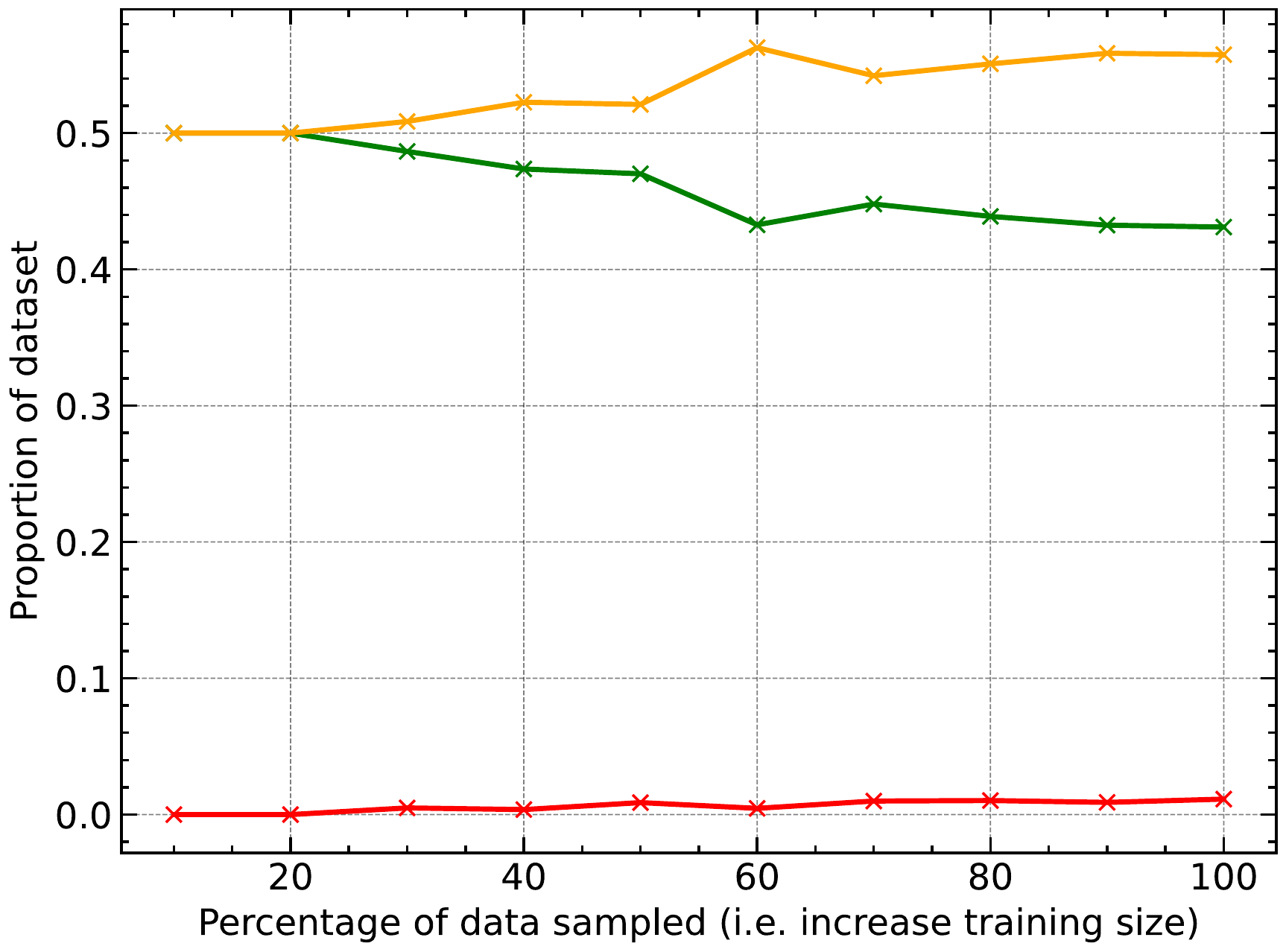}}
  \caption{The $\ambiguous$ subgroup proportion in fact increases as the dataset size increases, i.e. more examples are added. This is due to the increased probability of feature collisions.}
  \label{fig:more_data}
\end{figure*}

\paragraph{Experiment.} We sub-sample proportions of the dataset from 0-100\% of the original size of the dataset. We then recompute the Data-IQ subgroups on each proportion, as more data examples are added. We show two examples in Figure \ref{fig:more_data}, in which the subgroups evolve as more data is added. 

\paragraph{Takeaway.} As the number of data examples increases, the proportion covered by the $\ambiguous$ subgroup increases. This is due to the increased probability of ``feature collisions'' as more examples are added. Therefore, the result shows that the simple solution, in which we naively collect more data, can sometimes do more harm.

\subsection{Identifying subsets in an outcome driven manner: assessing quality}

\paragraph{Goal.} Data-IQ identifies subgroups in an outcome-driven manner, compared to the conventional unsupervised manner. We compare the cluster quality on the outcome-driven subgroups as obtained by Data-IQ compared to other benchmarks, namely: George \cite{sohoni2020no}, clustering in the latent space, clustering in a PCA space, and clustering the raw feature space. All methods use a Gaussian Mixture model for consistency.

We evaluated these methods based on two well-known cluster quality metrics, namely, (1) Silhouette score (SIL) and (2) Davies-Bouldin (DB) score. The results for two example datasets can be seen in Table \ref{tab:cluster}.
\begin{table}[!h]
    \centering
        \caption{Clustering results}
    \scalebox{0.85}{
    \begin{tabular}{lcccccc}
\toprule
Dataset &      Metric  & Data-IQ (Ours) &  Outcome latent space (George)  & Latent space  &      PCA  & Feature space   \\
\midrule
Covid-19 & SIL ($\uparrow$) &  0.68 & 0.48 & 0.47 & 0.51 & 0.02\\
 & DB ($\downarrow$) &   0.42 &  0.87 &  0.78 & 0.81 &  5.7\\
Prostate & SIL ($\uparrow$) &  0.80 & 0.6 & 0.56 & 0.57 & 0.54\\
 & DB ($\downarrow$) &   0.3 & 0.6 & 0.63 & 0.57 & 0.61
\\
\bottomrule
\end{tabular}}
    \label{tab:cluster}

\end{table}

\paragraph{Takeaway.} We can clearly see that Data-IQ has the highest SIL score and the lowest DB score. These results demonstrate that the outcome-driven clustering approach taken by Data-IQ, where the clusters are within the Data-IQ subgroups, results in higher quality clusters. 

\subsection{Data Insights: additional results}
\begin{figure*}[!h]
  \centering
  \subfigure[Easy ]{\includegraphics[width=0.255\textwidth]{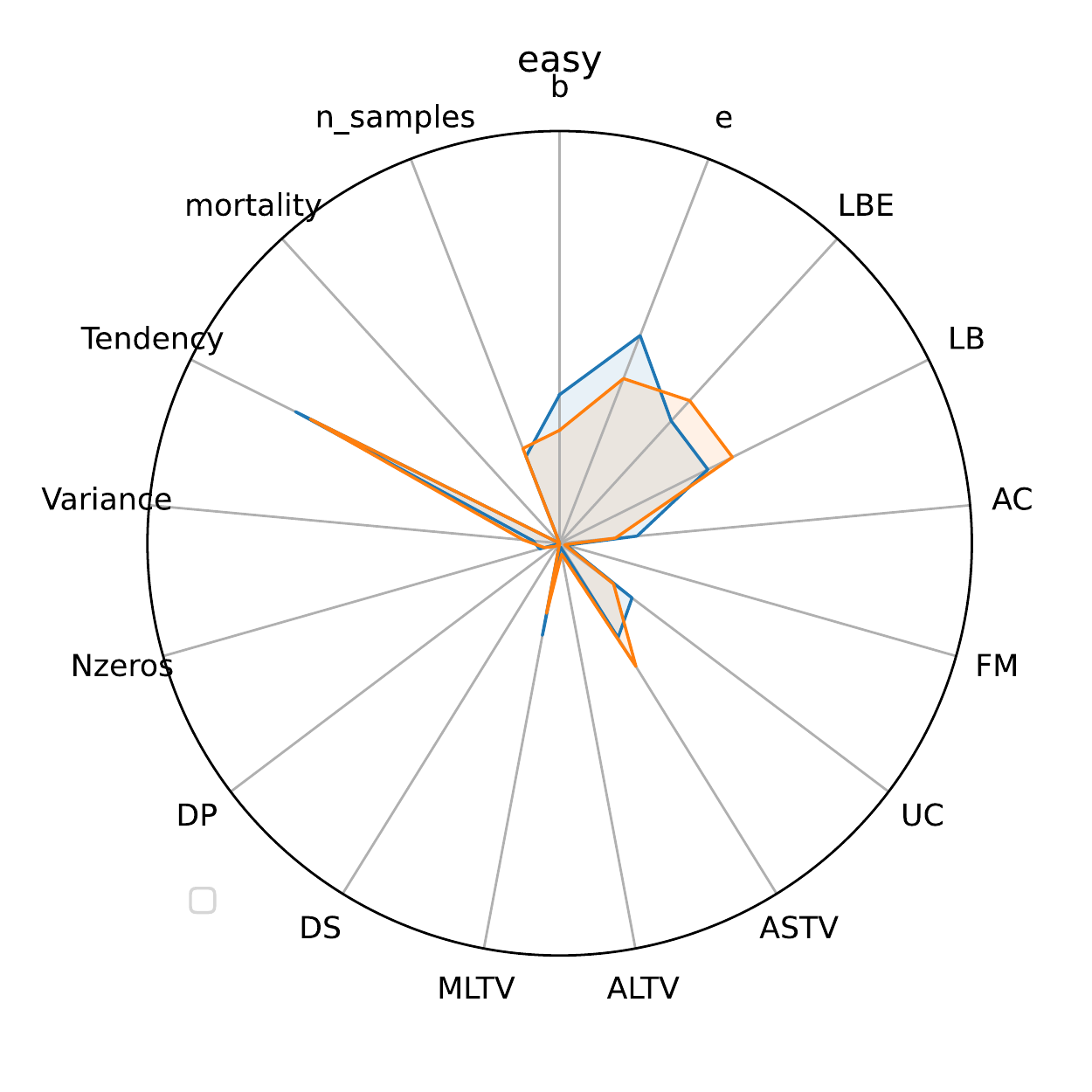}}\quad\quad
  \subfigure[Ambiguous ]{\includegraphics[width=0.255\textwidth]{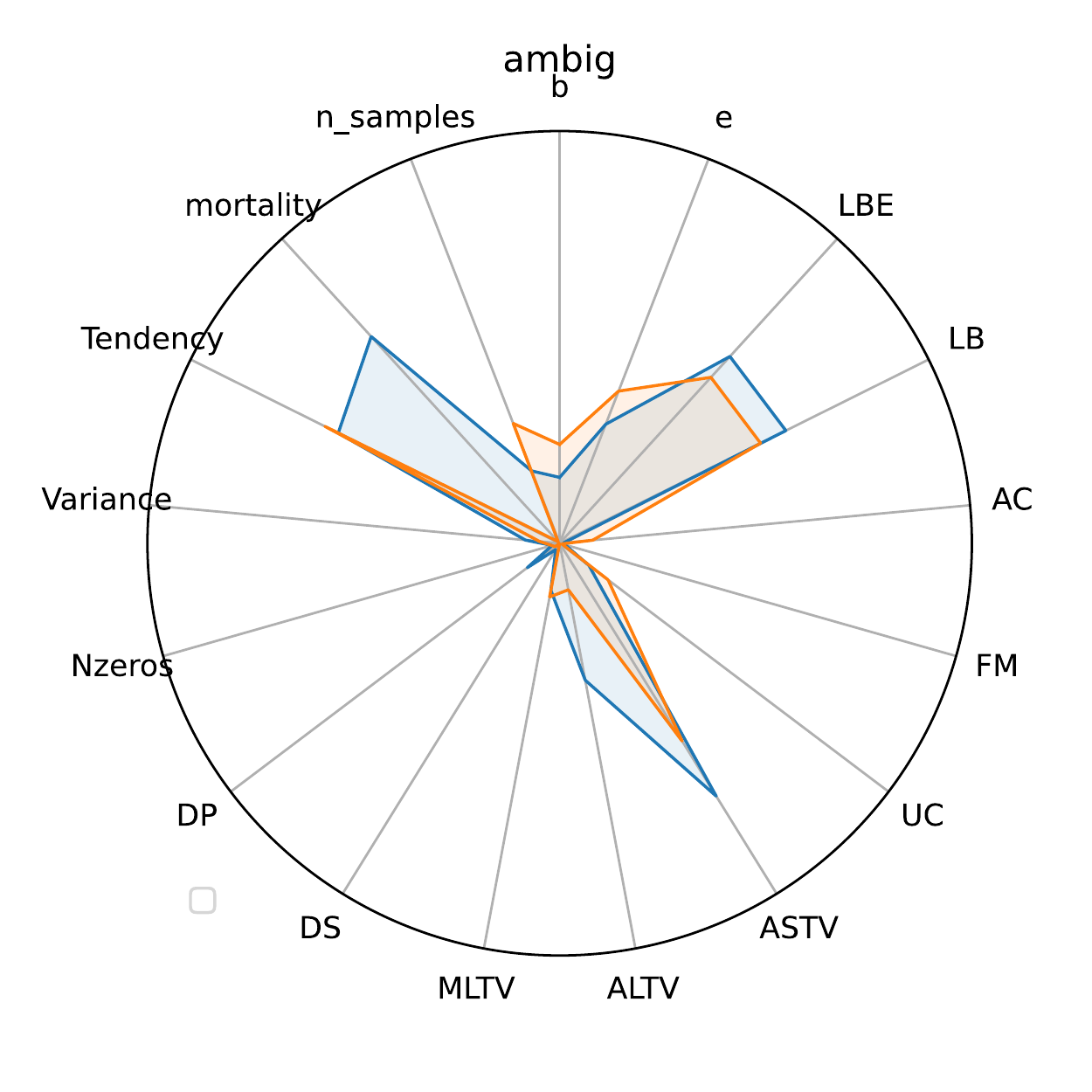}}\quad\quad
  \subfigure[Hard]{{\includegraphics[width=0.255\textwidth]{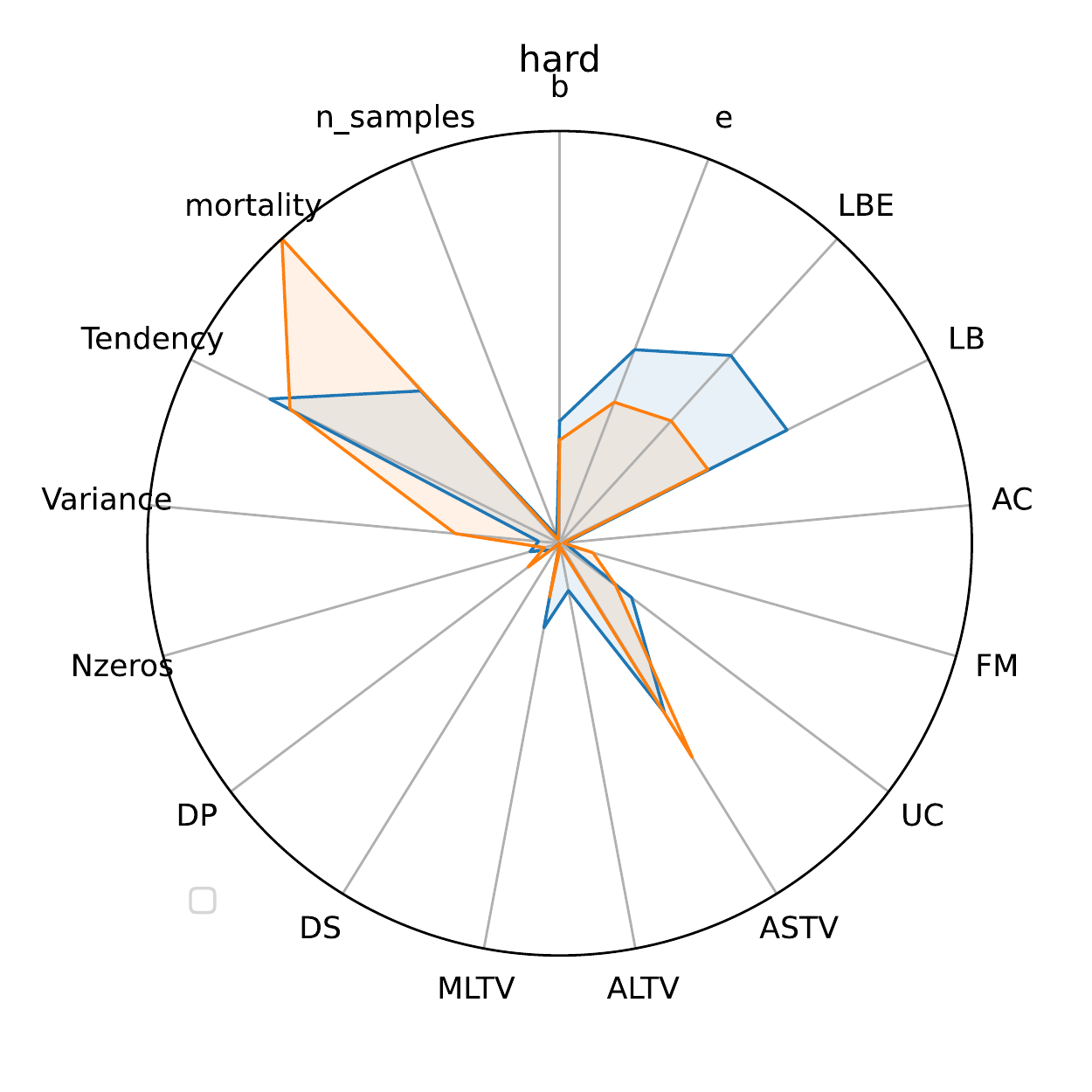}}}

  \caption{\footnotesize{Fetal subgroups identified by Data-IQ (descriptions above). Colors represent the GMM clusters}}
  \label{fig:fetal_radar}
\end{figure*}

\begin{figure*}[!h]
  \centering
  \subfigure[Easy ]{\includegraphics[width=0.255\textwidth]{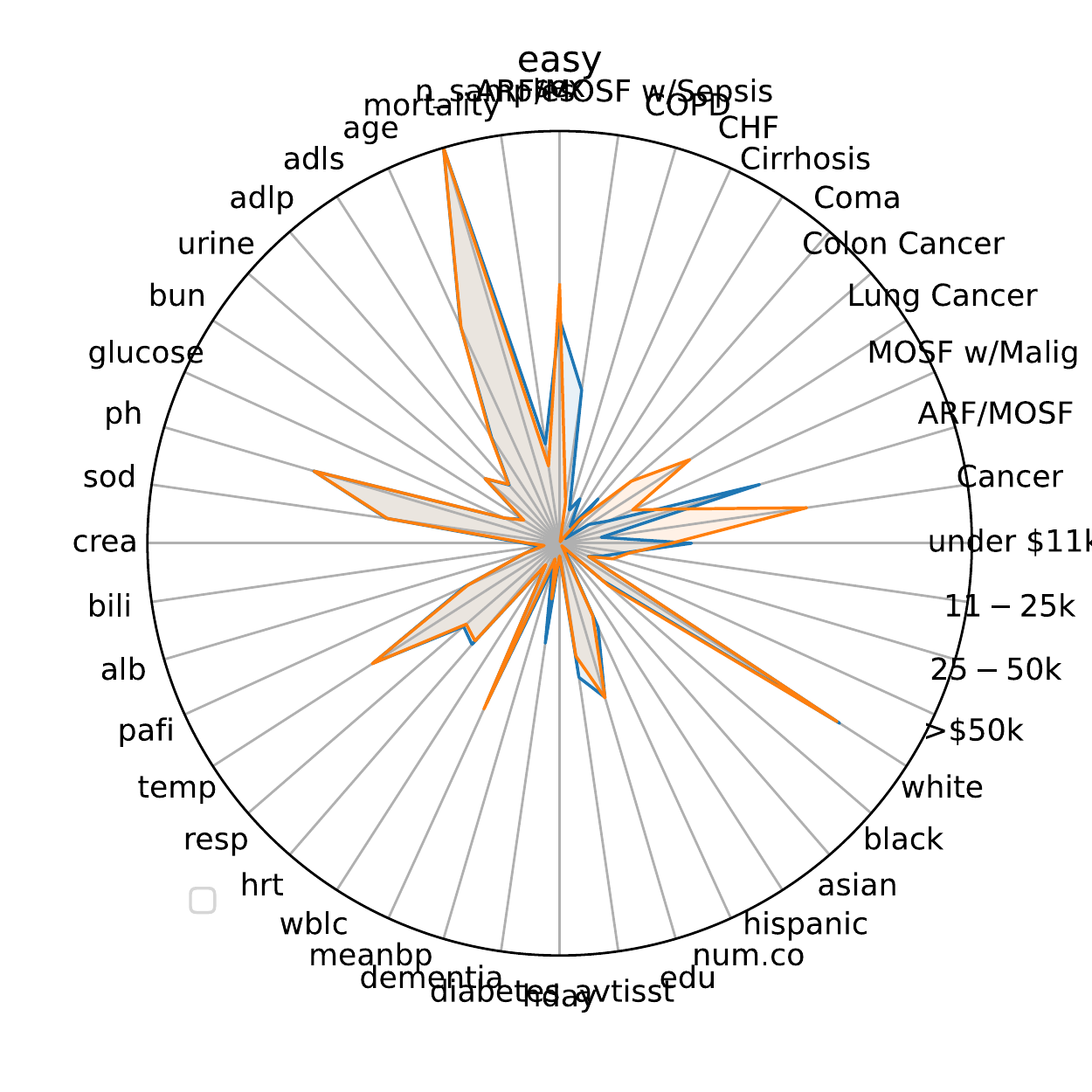}}\quad\quad
  \subfigure[Ambiguous ]{\includegraphics[width=0.255\textwidth]{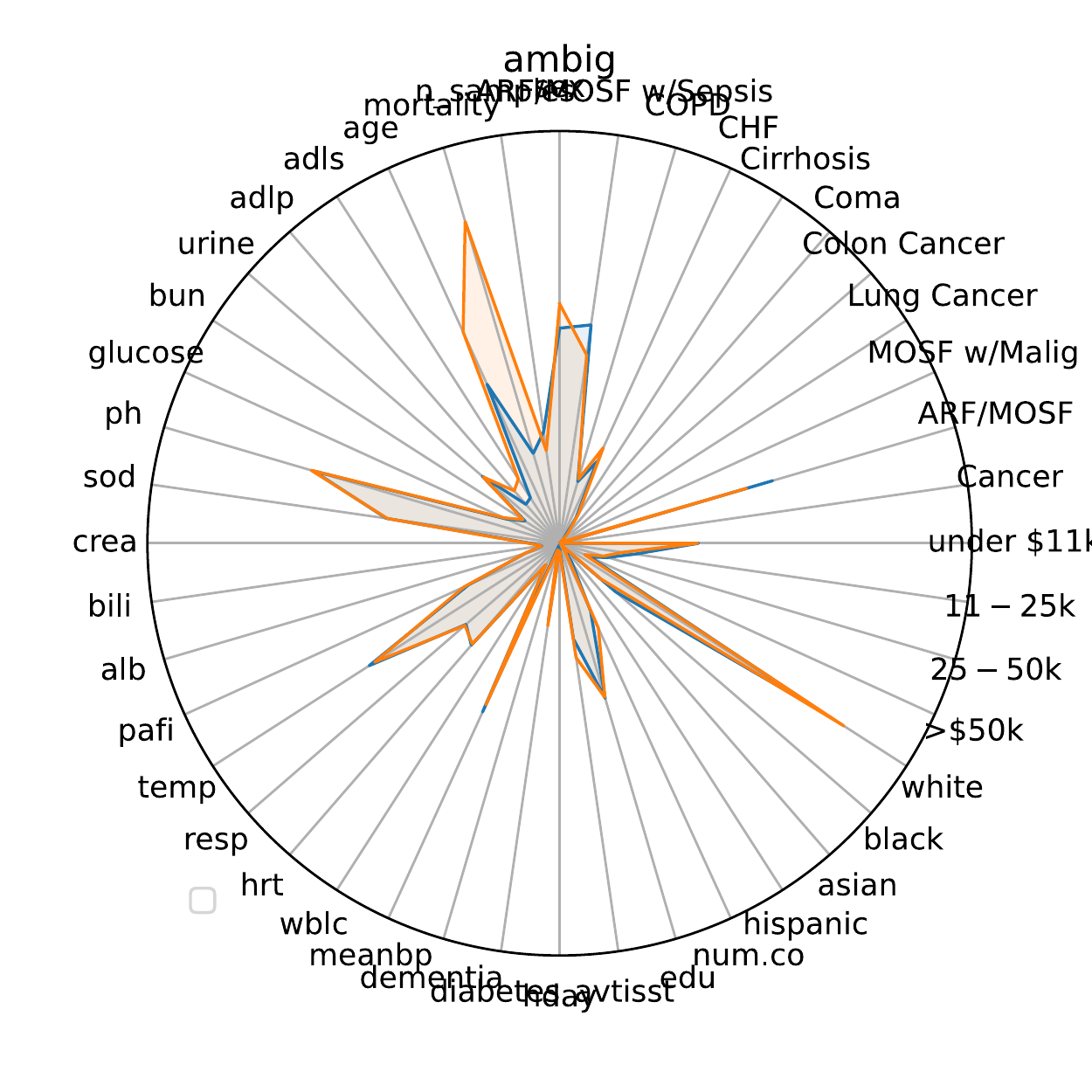}}\quad\quad
  \subfigure[Hard]{{\includegraphics[width=0.255\textwidth]{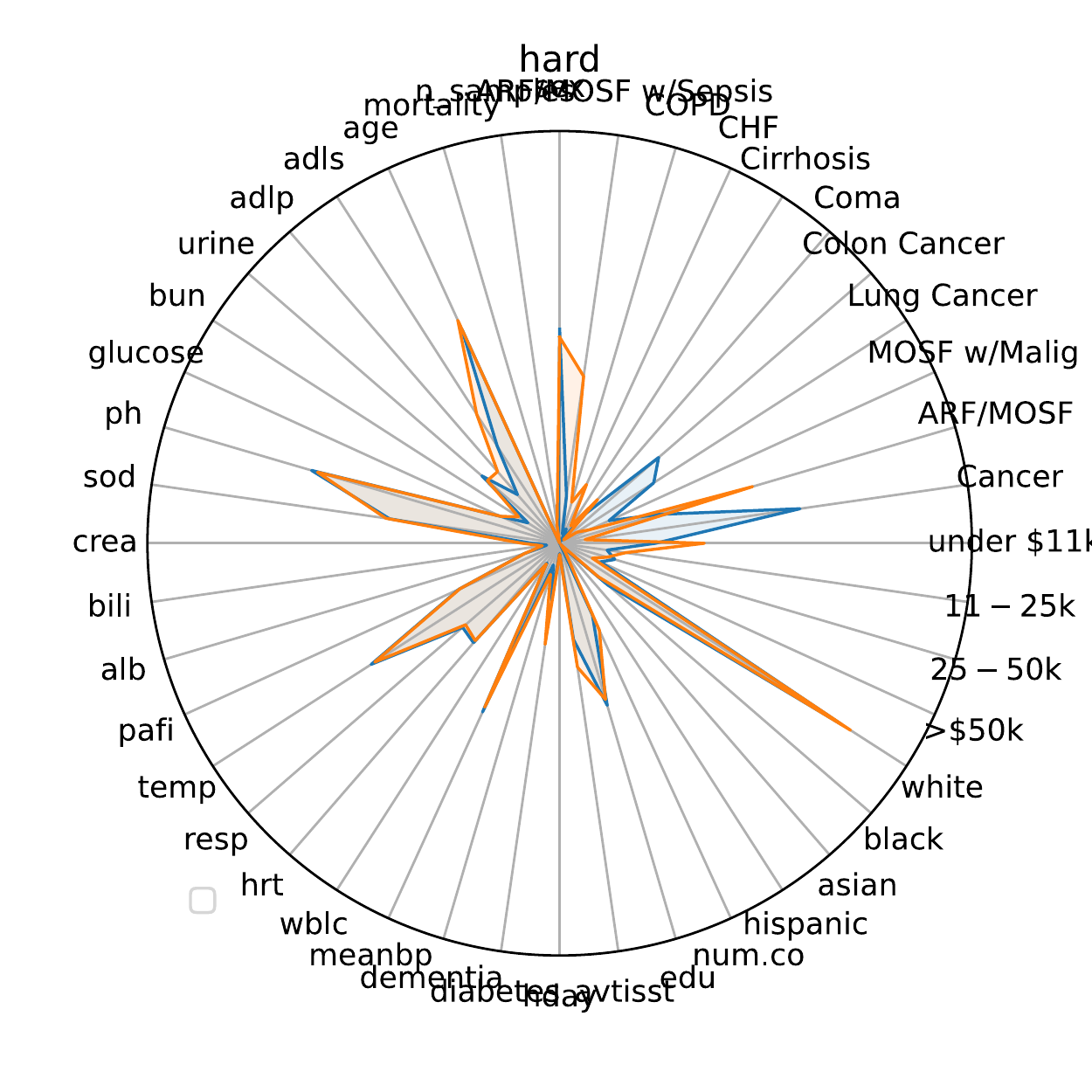}}}

  \caption{\footnotesize{Support subgroups identified by Data-IQ (descriptions above). Colors represent the GMM clusters}}
  \label{fig:support_radar}
\end{figure*}

\begin{figure*}[!h]
  \centering
  \subfigure[Easy ]{\includegraphics[width=0.255\textwidth]{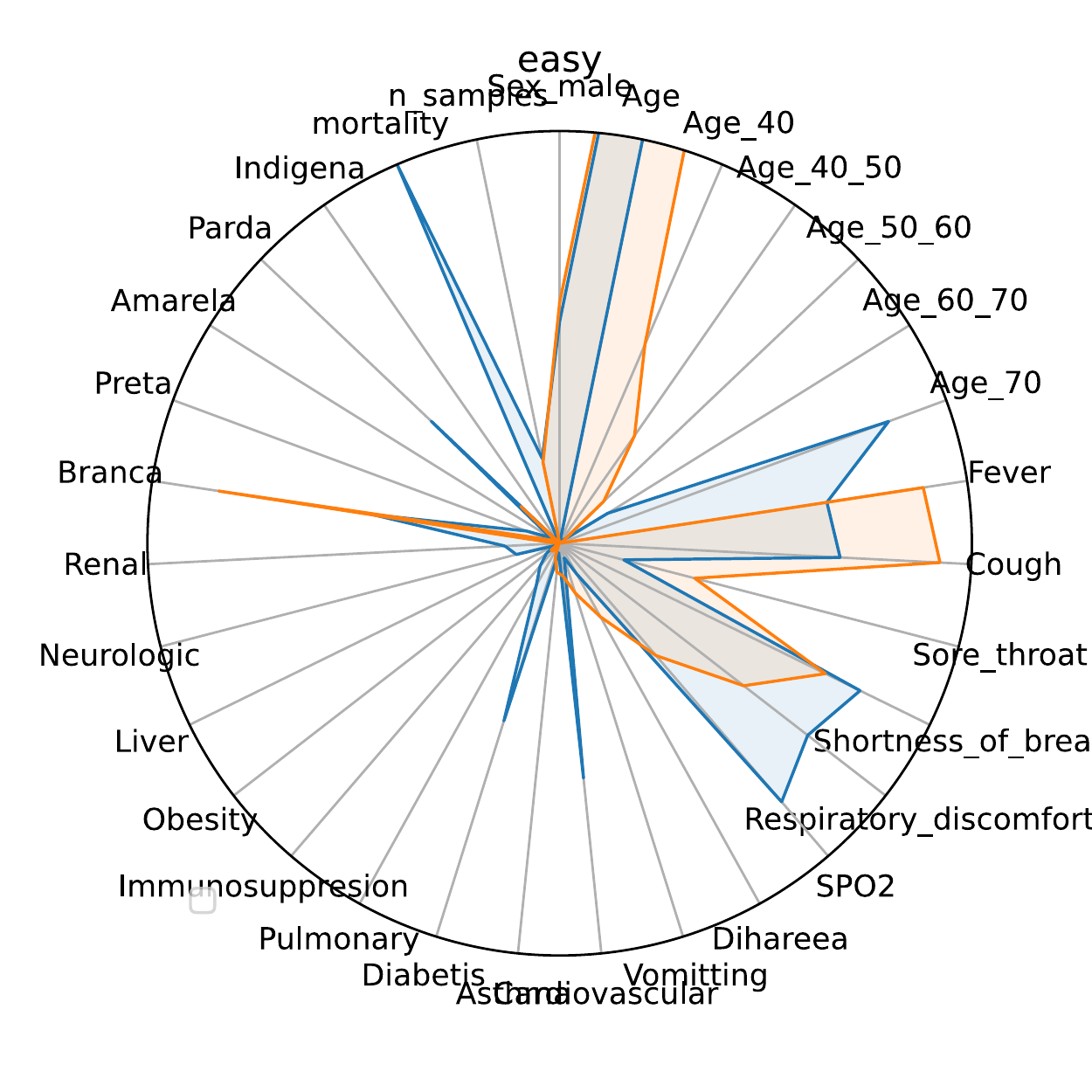}}\quad\quad
  \subfigure[Ambiguous ]{\includegraphics[width=0.255\textwidth]{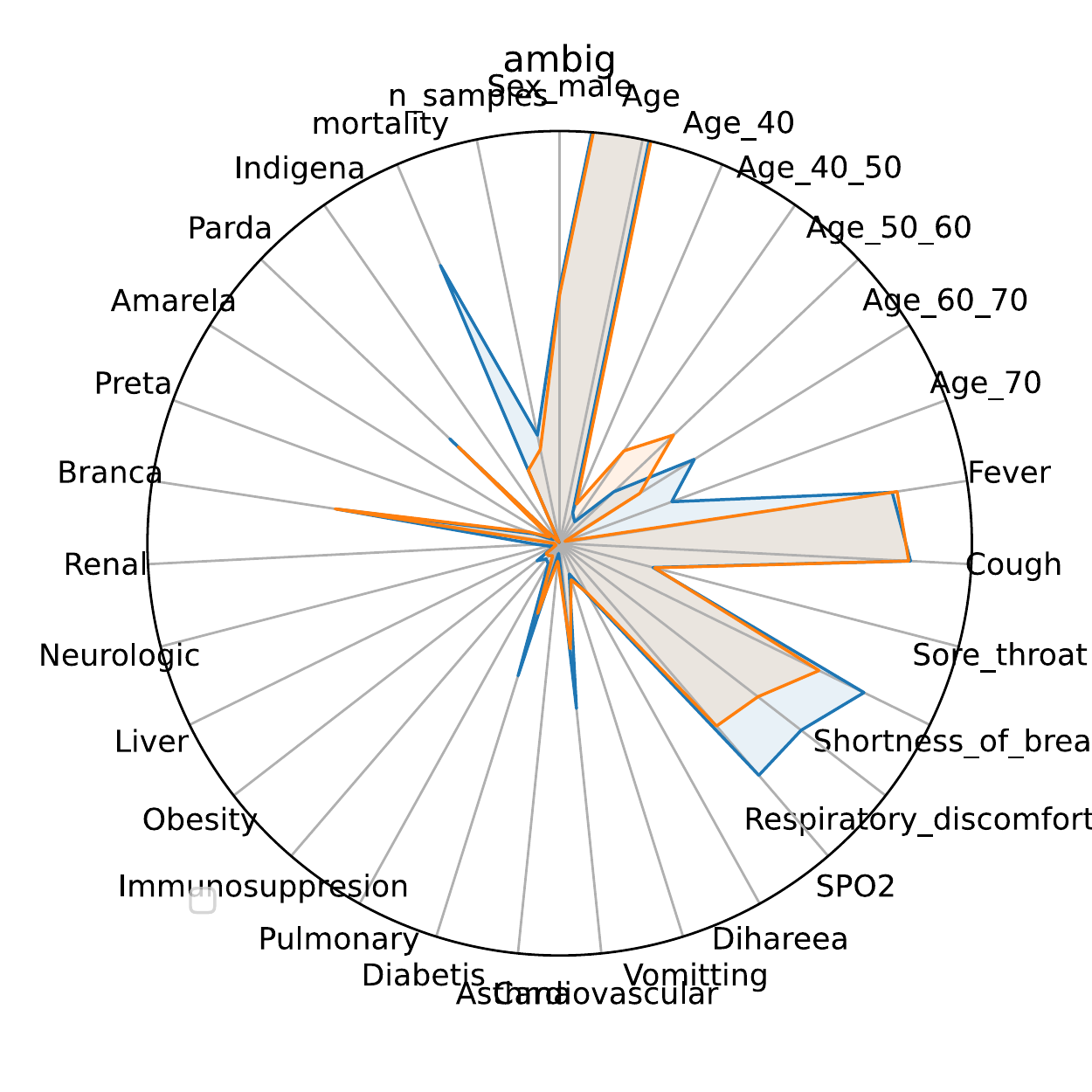}}\quad\quad
  \subfigure[Hard]{{\includegraphics[width=0.255\textwidth]{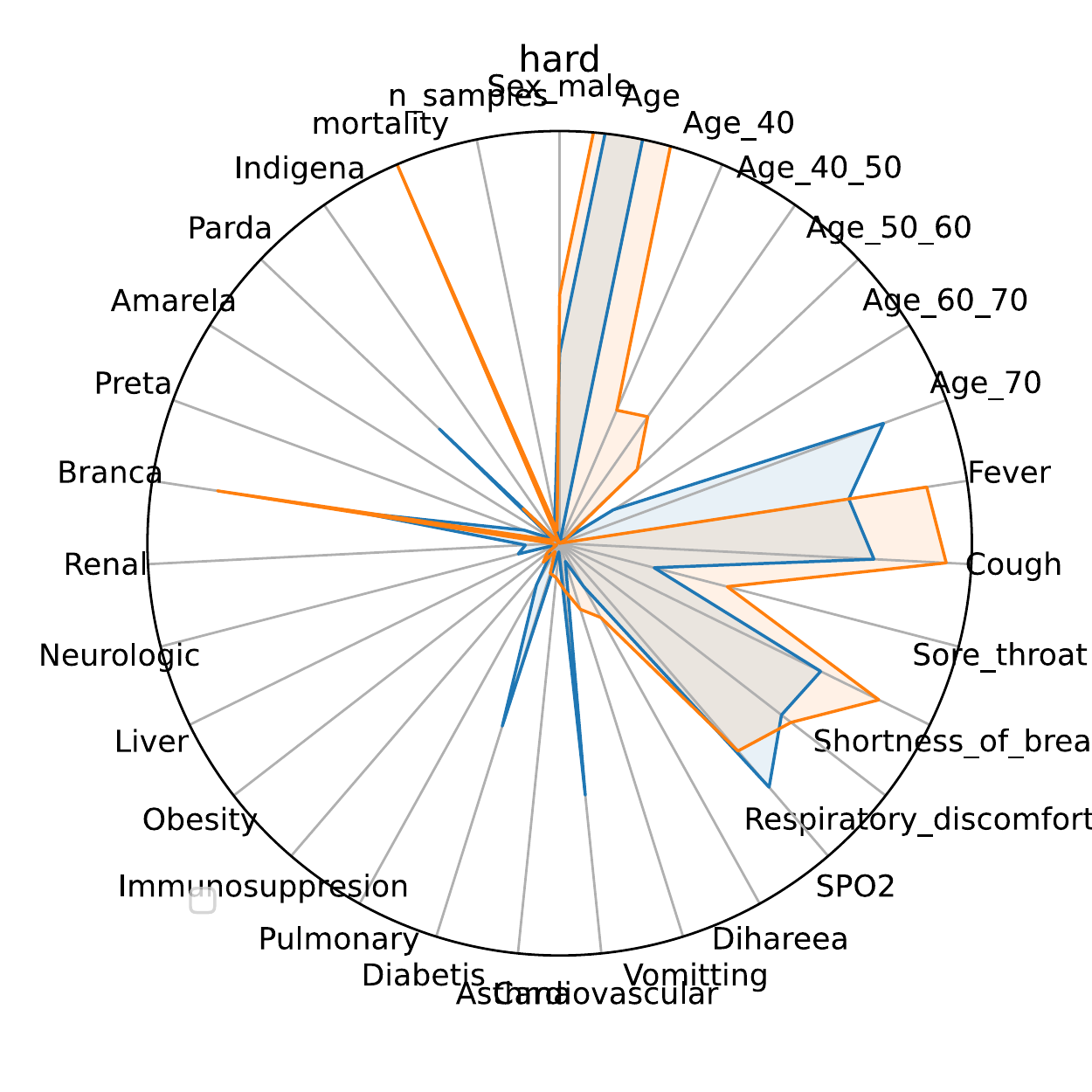}}}

  \caption{\footnotesize{Covid subgroups identified by Data-IQ (descriptions above). Colors represent the GMM clusters}}
  \label{fig:covid_radar}
\end{figure*}

\paragraph{Goal.} The main text (Section \ref{insights}) illustrated the data insights that can be obtained for a single dataset (Prostate). Here, we include the results for all other datasets (using the same setup). See Figures \ref{fig:fetal_radar}-\ref{fig:covid_radar}.

\paragraph{Takeaway.} In general, for each dataset, the subgroups represent: (1) $\easy$: Severe patients with a death outcome, and less severe patients with a survival outcome, (2) $\ambiguous$: Patients with similar features, but different outcomes. This could suggest that the features we have at hand are insufficient to separate the differences in outcomes and (3) $\hard$: Severe patients with a survival outcome, and less severe patients with a death outcome. i.e. opposite outcomes as expected due to randomness in the outcomes.

\newpage

\subsection{Principled feature acquisition (P2): additional results}
\paragraph{Goal.} The main text (Section \ref{p2-exp}) illustrated principled feature acquisition using Data-IQ for a single dataset. Here, we include the results for all other datasets (using the same semi-synthetic setup, based on correlation of the feature with the target).  Recall our formulation, that a valuable feature should decrease the ambiguity of an example, overall decreasing the proportion of $\ambiguous$ examples in the dataset. This permits a principled approach to feature acquisition. 

\paragraph{Takeaway.} We see similar results to the main text for all datasets. For Data-IQ, we see that as we acquire ``valuable'' features, the proportion of the $\ambiguous$ subgroup drops, while the proportion of the $\easy$ subgroup increases. There are significant changes for the important features. This shows that Data-IQ's subgroup characterization can be used to quantify a feature's value, by its ability to decrease ambiguity. In contrast, Data Maps, shows minimal response to feature acquisition, suggesting that it may not be sensitive enough to capture the feature's value.

\begin{figure}[!h]
  \centering
  \subfigure[
\scriptsize{Data-IQ subgroup ambiguity proportion is reduced as more informative features are acquired.}]{\includegraphics[width=0.35\textwidth]{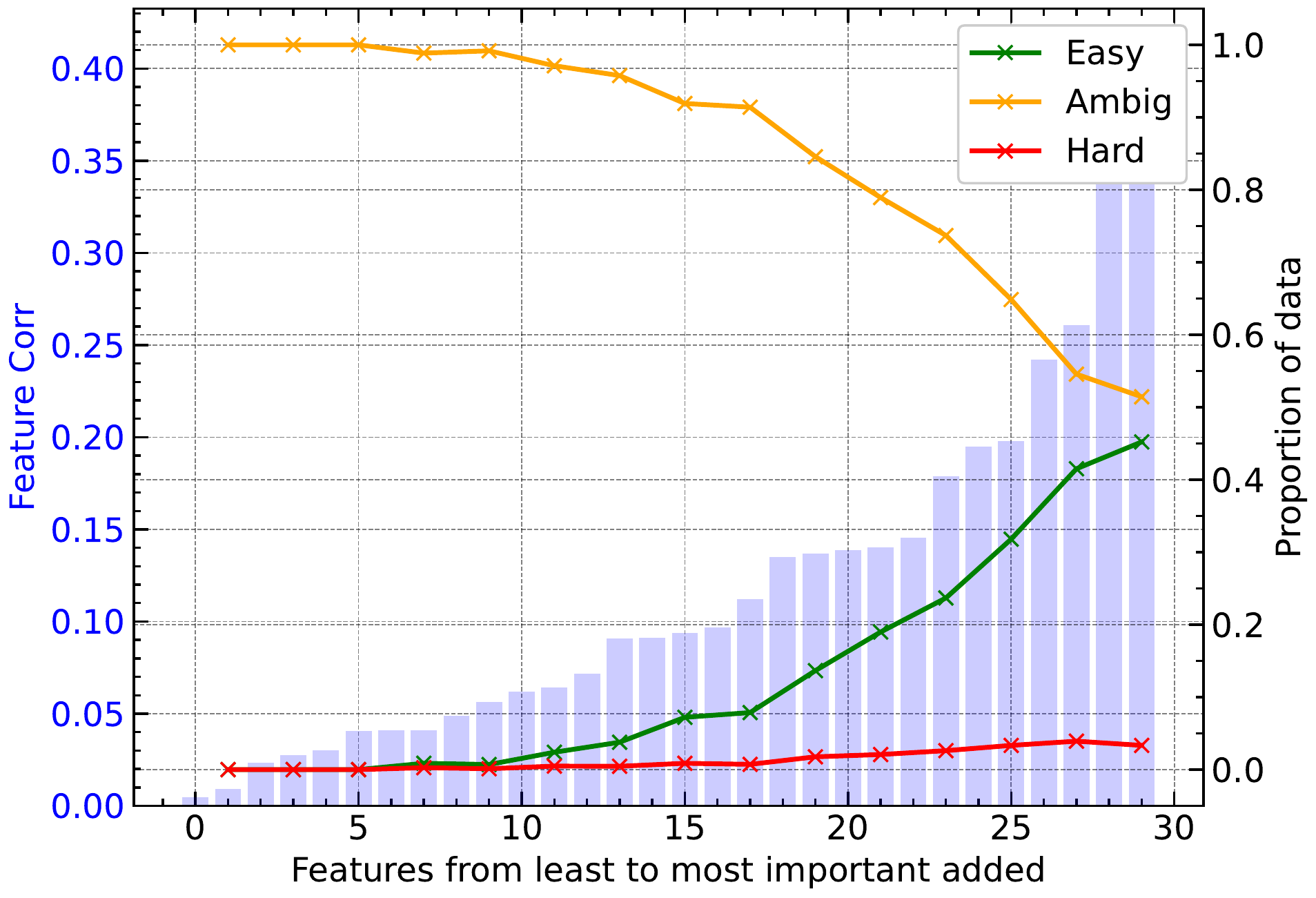}}\quad
  \subfigure[\scriptsize{Data-IQ aleatoric uncertainty remains stable for Ambiguous, reduces for others as features are acquired.} ]{\includegraphics[width=0.35\textwidth]{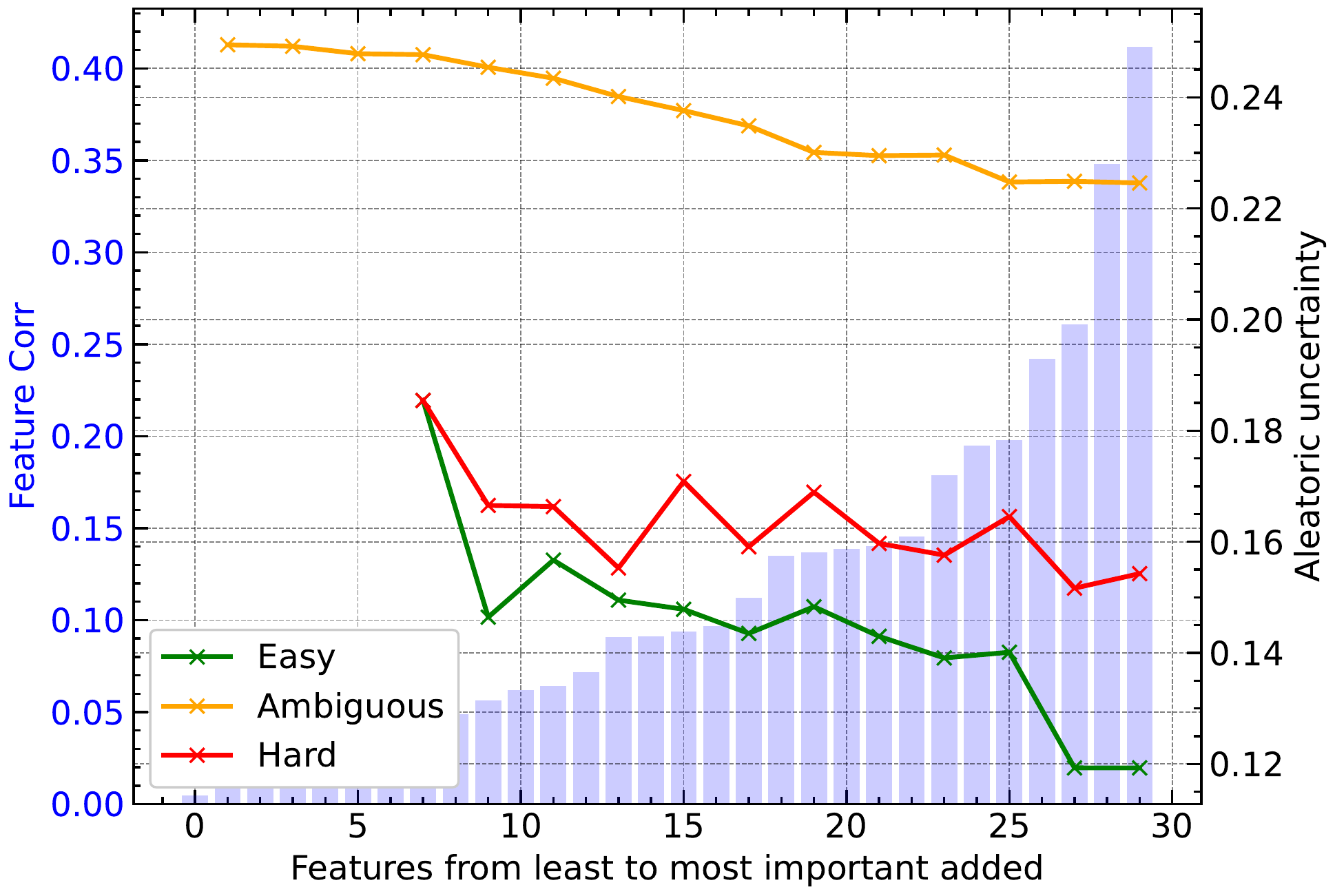}}\quad\\
  \subfigure[\scriptsize{Data Maps subgroup proportions largely unaffected as features acquired.}]{\includegraphics[width=0.35\textwidth]{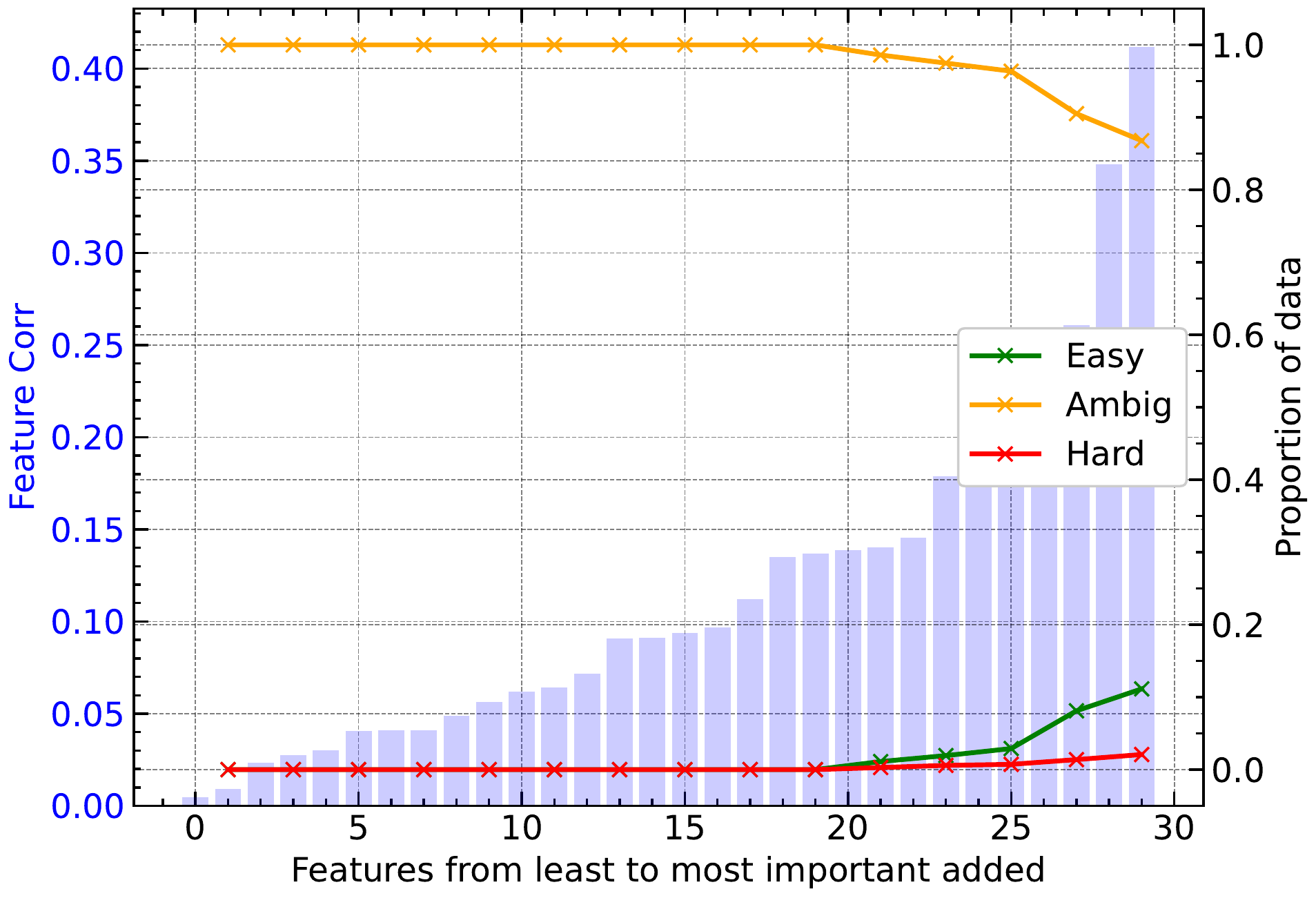}}\quad
  \subfigure[\scriptsize{Data Maps variability increases across subgroups as features acquired.} ]{\includegraphics[width=0.35\textwidth]{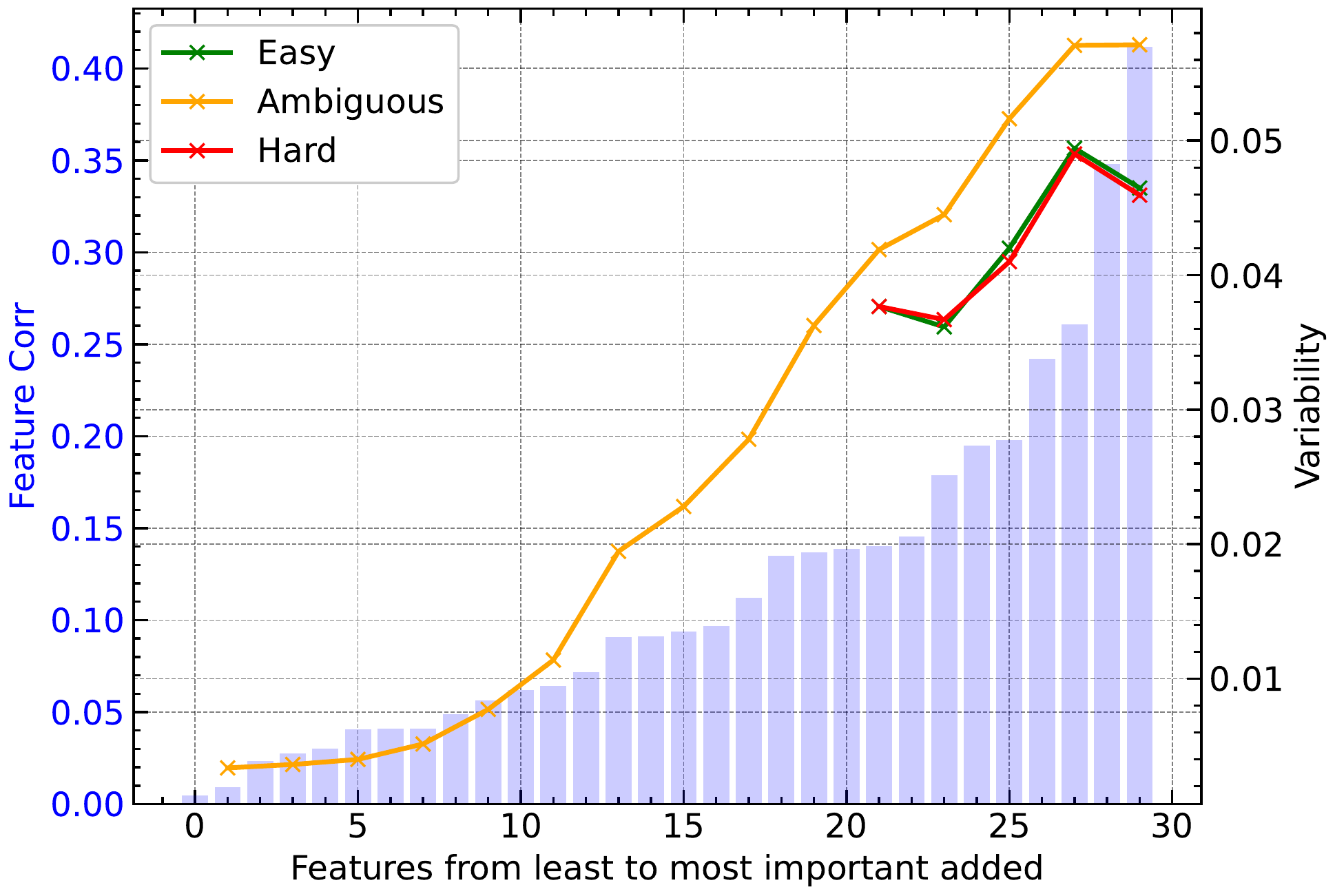}}\quad
  \caption{COVID Quantifying the value of feature acquisition based on change in ambiguity. Only Data-IQ captures this relationship.}
  \label{fig:covid_feats}
\end{figure}

\begin{figure}[!h]
  \centering
  \subfigure[
\scriptsize{Data-IQ subgroup ambiguity proportion is reduced as more informative features are acquired.}]{\includegraphics[width=0.35\textwidth]{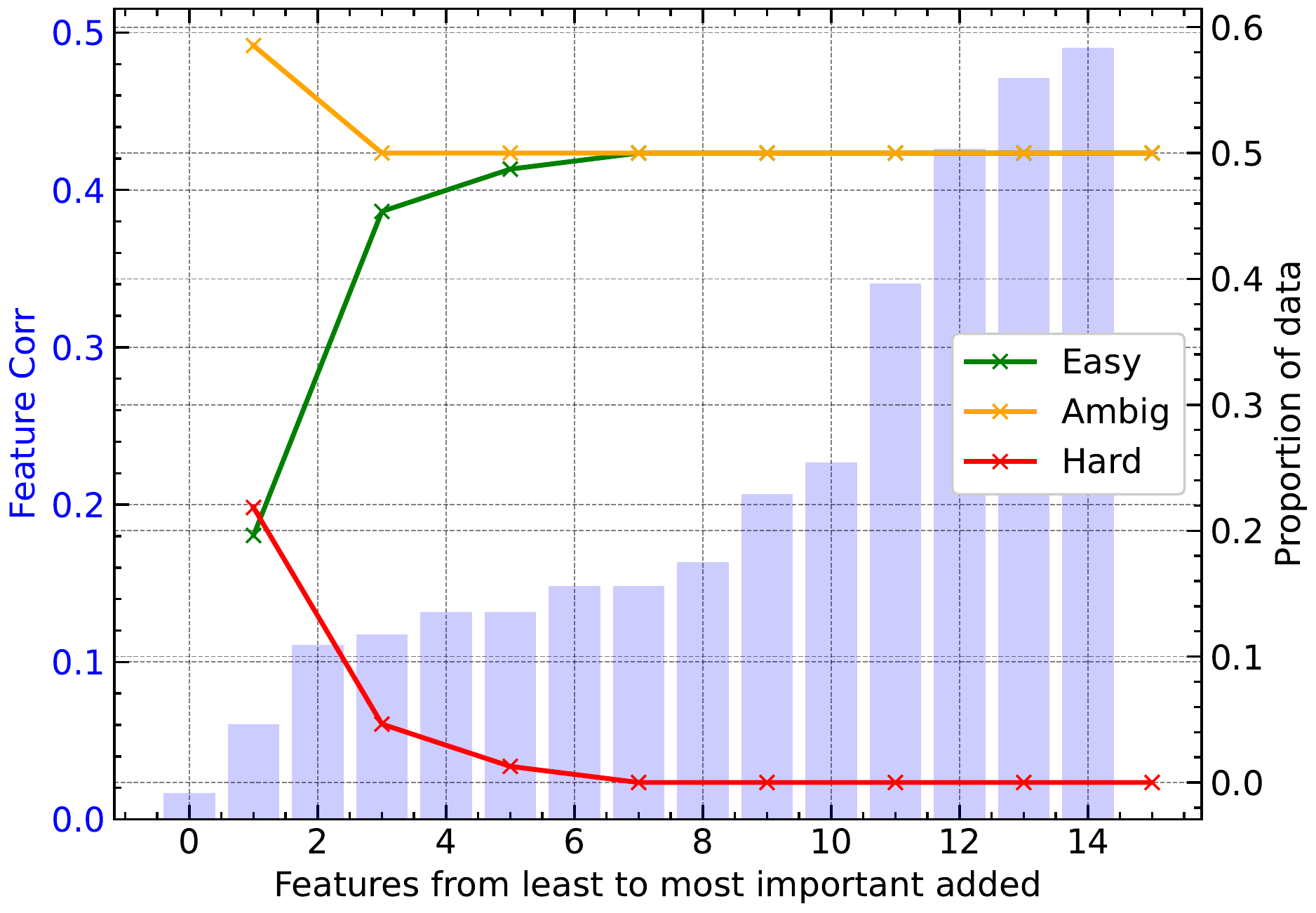}}\quad
  \subfigure[\scriptsize{Data-IQ aleatoric uncertainty remains stable for Ambiguous, reduces for others as features are acquired.} ]{\includegraphics[width=0.35\textwidth]{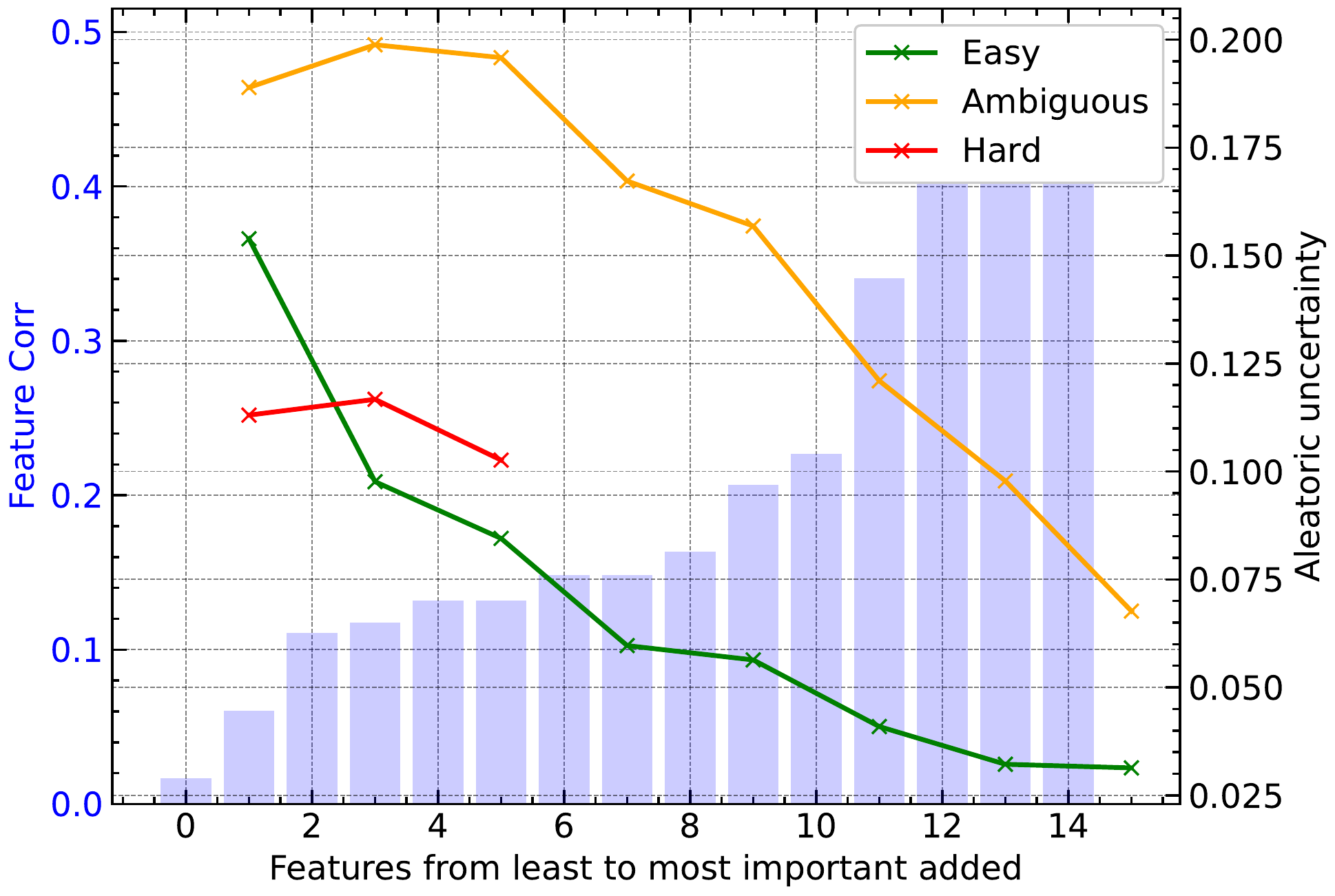}}\quad\\
  \subfigure[\scriptsize{Data Maps subgroup proportions largely unaffected as features acquired.}]{\includegraphics[width=0.35\textwidth]{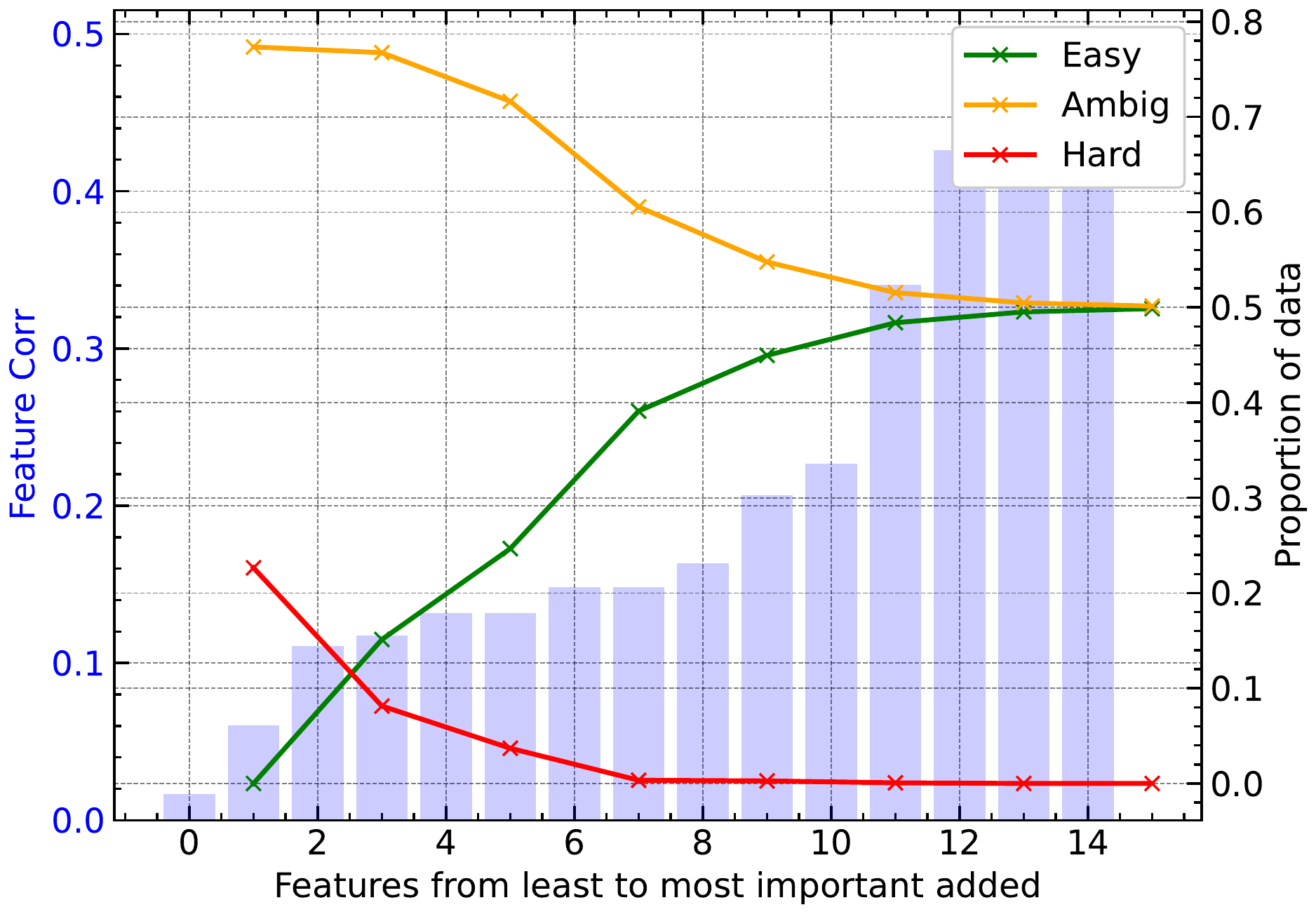}}\quad
  \subfigure[\scriptsize{Data Maps variability increases across subgroups as features acquired.} ]{\includegraphics[width=0.35\textwidth]{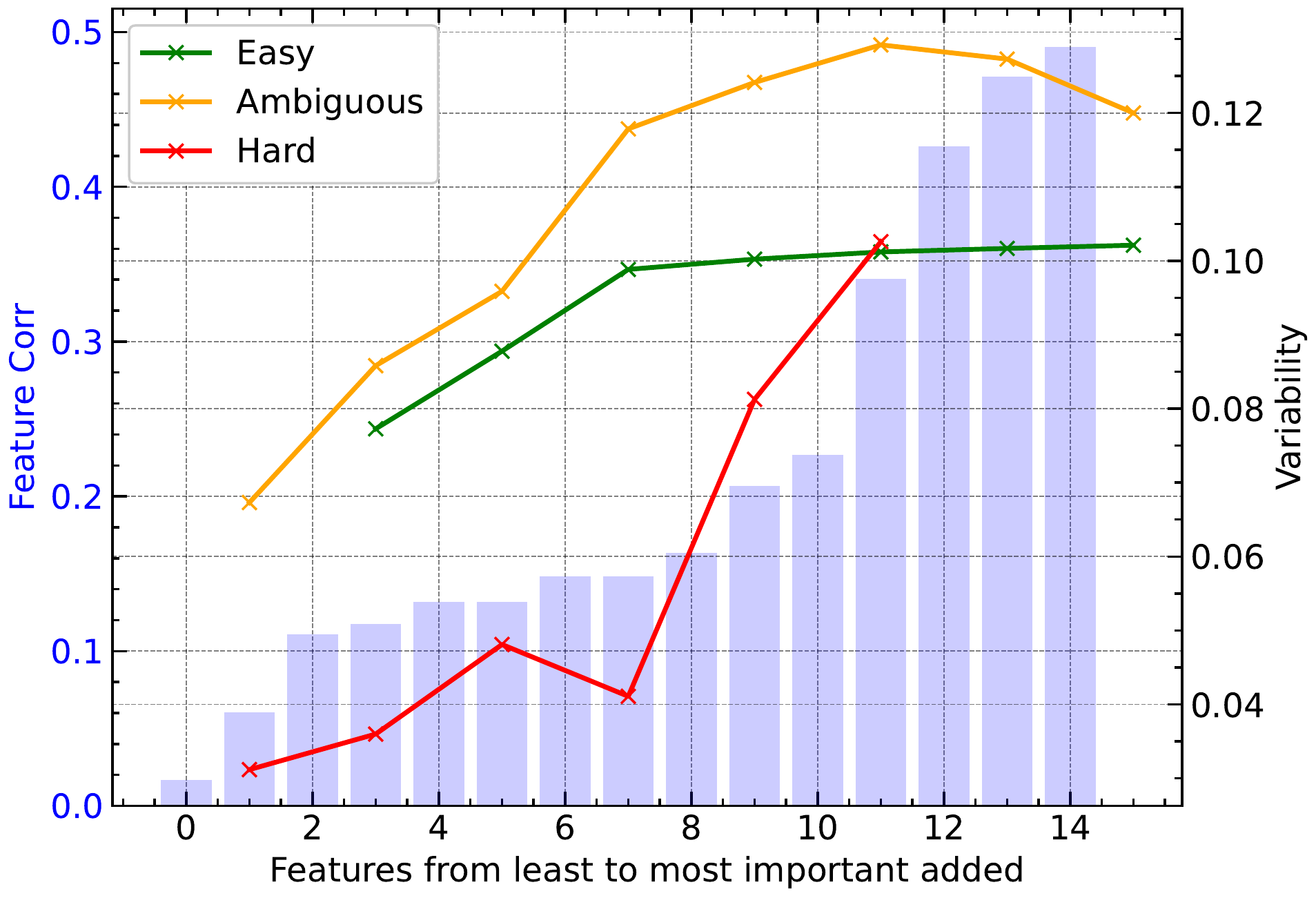}}\quad
  \caption{FETAL Quantifying the value of feature acquisition based on change in ambiguity. Only Data-IQ captures this relationship.}
  \label{fig:fetal_feats}
\end{figure}

\begin{figure}[!h]
  \centering
  \subfigure[
\scriptsize{Data-IQ subgroup ambiguity proportion is reduced as more informative features are acquired.}]{\includegraphics[width=0.35\textwidth]{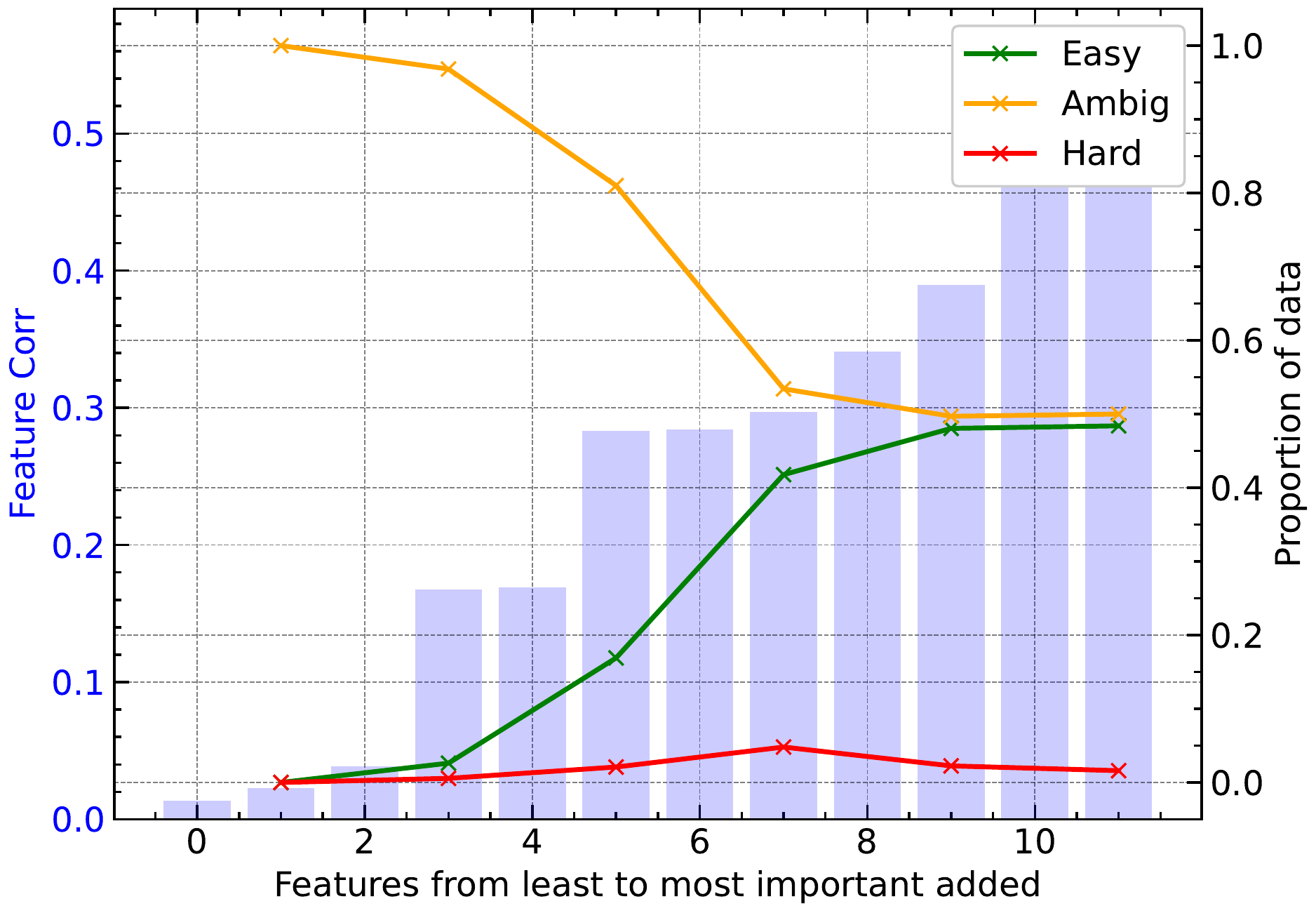}}\quad
  \subfigure[\scriptsize{Data-IQ aleatoric uncertainty remains stable for Ambiguous, reduces for others as features are acquired.} ]{\includegraphics[width=0.35\textwidth]{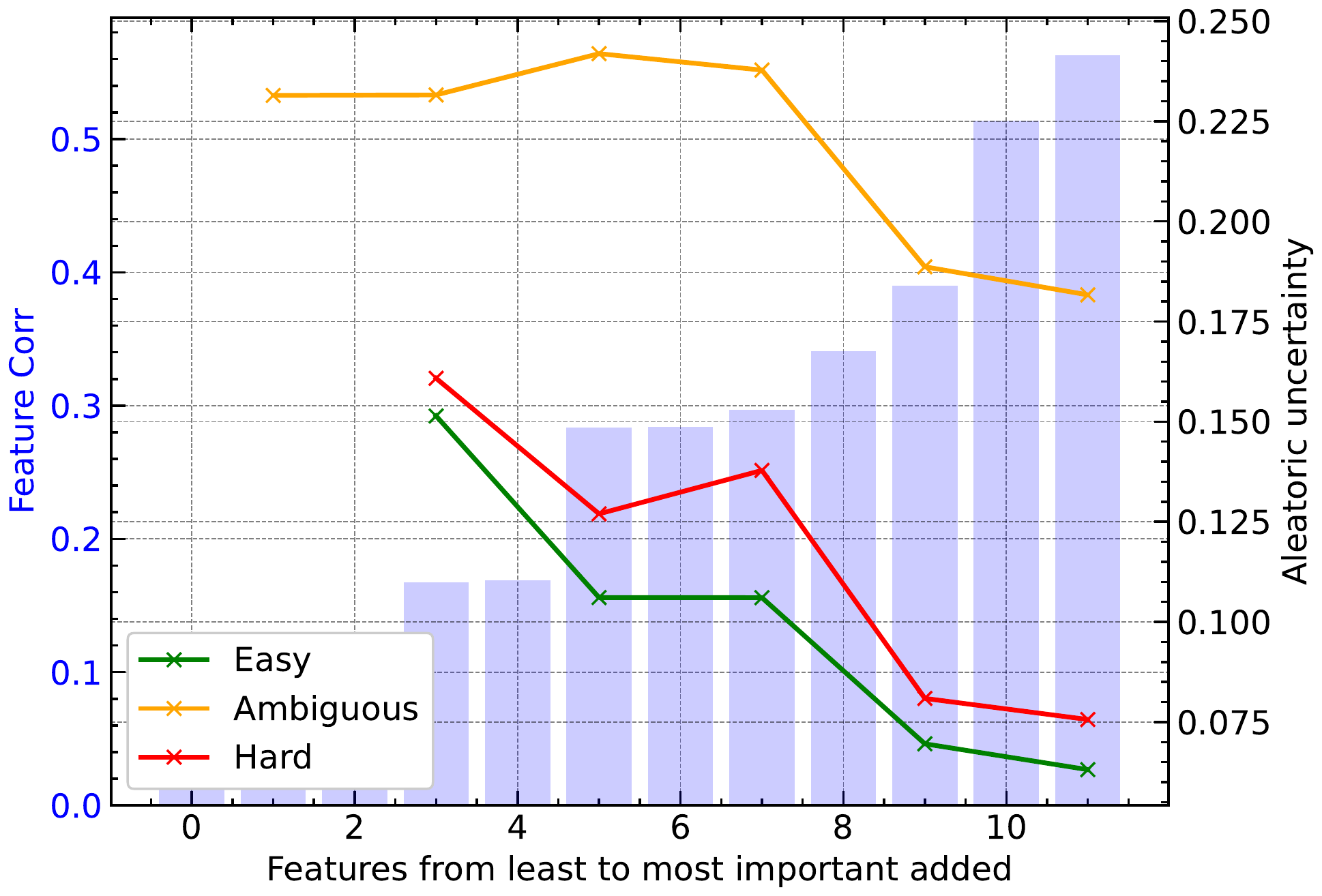}}\quad\\
  \subfigure[\scriptsize{Data Maps subgroup proportions largely unaffected as features acquired.}]{\includegraphics[width=0.35\textwidth]{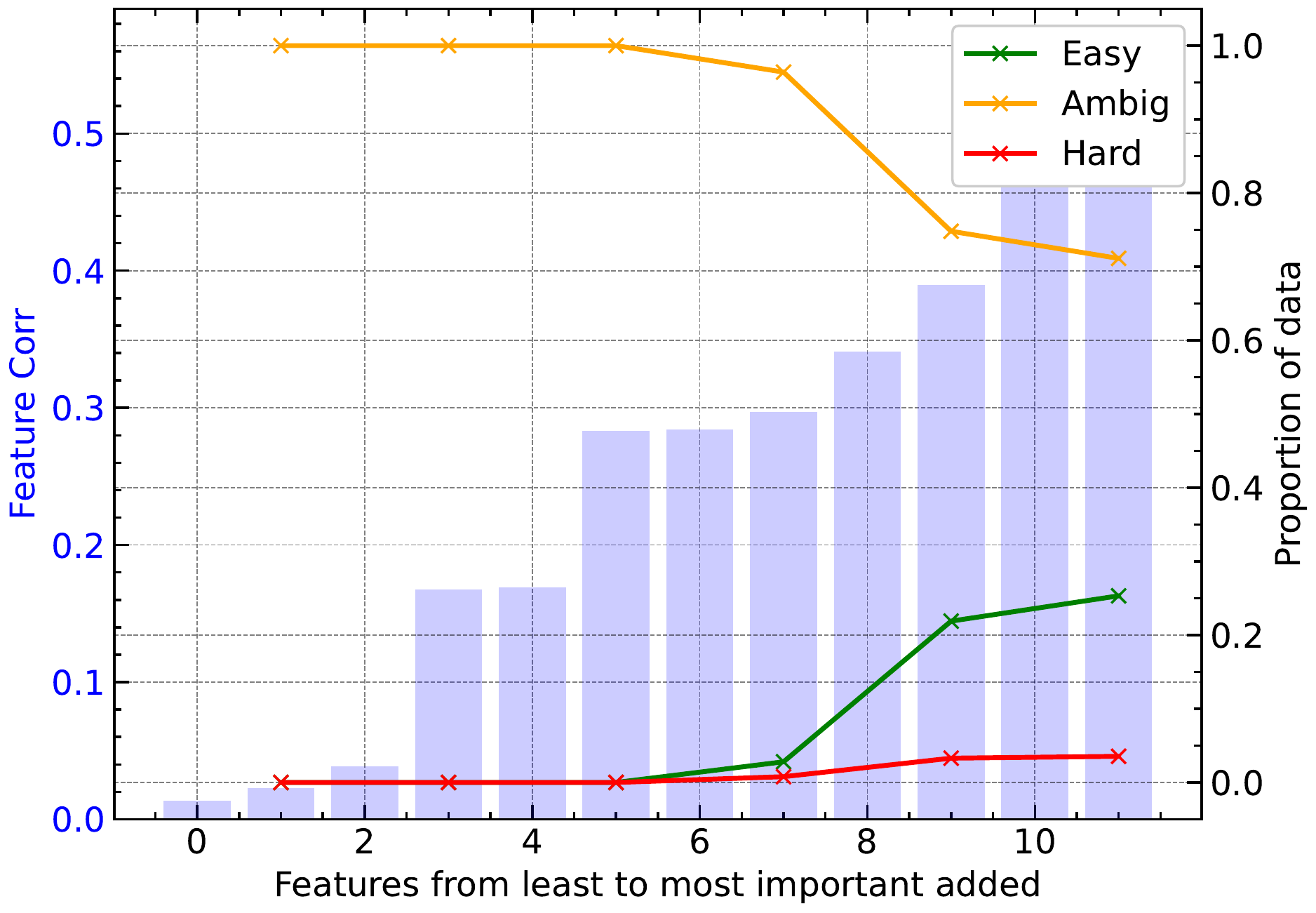}}\quad
  \subfigure[\scriptsize{Data Maps variability increases across subgroups as features acquired.} ]{\includegraphics[width=0.35\textwidth]{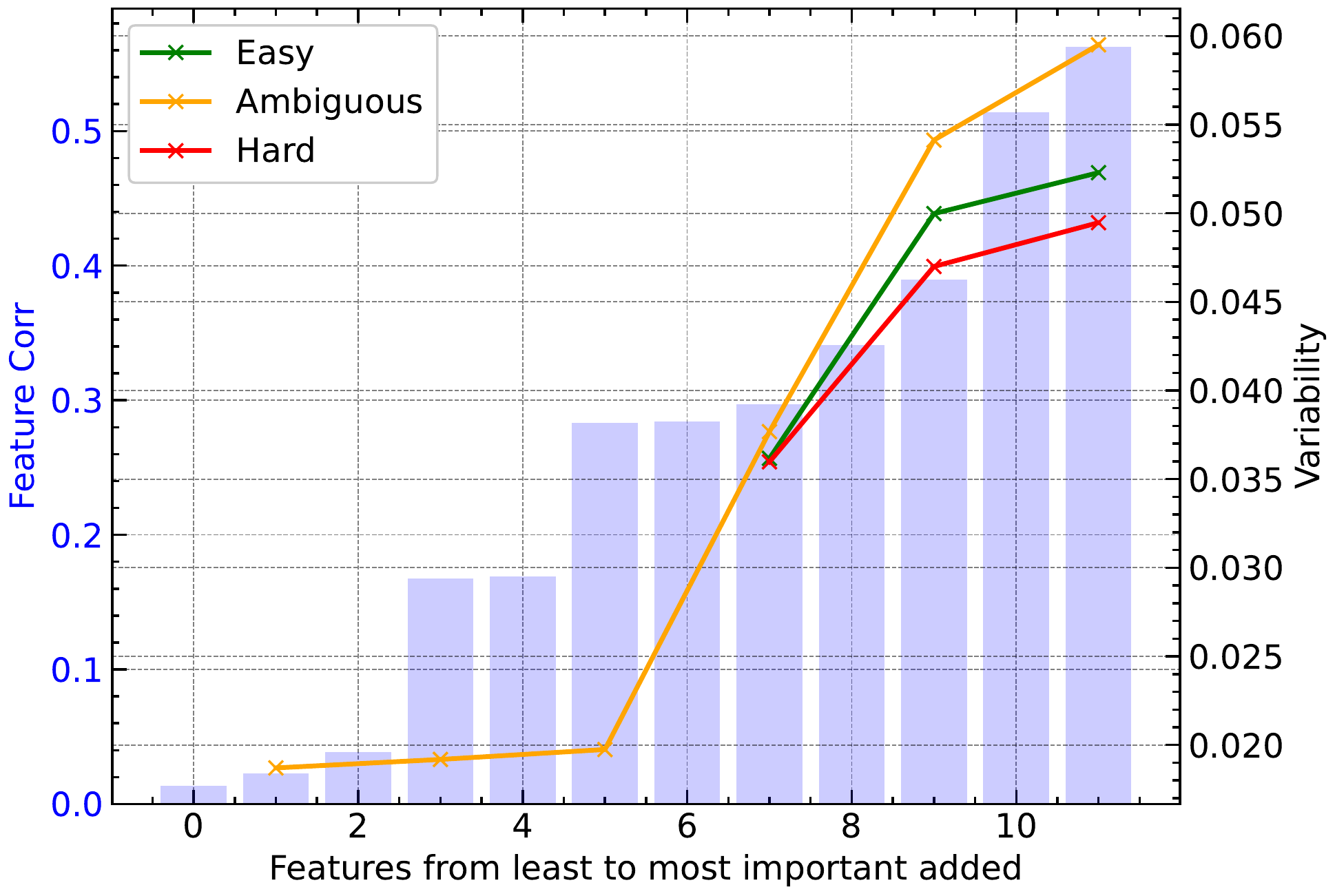}}\quad
  \caption{Prostate Quantifying the value of feature acquisition based on change in ambiguity. Only Data-IQ captures this relationship.}
  \label{fig:prostate_feats}
\end{figure}

\begin{figure}[!h]
  \centering
  \subfigure[
\scriptsize{Data-IQ subgroup ambiguity proportion is reduced as more informative features are acquired.}]{\includegraphics[width=0.35\textwidth]{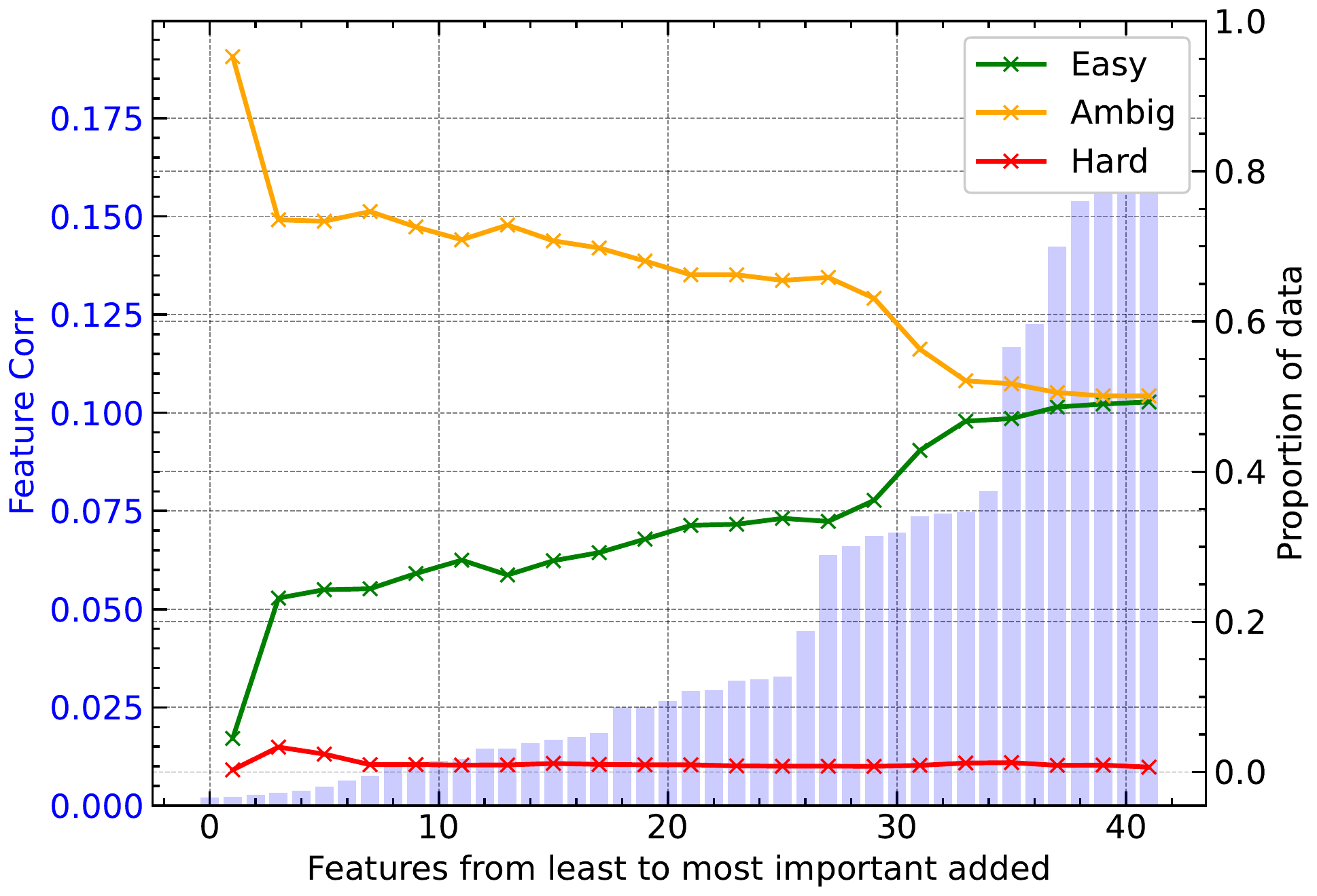}}\quad
  \subfigure[\scriptsize{Data-IQ aleatoric uncertainty remains stable for Ambiguous, reduces for others as features are acquired.} ]{\includegraphics[width=0.35\textwidth]{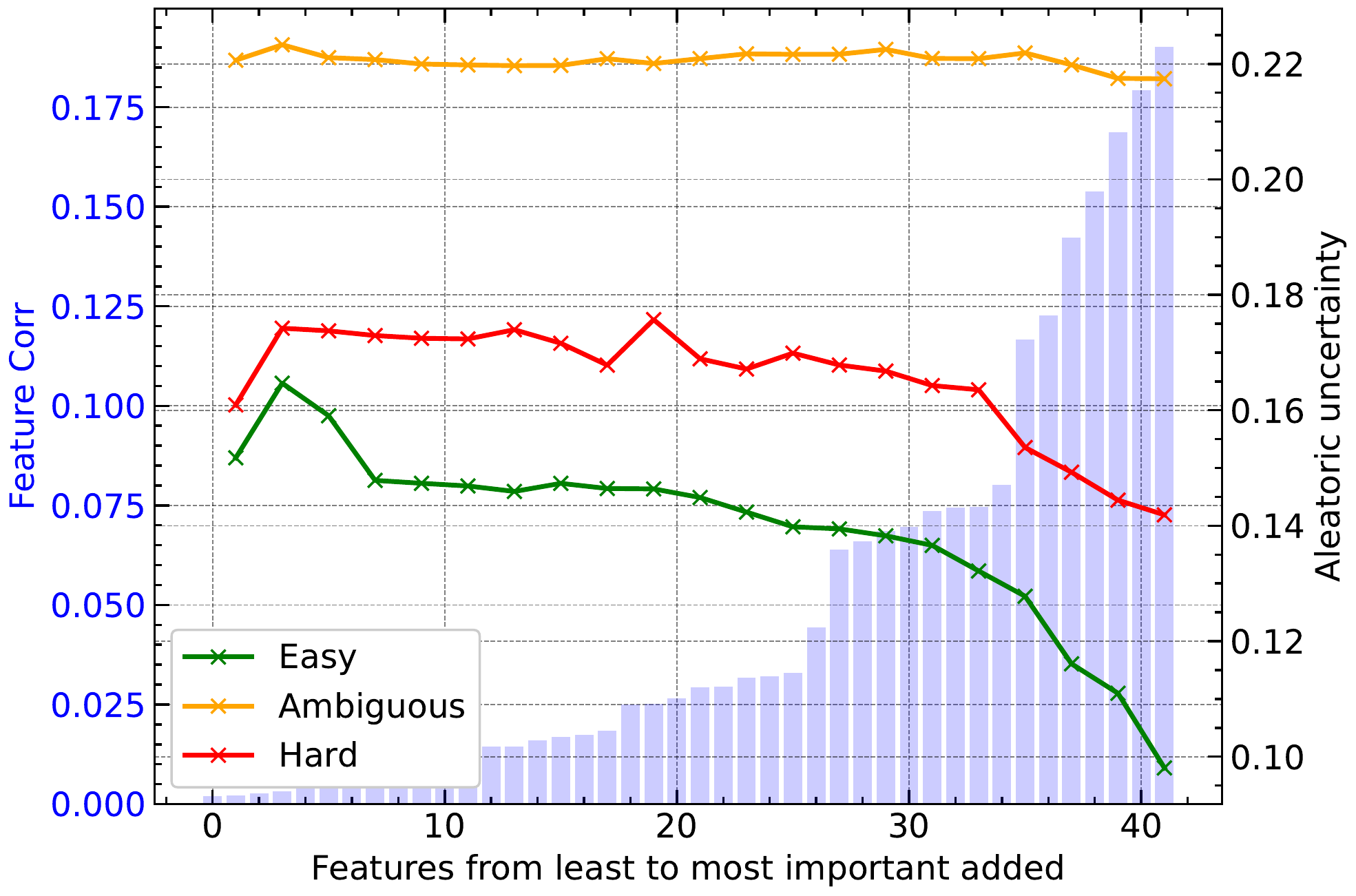}}\quad\\
  \subfigure[\scriptsize{Data Maps subgroup proportions largely unaffected as features acquired.}]{\includegraphics[width=0.35\textwidth]{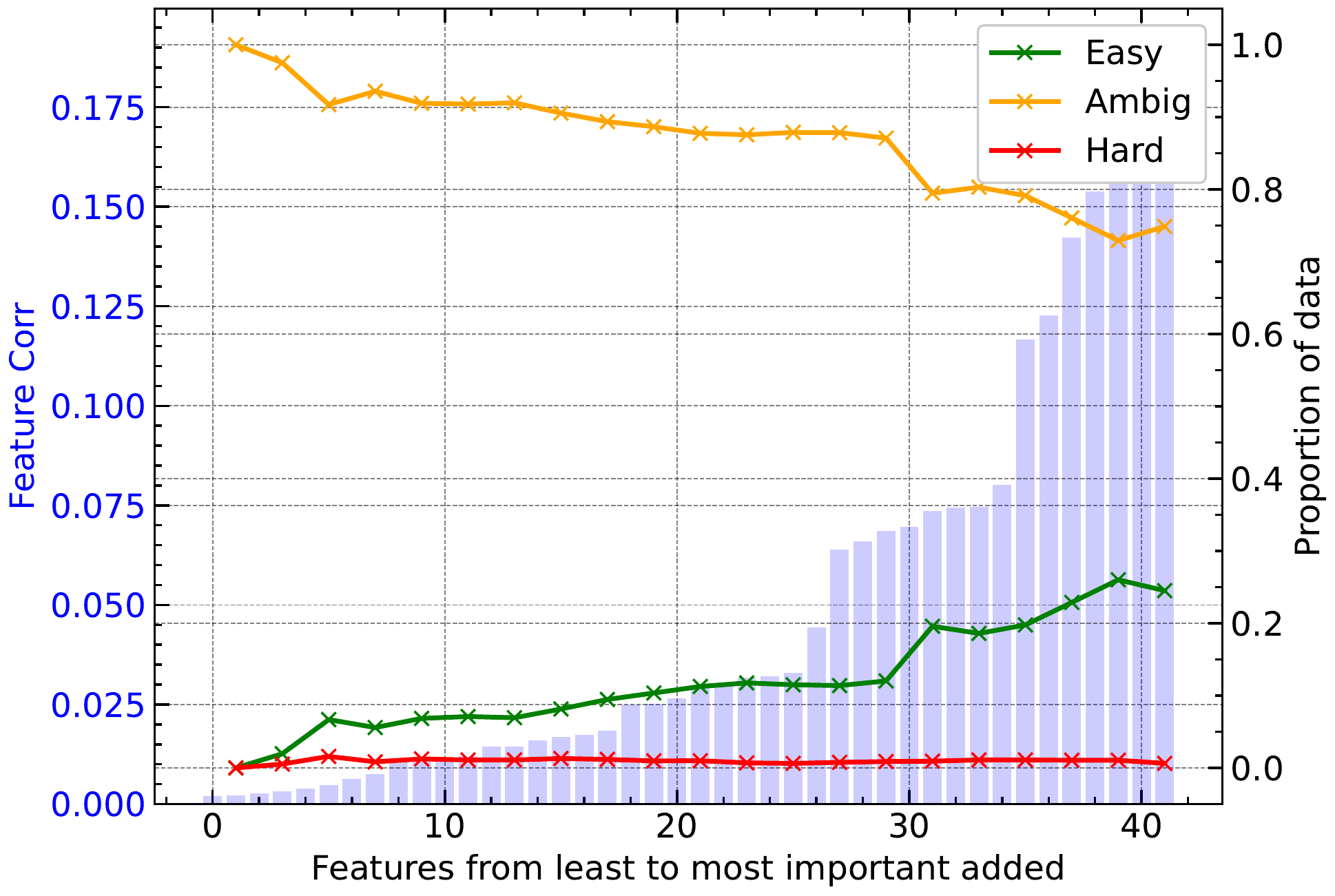}}\quad
  \subfigure[\scriptsize{Data Maps variability increases across subgroups as features acquired.} ]{\includegraphics[width=0.35\textwidth]{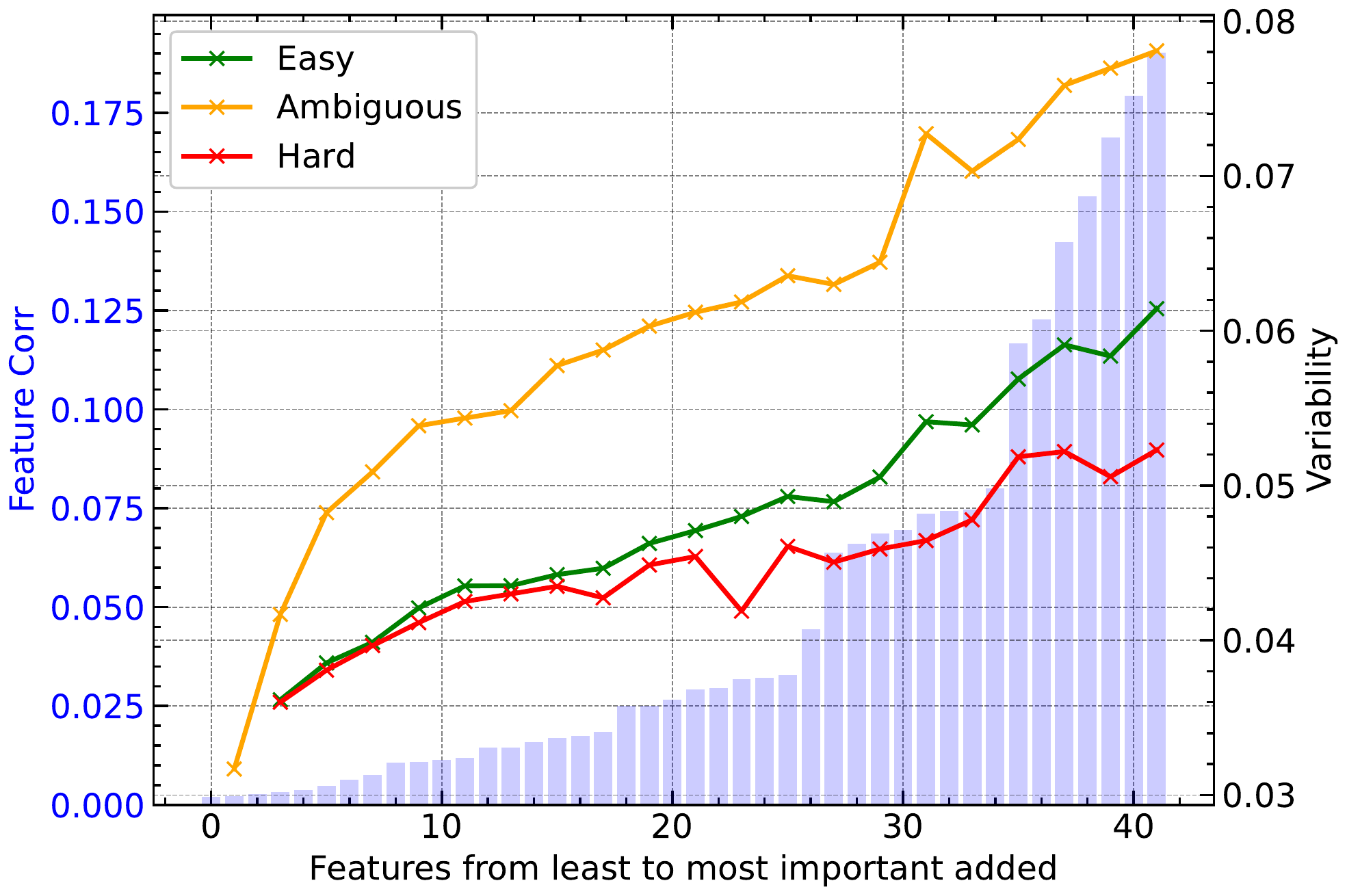}}\quad
  \caption{Support Quantifying the value of feature acquisition based on change in ambiguity. Only Data-IQ captures this relationship.}
  \label{fig:support_feats2}
\end{figure}

\begin{figure*}[!h]

  \centering 
  \subfigure[Data-IQ: TEXT ]{\includegraphics[width=0.40\textwidth]{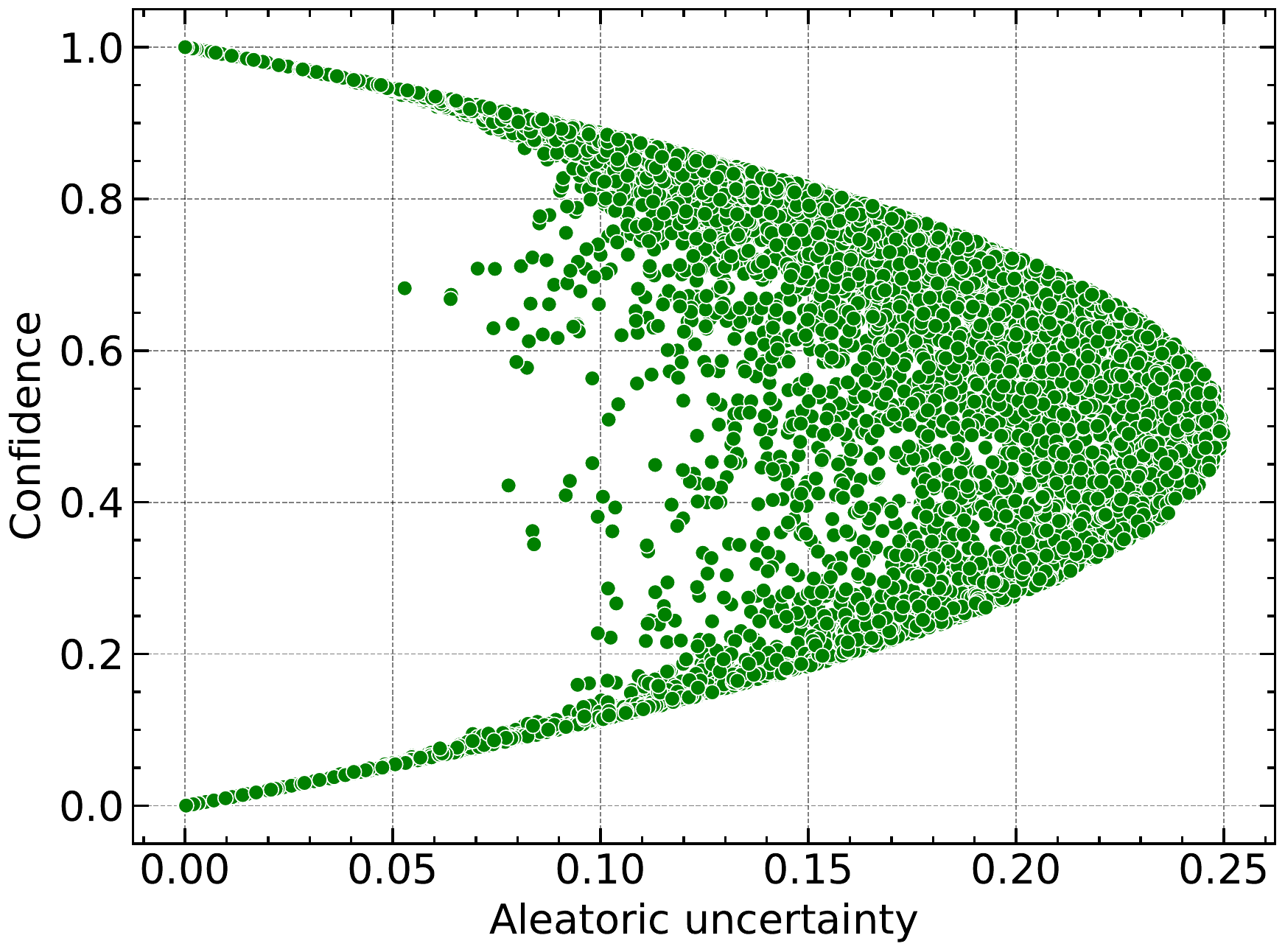}}\quad\quad
  \subfigure[Data Maps: TEXT]{\includegraphics[width=0.40\textwidth]{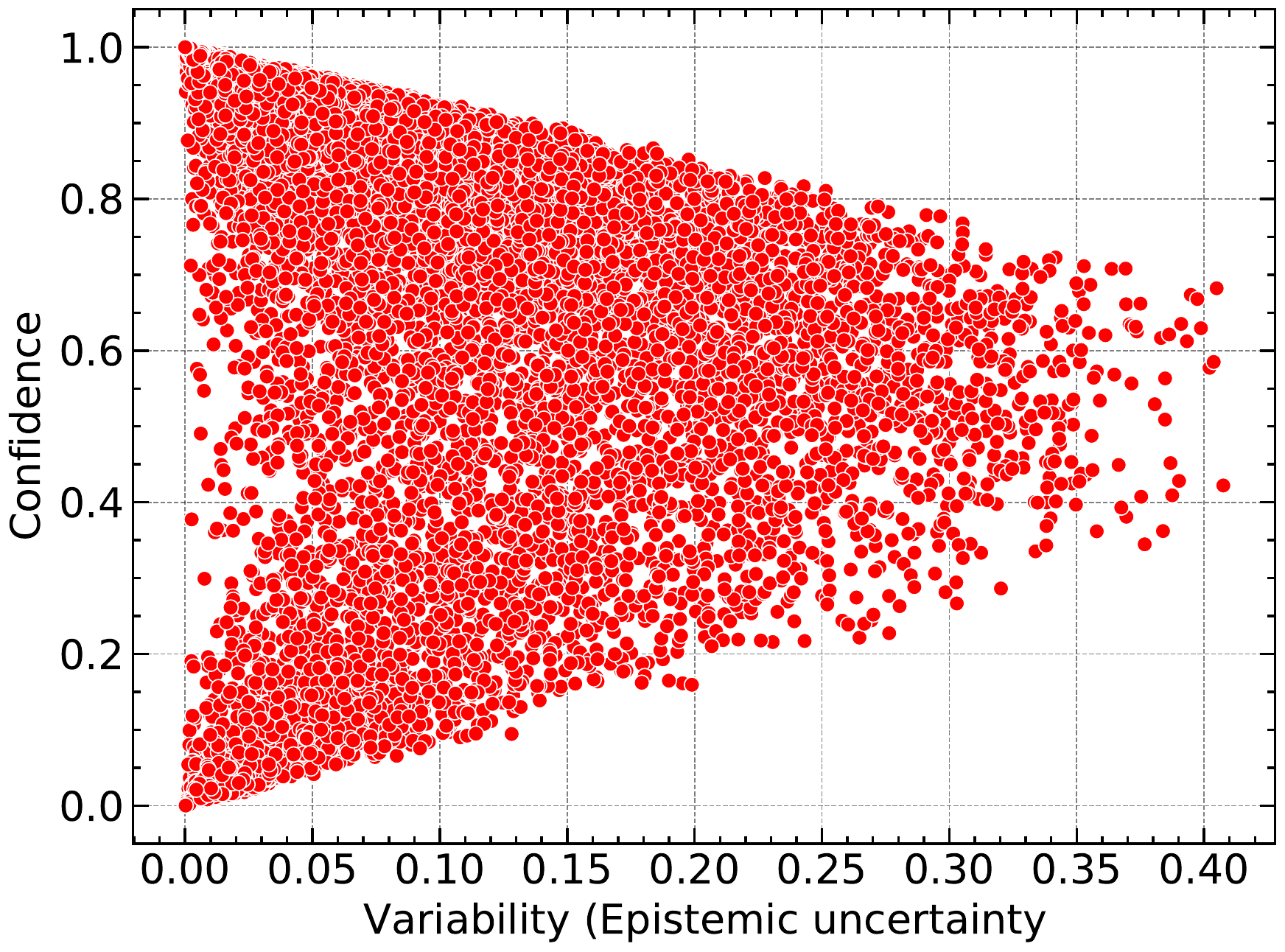}}
  \subfigure[Data-IQ: IMAGE ]{\includegraphics[width=0.40\textwidth]{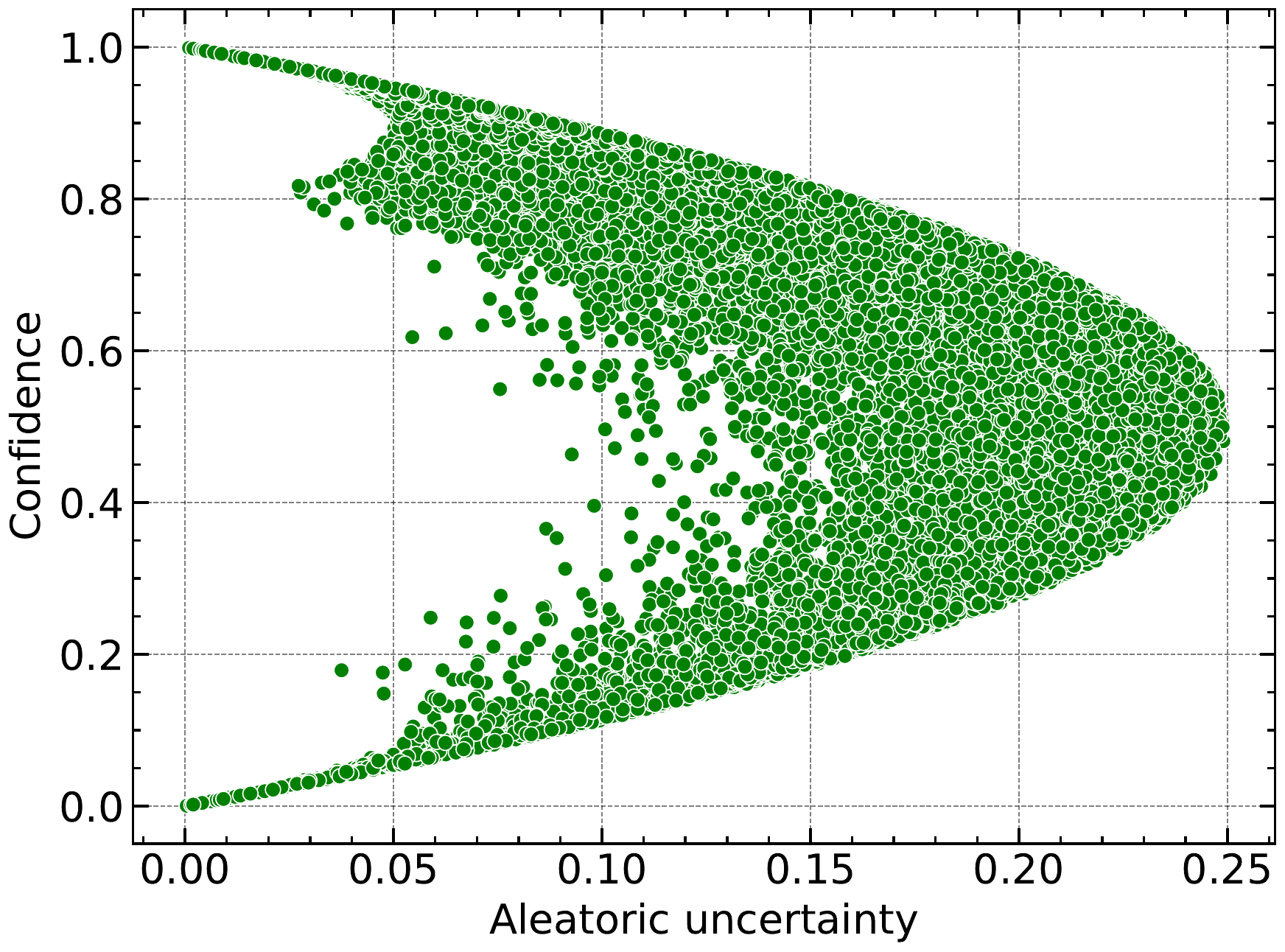}}\quad\quad
  \subfigure[Data Maps: IMAGE]{\includegraphics[width=0.40\textwidth]{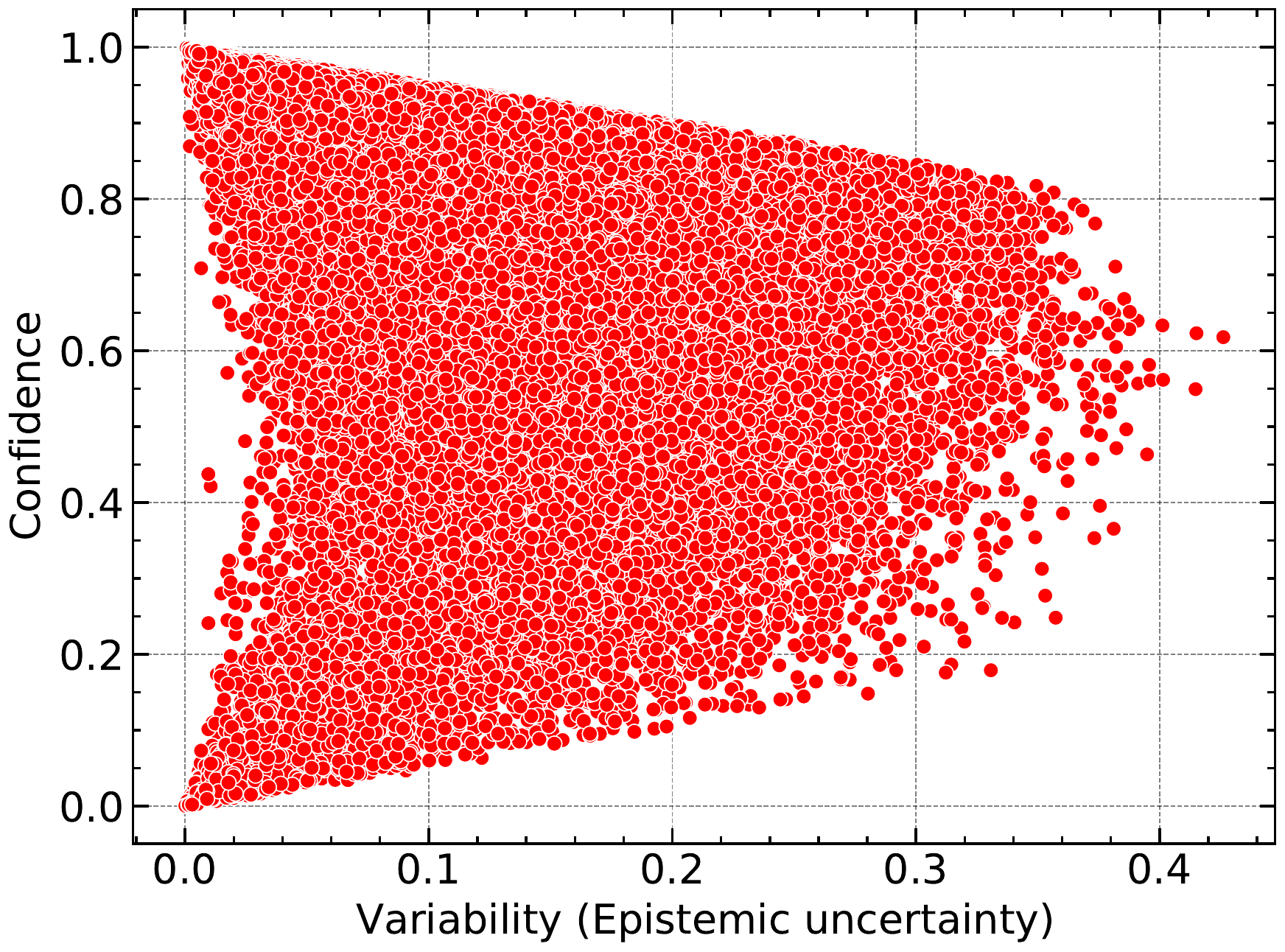}}
  \caption{Data-IQ vs Data Maps compared on text and image modalities}
  \label{fig:image_text_figs}
\end{figure*}

\clearpage

\subsection{Data-IQ on text data (NLP)}\label{nlp}

\paragraph{Goal.} We wish to assess the generality of Data-IQ. Whilst, in the main paper we mainly tackled tabular data, we wish to showcase the potential utility of Data-IQ on text data for NLP tasks.

\paragraph{Experiment.} We train a bidirectional RNN on the IMBDb 50K dataset, where the task is to predict the sentiment on the basis of the reviews. We then use Data-IQ to analyze the training dynamics and categorize the data into subgroups - $\easy$, $\ambiguous$~ and $\hard$

\paragraph{Takeaway.} We notice that the $\easy$, $\ambiguous$~ and $\hard$ groups represent different phenomena in the text data. We summarize these below.

\textbf{EASY: Average review length - 163 words}

- Easy to predict review. Very negative sentiment with a negative label

\noindent\fbox{%
    \parbox{\textwidth}{%
    \textcolor{red}{\textbf{NEGATIVE (0)}}:
This movie is \textcolor{red}{pathetic} in every way possible. \textcolor{red}{Bad acting, horrible script (was there one?), terrible editing, lousy cinematography, cheap humor}. Just \textcolor{red}{plain horrible}. I had seen 'The Wishmaster' a couple weeks before this movie and I thought it was a dead-ringer for \textcolor{red}{worst movie of the yea}r. Then, I saw 'The Pest' and suddenly 'The Wishmaster' didn't seem so bad at all. \textcolor{red}{Bad Bad Bad. Excruciatingly bad}.
    }%
}

\noindent\fbox{%
    \parbox{\textwidth}{%
\textcolor{ForestGreen}{\textbf{POSITIVE (1)}}
When I saw this movie for the first time I was both surprised and a little shocked by the blatant \textcolor{ForestGreen}{vibrance} of the story. It is a very artistic drama with \textcolor{ForestGreen}{incredible special effects}, \textcolor{ForestGreen}{spectacular acting}, not to mention a very \textcolor{ForestGreen}{excellent job} in the makeup department. Jennifer Lopez has pulled herself out of past roles that dug into her career with this movie, portraying a very sensitive child psychologist who works with a team of engineers to enter the minds of comatose patients to treat them. Vincent D'onofrio \textcolor{ForestGreen}{played amazingly well}. His portrayal of a sadist serial killer was \textcolor{ForestGreen}{perfect to a T}. The sheer emotion conveyed by his \textcolor{ForestGreen}{performance is astounding}. Vince Vaughn isn't my favorite, but still \textcolor{ForestGreen}{performed exceptionally well}. The symbolism and artistry was \textcolor{ForestGreen}{intriguing} and \textcolor{ForestGreen}{titillating}, sometimes surprising, and other times shocking. Overall, I say this is a \textcolor{ForestGreen}{wonderful movie}, with \textcolor{ForestGreen}{excellent acting} and \textcolor{ForestGreen}{beautiful artwork}

    }%
}

\noindent\fbox{%
    \parbox{\textwidth}{%
 \textcolor{red}{\textbf{NEGATIVE (0)}}:
I've tried to watch this so-called comedy, but it's very  
\textcolor{red}{hard to bear}. This is a  \textcolor{red}{bad, narrow-minded, cliché-ridden movie}. Definitively  \textcolor{red}{not 
funny}, but very much  \textcolor{red}{boring} and  \textcolor{red}{annoying}, indeed.  \textcolor{red}{Bad script, bad acting}. It's a complete  \textcolor{red}{waste of time} - and there remains nothing more to say, I'm afraid 1 out of 10 points

    }%
}

\textbf{AMBIGUOUS: Average review length - 283 words}
- Linguistic ambiguity. A mix of positive and negative comments. Other reviews of AMBIGUOUS even have text which simply describes the story without any positive/negative sentiment.
- Ambiguous reviews also are much longer (in words).

\noindent\fbox{%
    \parbox{\textwidth}{%
 \textcolor{red}{\textbf{NEGATIVE (0)}}
 I saw this movie on the film festival of Rotterdam (jan '06) and followed the discussion between director and public afterwards. Many people reacted shocked and protesting. He will get a lot of  \textcolor{red}{negative critics}. But: the world is cruel like this, and it's not funny. People don't like it. That itself doesn't mean that the movie is bad. I can see that difference. Don't shoot the messenger that shows us the world outside our 'hubble'! Nevertheless I think this a  \textcolor{red}{bad movie}. \textcolor{ForestGreen}{Film-technically it's a good one}. \textcolor{ForestGreen}{Nice shots and script}, most \textcolor{ForestGreen}{good fitting music, great actors}. The director pretends to make a psychological movie, - the psychology however is of  \textcolor{red}{poor quality}. Describing such a powerful violence itself is not the art. The art would be a powerful description of the psychological process behind that violence. How does a shy boy come to such a cruelty? The director pretends to describe that, - but is not good in that.  \textcolor{orange}{The director used several times the word the 'selfishness' of people, mentioning for instance the teacher. Only: this teacher wasn't selfish,- just someone in several roles, caring for his pupils, ánd worried about his script. I think it's a simplification to call him selfish. The atmosphere in the village is creepy, and the mother made awful mistakes ('you terribly let me down') but it doesn't become believable for me, that there is caused súch a lot of pain, that the shyest boy comes to such terrible things.} In fact, reality is far more complex than the way, this film describes and it  \textcolor{red}{needs far better descriptions}. The interesting thing would be: how does it work? Describe that process for me please, so that we understand.With the written phrase on the end, the director said to point to an alternative way of life. It was the other extreme, and confirmed for me that director and scriptwriter are bad psychologists, promoting black/white-thinking. The connection between violence in films and in society has been proved. Use such a violence gives the responsibility to use it right. There are enough black/white-thinkers in the world, causing lots of war and misery.  \textcolor{red}{I hope, this movie won't be successful}.
    }%
    }

\noindent\fbox{%
    \parbox{\textwidth}{%
 \textcolor{red}{\textbf{NEGATIVE (0)}}
 
 When the film started I got the feeling this was going to be \textcolor{ForestGreen}{something special}. The \textcolor{ForestGreen}{acting and camera work were undoubtedly good}. I also \textcolor{ForestGreen}{liked the characters} and could have grown to empathise with them. The film had a \textcolor{ForestGreen}{good atmosphere} and there was a hint of fantasy. However, as the film went on, the  \textcolor{red}{plot never appeared to takeoff} and just rolled on scene by scene. I was  \textcolor{red}{unable to understand the connection} between the stories. All I could see was the characters occasionally bumping into each other and references to ships in bottles. Without that connection, I was just left with a few  \textcolor{red}{unremarkable }short stories. 
    }%
}

\noindent\fbox{%
    \parbox{\textwidth}{%
    \textcolor{ForestGreen}{\textbf{POSITIVE (1)}}
    Artistically speaking, this is a \textcolor{ForestGreen}{beautiful movie}--the \textcolor{ForestGreen}{cinematography, music and costumes are gorgeous}. In fact, this movie is \textcolor{ForestGreen}{prettier} than those directed by Akira Kurasawa himself. In this case, he only wrote the movie as it was made several years after his death.So, as far as the writing goes, the dialog was \textcolor{ForestGreen}{well-written} and the story, at times, was \textcolor{ForestGreen}{interesting}. However, the story was also rather \textcolor{red}{depressing yet uninvolving} in some ways--after all, \textcolor{orange}{it's the story of a group of women who work in a brothel. It's interesting that although prostitution has been seen as a much more acceptable business in Japan, the women STILL long for a better life. }This reminds me a lot of the movie Streets Of Shame, though Streets Of Shame's characters are a lot less likable and more one-dimensional.So, overall it gets a 7--mostly due to everything BUT the writing. It's too bad that the \textcolor{red}{weakest link in this movie is the story} by the great Kurasawa.

    }%
}

\textbf{HARD: Average review length - 124 words}
- Hard to predict review. Very positive sentiment with a negative label. Might reflect mislabeling.

\noindent\fbox{%
    \parbox{\textwidth}{%
    \textcolor{red}{\textbf{NEGATIVE (0)}}
   this one of the \textcolor{ForestGreen}{best celebrity's reality shows a ever saw}. we can see the concerts we can see the life of Britney, i \textcolor{ForestGreen}{love the five episodes}. i was always being surprised by Britney and the subjects of the show i think that some people don't watch the show at all we can how a great person she his. she his really funny really gentle and she loves her fans and we can see how she loves her work. i just don't give a 10 because of k-fed he his a real jerk he doesn't seem to like Britney at all. I they make a second season of this \textcolor{ForestGreen}{great show} because it shows at some people how Britney really is. Go Britney \textcolor{ForestGreen}{your the best} and you will never leave our hearts.
    }%
}

\noindent\fbox{%
    \parbox{\textwidth}{%
    \textcolor{ForestGreen}{\textbf{POSITIVE (1)}}
   Okay, first of all I got this movie as a Christmas present so it was FREE! FIRST - This movie was meant to be in stereoscopic 3D. It is for the most part, but whenever the main character is in her car the movie falls flat to 2D! What!!?!?! It's not that hard to film in a car!!! SECOND - The \textcolor{red}{story isn't very good}. There are a \textcolor{red}{lot of things wrong with it}. THIRD - Why are they showing all of the deaths in the beginning of the film! It made the \textcolor{red}{movie suck} whenever some was going to get killed!!! Watch it for a good laugh , but \textcolor{red}{don't waste your time buying it}. Just download it or something for cheap

    }%
}

\noindent\fbox{%
    \parbox{\textwidth}{%
    \textcolor{red}{\textbf{NEGATIVE (0)}}
    this is a \textcolor{ForestGreen}{great movie}. I \textcolor{ForestGreen}{love the series} on tv and so I \textcolor{ForestGreen}{loved the movie}. One of the best things in the movie is that Helga finally admits her deepest darkest secret to Arnold!!! that was \textcolor{ForestGreen}{great}. \textcolor{ForestGreen}{i loved it it was pretty funny too}. It's a \textcolor{ForestGreen}{great movie!} Doy!

    }%
}

\subsection{Data-IQ on images (computer vision)}\label{cv}

\paragraph{Goal.} We wish to assess the generality of Data-IQ. Whilst, in the main paper we mainly tackled tabular data, we wish to showcase the potential utility of Data-IQ on image data for computer vision tasks.

\paragraph{Experiment.} We train a convolutional neural network on the CelebA dataset, where the task is to predict the gender based on the image of the celebrity. We then use Data-IQ to analyze the training dynamics and categorize the data into subgroups - $\easy$, $\ambiguous$~ and $\hard$

\begin{figure}[!h]
    \centering
    \includegraphics[width=0.75\textwidth]{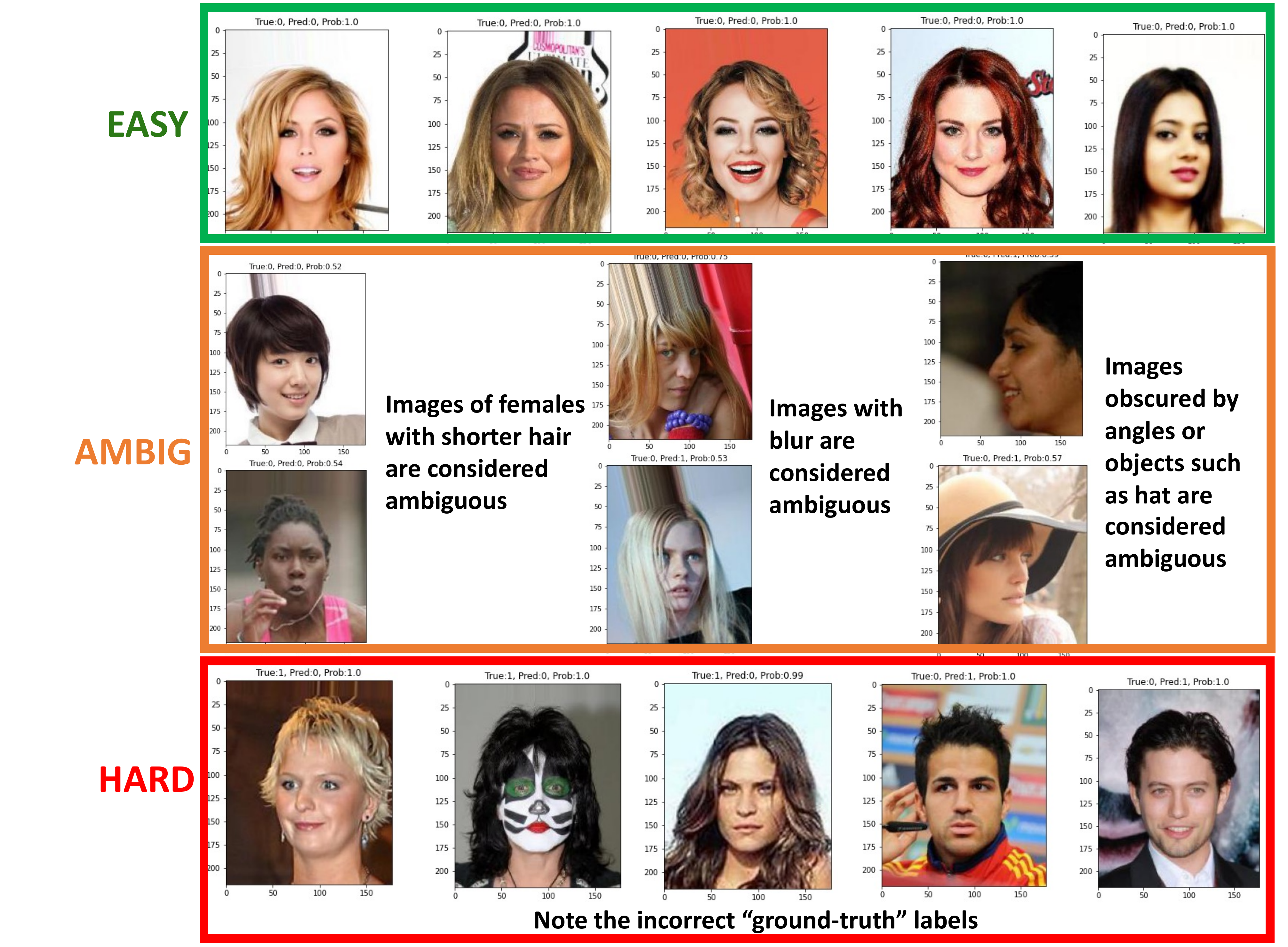}
    \caption{Subgroups highlighted by Data-IQ on the CelebA dataset}
    \label{fig:cv_task}
\end{figure}

\paragraph{Takeaway.} We notice that the $\easy$, $\ambiguous$~ and $\hard$ groups represent different phenomena in the image data. We summarize the findings in Figure \ref{fig:cv_task}

\newpage
\subsection{Effect of averaging over the uncertainty metric}

\paragraph{Goal.} We wish to assess the impact of averaging/taking an expectation over the metric such as aleatoric uncertainty over training. Specifically, we assess the case where an example has high variance at the beginning of training and flattens with time, compared to an example that remains at a medium level throughout training. These represent the EASY/HARD (high variance in the beginning and then flattening) and AMBIGUOUS  (medium level throughout) subgroups respectively. The question is, by taking an average will the uncertainties be similar, despite the training dynamics being different? If the average uncertainties are not similar that means we can distinguish the groups with different training dynamics.

\paragraph{Experiment.} We conduct an experiment to assess this by sampling 100 examples from each of these groups and plot the following three metrics per sample:
\begin{itemize}
    \item  Training dynamic (i.e. the aleatoric uncertainty over training epochs).
    \item The distribution over aleatoric uncertainty scores across all training steps.
    \item The “average” aleatoric uncertainty. 
\end{itemize}

\paragraph{Takeaway.} The result as shown in Figure \ref{fig:avg_exp} below shows that even by taking an average, there is still a distinct separation to be able to distinguish between the two types of samples with different training dynamics. This suggests that averaging over the uncertainty is indeed a useful metric to evaluate the uncertainty of an individual data sample.

\begin{figure}[!h]
    \centering
    \includegraphics[width=1\textwidth]{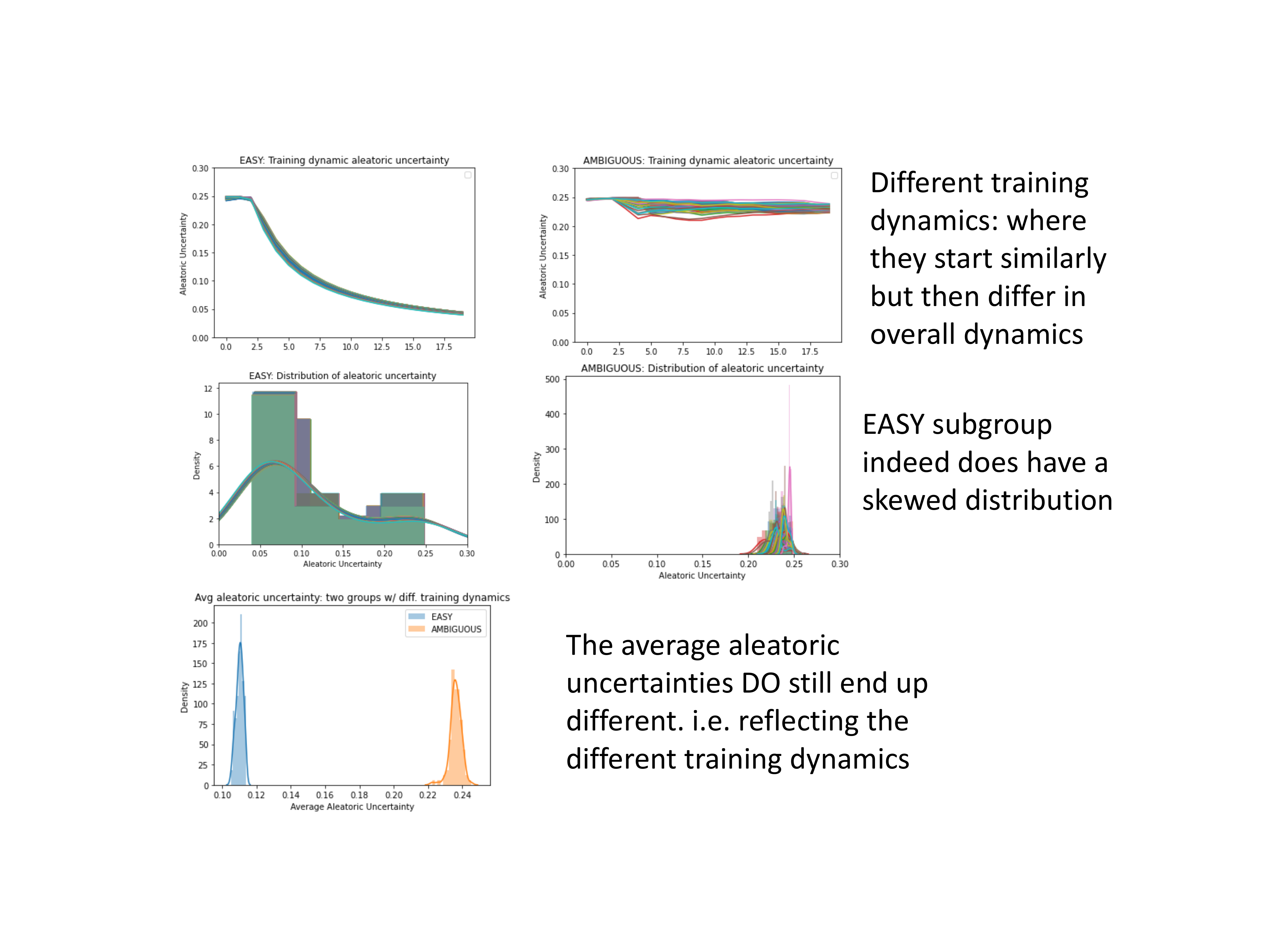}
    \vspace{-2cm}
    \caption{Assessment of whether we can distinguish groups with different training dynamics when we average over the metric}
    \label{fig:avg_exp}
\end{figure}

\subsection{Effect of additional sources of randomness}

\paragraph{Goal.} We wish to assess the impact of incorporating additional sources of randomness - for example dropout.

\paragraph{Experiment.} We have test the model parameterizations from Section 4.1, but additionally trained an additional 3 models with the same architecture, yet with the inclusion of different dropout masks with dropout probability: $p=0.1, p=0.2, p=0.3$. 

We then assess the Spearman correlation of the metrics across all model combinations, as was done in the experiment on robustness to variation from Section 4.1. The higher correlation, implies better robustness to variations. 

\paragraph{Takeaway.}  Data-IQ remains most robust to variation, even with this additional randomness. We highlight the results on all four datasets below in Table \ref{dropout_exp}.

\begin{table}[!h]
 \centering
\caption{Comparison of robustness/consistency across different models on the basis of the Spearman rank correlation}
\scalebox{1}{
\begin{tabular}{ccc}
\toprule
Dataset &  (Ours) Data-IQ & Data Maps    \\ \hline
\midrule
Covid & \bf $0.93 \pm 0.07$  & $0.66 \pm 0.16$\\ \hline
Support & \bf $0.92 \pm 0.02$  & $0.81 \pm 0.11$\\ \hline
Prostate & \bf $0.97 \pm 0.01$  &$0.96 \pm 0.01$ \\ \hline
Fetal & \bf $0.86 \pm 0.06$  & $0.81 \pm 0.13$
\\ \hline
\bottomrule
\end{tabular}}
\label{dropout_exp}
\end{table}

\subsection{Effect of hyper-parameters on the recovery of estimated groups}

\paragraph{Goal.} We wish to assess how model choices/parameterizations might affect the recovery of the different subgroups. i.e. we desire that if a sample is categorized as EASY with one parameterization, that similarly it should be categorized as EASY for another parameterization. Similarly, for the other subgroups - we desire that their subgroup characterization stays the same across different model parameterizations. 

\paragraph{Experiment.} We use the same sets of model parameterizations as assessed in Section 4.1 (Robustness to variation experiment). We then compute the overlap of sample ids for each subgroup ($\easy$, $\ambiguous$, $\hard$) between the different parameterizations. This can be thought of as accuracy or sample level stability for assignment to a specific subgroup. 

\paragraph{Takeaway.} We present the results across all four datasets below in Table \ref{overlap} which suggest that recovery of subgroups is stable, with more than 90\% of samples being assigned to the same subgroup between models.

\begin{table}[!h]
 \centering
\caption{Assessment of stability in group assignment across different parameterizations. The score reflects the average overlap of id characterization}
\scalebox{1}{
\begin{tabular}{ccc}
\toprule
Dataset &  (Ours) Data-IQ     \\ \hline
\midrule
Covid & \bf $0.94 \pm 0.03$ \\ \hline
Support & \bf $0.91 \pm 0.01$ \\ \hline
Prostate & \bf $0.95 \pm 0.01$   \\ \hline
Fetal & \bf $0.77 \pm 0.14$ \\ \hline
\bottomrule
\end{tabular}}
\label{overlap}
\end{table}

\subsection{Correlation of weights through training}

\paragraph{Goal.} We wish to assess the correlation of weights through model training.

\paragraph{Experiment.} To provide this assessment we follow Jin et al \cite{jin2020does}, where weight correlation is defined as the average cosine similarity between weight vectors of neurons. We conduct experiment to assess the correlation of model weights over epochs $e \in \{1,2...E\}$. i.e. we assess the average correlation of model weights $(e_1, e_2), (e_2, e_3)$ and so on. 

\paragraph{Takeaway.} We note that \cite{jin2020does}, shows that LOW correlation is less than 0.2. Our validation shows that the weights have LOW correlation, with an average weight correlation across the 4 datasets of $0.1 \pm 0.02$. Thus, even though our goal is different from a typical sampling setting (i.e. we just want to measure the dispersion of observed variables) - we still wish to highlight the low correlation of weights between epochs. 

However, as a recommendation for practitioners if this is a concern: a practical solution is that the practitioner could decide to simply increase the intervals for which the dynamics are sampled, thereby decreasing this correlation even further.

\subsection{Data sculpting: Absolute number of samples}\label{sculpt-nums}

\paragraph{Goal.} When performing data sculpting as in Section 4.3 we desire to know the absolute number of samples removed in order to quantify the importance of the effect of removing such datapoints on model performance.

\paragraph{Experiment.} We present this analysis for two scenarios:
\begin{itemize}
    \item Balanced SEER-CUTRACT: 1000 samples each
    \item Unbalanced SEER-CUTRACT: 10000 samples in SEER and 1000 in CUTRACT
\end{itemize}

\paragraph{Takeaway.}
For the balanced experiment the full Ambiguous subgroup (i.e. proportion=1) has a total of 500 samples. Hence, when we sweep the x-axis representing the proportions of Ambiguous samples; $p = \{0, 0.2, 0.4, 0.6, 0.8, 1\}$, this is reflective of $\{0, 100, 200, 300, 400, 500\}$ absolute samples. This result shows that the number of samples that can be excluded is fairly substantial.

\begin{figure*}[!h]

  \centering
  \subfigure[US-UK Balanced ]{\includegraphics[width=0.4\textwidth]{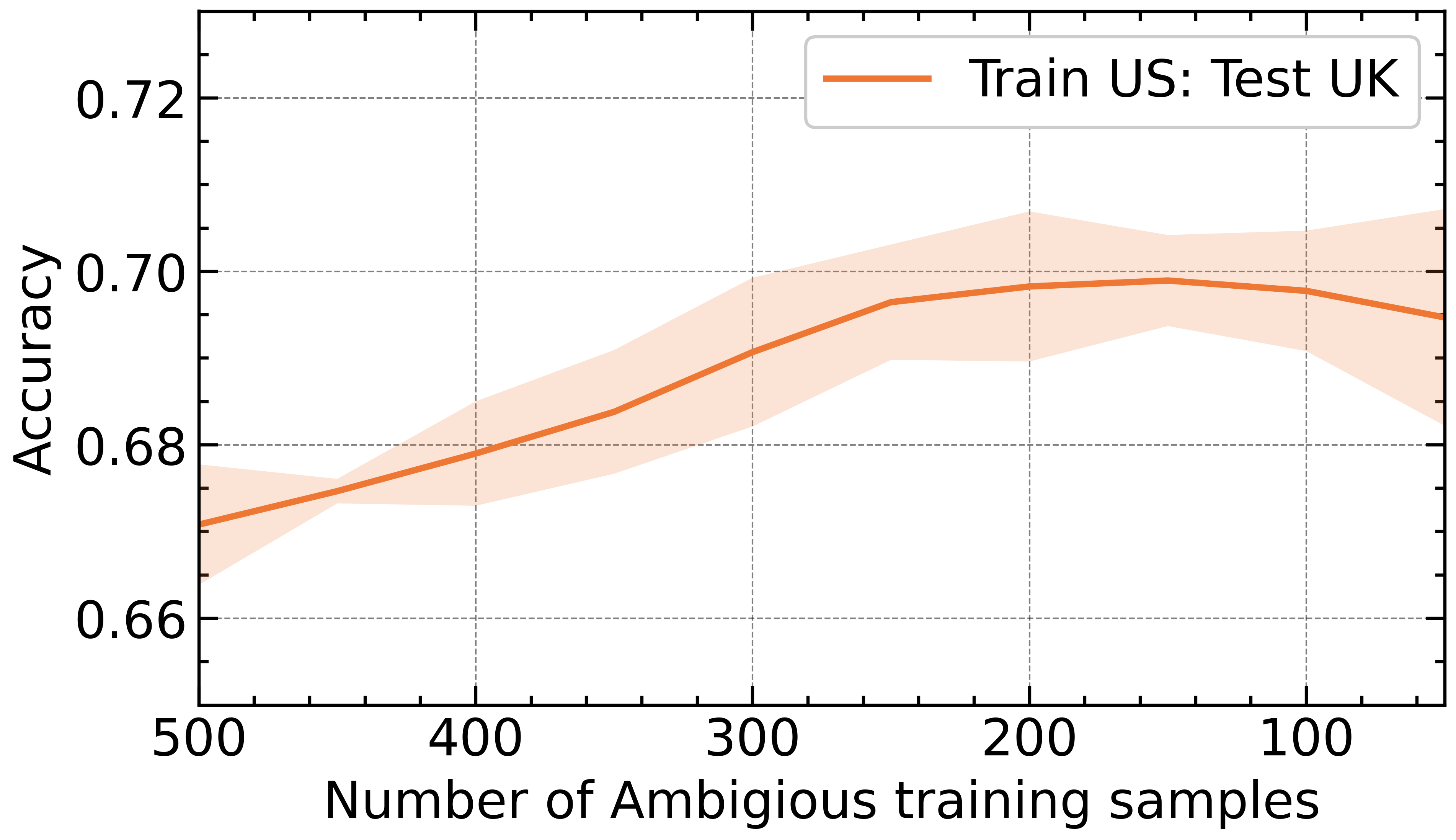}}\quad\quad
  \subfigure[US-UK Unbalanced]{\includegraphics[width=0.4\textwidth]{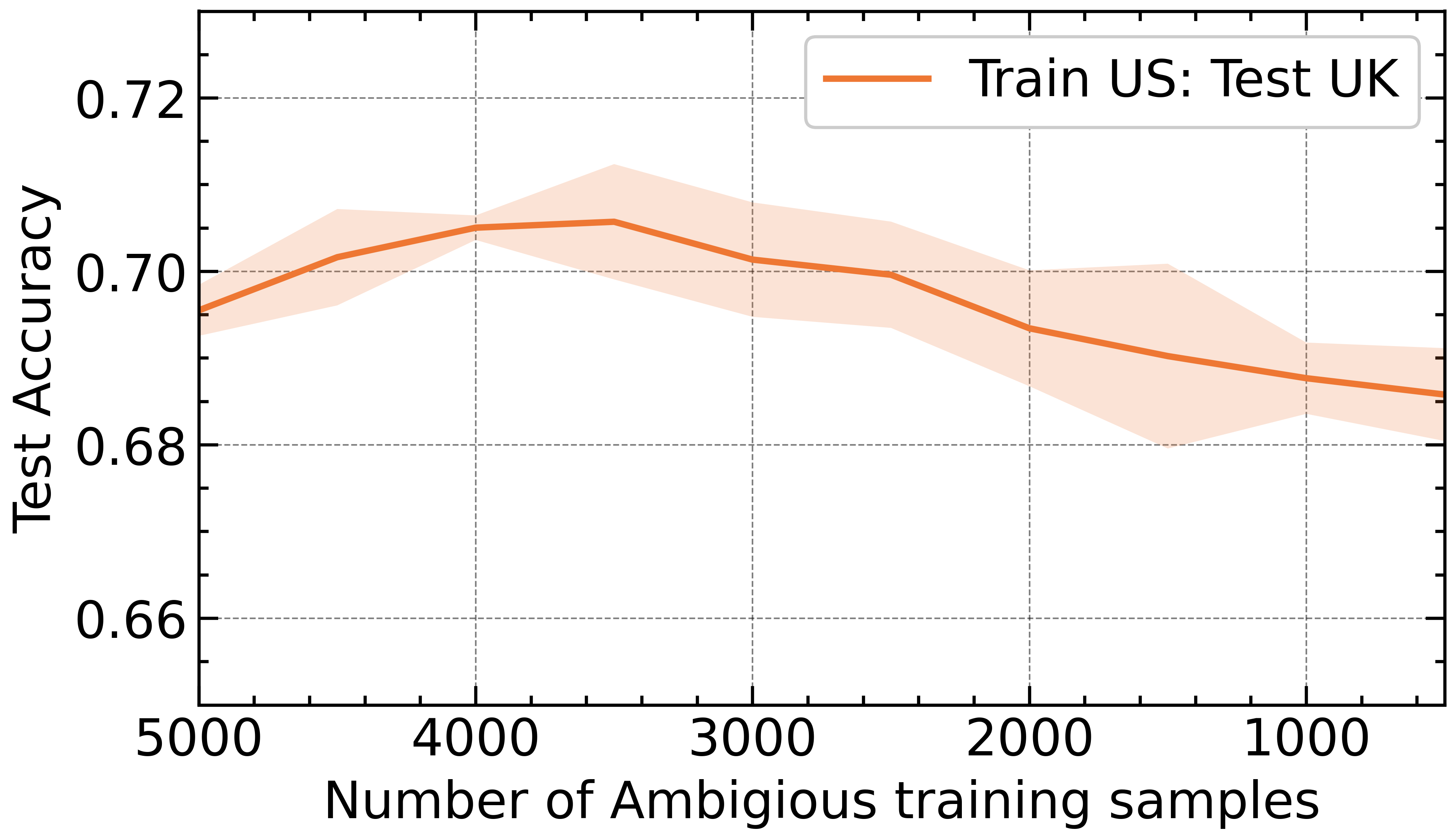}}\quad\quad
  \subfigure[UK-US Balanced]{\includegraphics[width=0.4\textwidth]{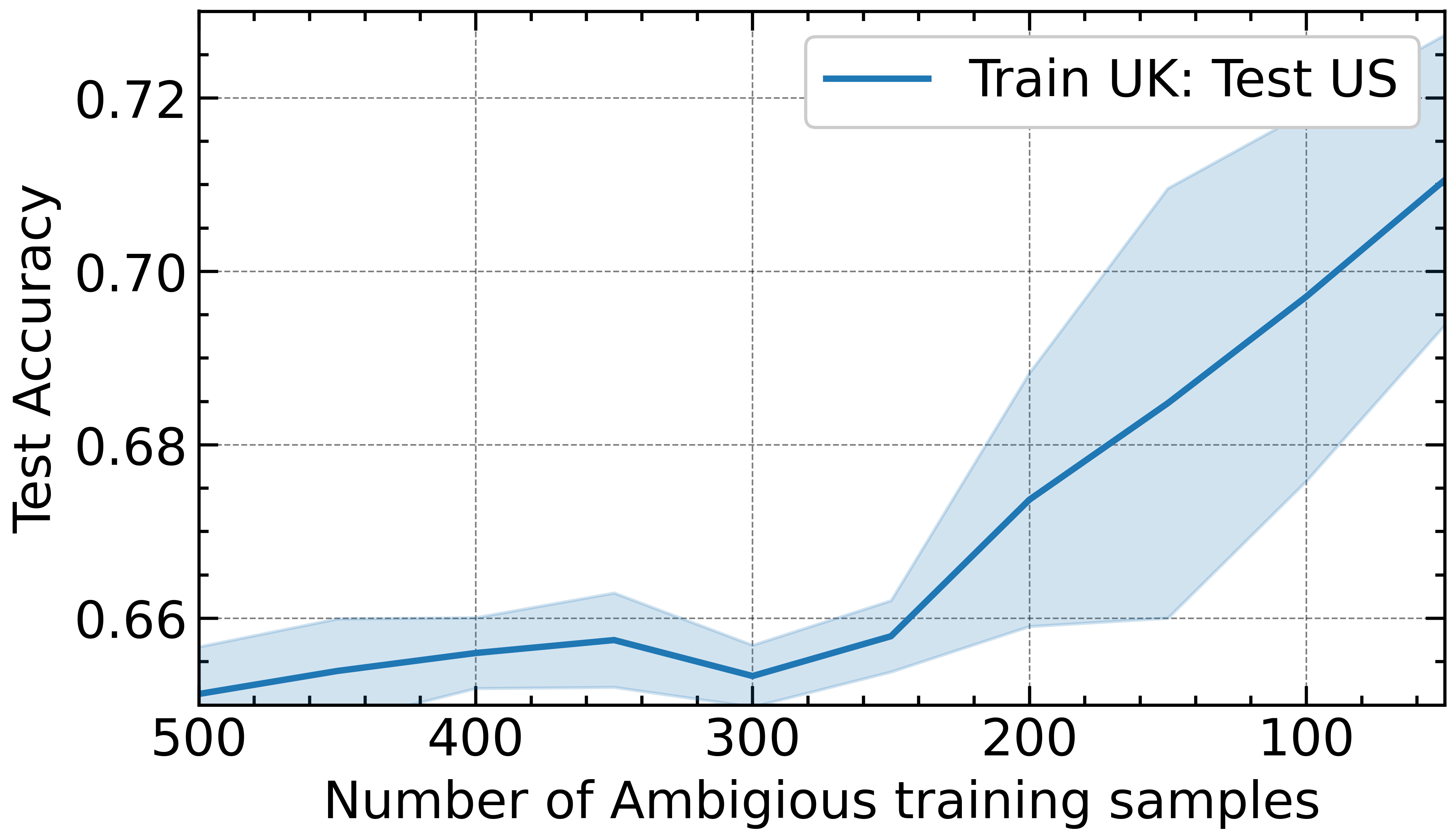}}\quad\quad
  \subfigure[UK-US Unbalanced]{\includegraphics[width=0.4\textwidth]{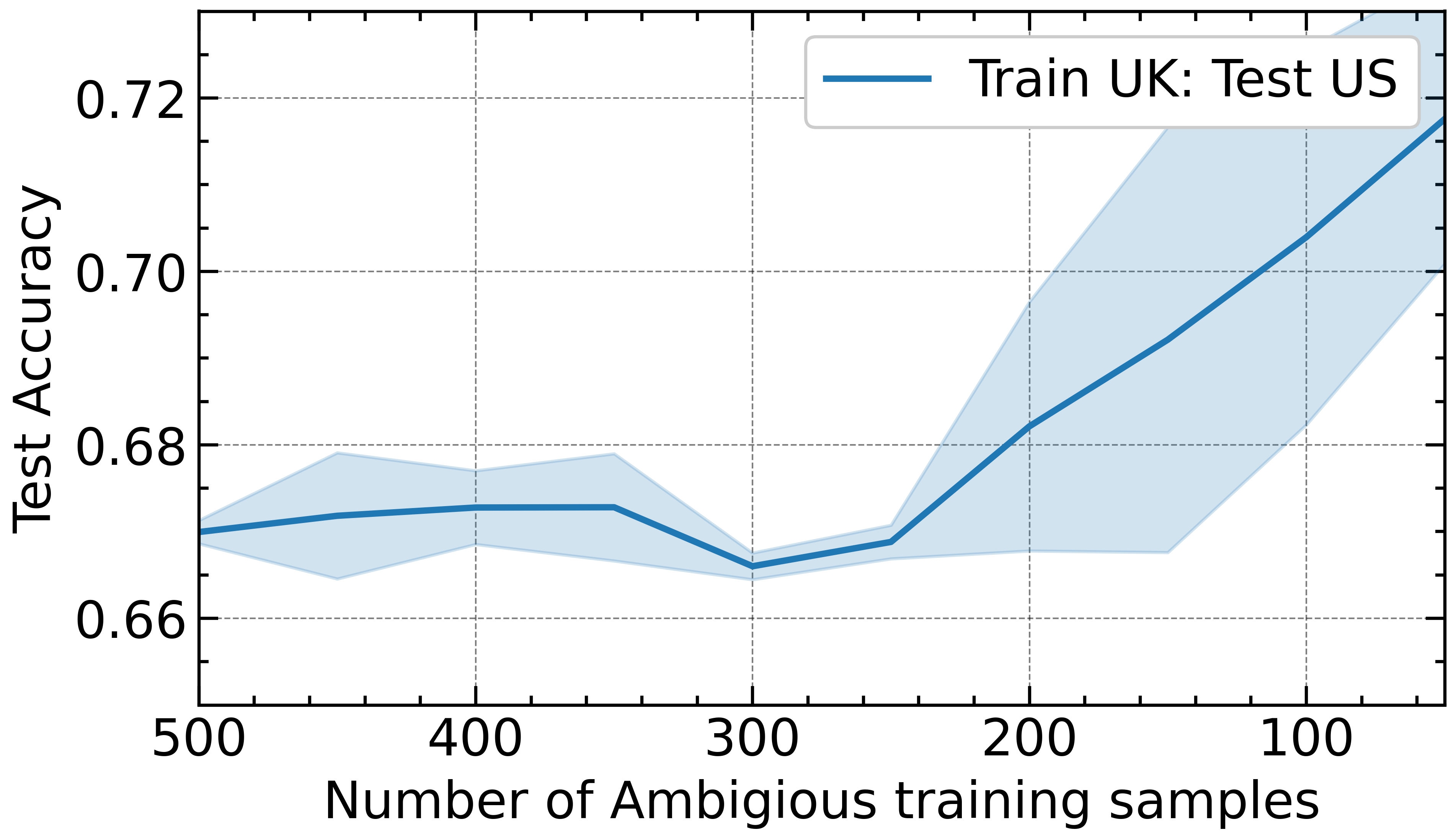}}
  \caption{Data sculpting - absolute number of $\ambiguous$ samples removed vs test accuracy}
  \label{fig:less_is_more_abs}
\end{figure*}

We compare this to the unbalanced experiment. The results are in Figure \ref{fig:less_is_more_abs}. It shows that a substantial number of Ambiguous samples can be removed to improve performance. However, in the case of (b) when there are a lot of samples, removing too many can also be harmful.

\clearpage

\end{document}